\documentclass[10pt,twocolumn,letterpaper]{article}

\usepackage[pagenumbers]{cvpr} 

\def\sd{{\mathrm{d}}}
\def\vzero{{\bm{0}}}

\def\vmu{{\bm{\mu}}}

\def\vepsilon{{\bm{\epsilon}}}

\def\mSigma{{\bm{\Sigma}}}
\def\va{{\bm{a}}}
\def\vb{{\bm{b}}}

\def\vv{{\bm{v}}}
\def\vw{{\bm{w}}}
\def\vx{{\bm{x}}}
\def\vy{{\bm{y}}}
\def\vz{{\bm{z}}}

\def\mA{{\bm{A}}}
\def\mB{{\bm{B}}}

\def\mI{{\bm{I}}}

\def\mR{{\bm{R}}}

\def\mX{{\bm{X}}}

\definecolor{cvprblue}{rgb}{0.21,0.49,0.74}
\usepackage[pagebackref,breaklinks,colorlinks,allcolors=cvprblue]{hyperref}
\usepackage{amssymb,amsmath,amsthm,bm,bbm}
\usepackage{enumitem}
\usepackage{algpseudocode}
\usepackage{algorithm}
\usepackage{booktabs,multirow}

\renewcommand{\epsilon}{\varepsilon}
\usepackage{placeins}
\usepackage{graphicx}
\usepackage{xcolor}
\newsavebox{\ffhqbox}
\newsavebox{\inetbox}
\newlength{\labelw}
\newlength{\imgw}

\newcommand{\dsetlabel}[2]{%
  \begingroup\setlength{\fboxsep}{0pt}%
  \colorbox{gray!15}{%
    \parbox[b][\dimexpr\ht#2+\dp#2\relax][c]{\labelw}{%
      \centering\rotatebox{90}{\small\textbf{#1}}}}%
  \endgroup}

\DeclareMathOperator{\dist}{dist}

\newtheorem{proposition}{Proposition}
\newtheorem{theorem}{Theorem}
\newtheorem{corollary}{Corollary}

\def\confName{CVPR}
\def\confYear{2027}

\title{Denoising Diffusion with Progressive Posterior Sampling}

\author{Zeyu Luo, Shuai Huang\\
Auburn University\\
{\tt\small zel0010@auburn.edu, shuai.huang@auburn.edu}
}

\begin{document}

\maketitle

\begingroup
\makeatletter
\let\addcontentsline\@gobblethree

\begin{abstract}
Diffusion models provide effective generative priors for solving noisy linear inverse problems. However, existing posterior sampling methods may require singular value decomposition (SVD) of the measurement operator or prior knowledge of the measurement noise variance, while guidance based on local gradients provides only a limited view of posterior geometry. We propose progressive posterior sampling (PPS) with manifold projection, which automatically estimates the measurement noise variance and avoids SVD. During reverse diffusion, PPS constructs a Gaussian approximation to the posterior distribution in closed form and guides a sample drawn from the local prior toward the approximate posterior mean. We then project the guided sample onto a spherical level set that approximates the local noisy image manifold, promoting consistency with the local geometry of the diffusion prior. Experiments on FFHQ and ImageNet across four restoration tasks and three measurement noise levels demonstrate competitive reconstruction accuracy and perceptual quality by PPS. 
\end{abstract}    
\vspace{-1em}

\section{Introduction}
\label{sec:intro}
Diffusion models implicitly represent data distributions through iterative denoising and have achieved state-of-the-art performance in image generation and restoration~\cite{sohl2015deep,song2020improved,ho2020denoising,rombach2022latentdiffusion}. Many image restoration tasks, including super-resolution, deblurring, and inpainting, can be formulated as ill-posed linear inverse problems that seek to recover an underlying image from noisy measurements. Solving these problems via posterior sampling requires characterizing the posterior distribution of the image given the measurements. However, training a dedicated diffusion model to learn this posterior for every measurement setting and restoration task is impractical. Instead, the posterior can be expressed using Bayes' theorem by combining an image prior learned by a diffusion model with the measurement likelihood specified by the imaging model.

Diffusion-based posterior sampling methods for noisy linear inverse problems operate in either the spectral domain or the image domain. Kawar \etal~\cite{kawar2022denoising} proposed Denoising Diffusion Restoration Models (DDRM), which transform the measurements and the signal into their respective spectral domains using the singular value decomposition (SVD) of the measurement matrix. This formulation enables the exact derivation of the variational distributions used for posterior sampling. However, computing the SVD can be computationally expensive for general measurement operators, limiting the applicability of DDRM to complex motion-blur models. DDRM also requires knowledge of the measurement noise variance, which is often unknown in practice. In the image domain, Chung \etal~\cite{chung2023diffusion,chung2022improving,chung2024decomposed} proposed Diffusion Posterior Sampling (DPS), which uses manifold-constrained gradients. DPS first performs an ancestral sampling step and then updates the resulting sample along the tangent space of the noisy-data manifold. Its update direction is locally determined, and the gradient step-size schedule must be carefully tuned to prevent the sample from drifting off the manifold. He \etal~\cite{he2024manifold} subsequently proposed Manifold Preserving Guided Diffusion (MPGD), which updates the estimated clean signal along the clean-data manifold before ancestral sampling, thereby improving robustness. Without relying on the manifold hypothesis, Wang \etal~\cite{wang2023zeroshot} proposed the Denoising Diffusion Null-Space Model (DDNM), which directly refines the estimated clean signal through range-space correction. However, this approach also requires knowledge of the measurement noise variance and an SVD of the measurement matrix to compute scaling matrices for the correction term and the reverse diffusion variance.

To address these challenges, we propose progressive posterior sampling (PPS) with manifold projection, which automatically estimates the measurement noise variance and avoids SVD of the forward operator. Our method derives a closed-form Gaussian approximation to the posterior distribution and uses its mean to determine a more robust update direction. We further approximate the local noisy image manifold of the pretrained diffusion model by a spherical level set, and project each guided sample onto this set to promote consistency with the local geometry of the diffusion prior. Experiments demonstrate competitive performance across four restoration tasks, with PPS consistently outperforming the evaluated baselines for motion deblurring across all tested noise levels.
\section{Background}
\label{sec:bg}
\subsection{Noisy Linear Inverse Problem}

We aim to recover a signal of interest $\vx\in\mathbb{R}^N$ from noisy linear measurements $\vy\in\mathbb{R}^M$:
\begin{align}
\label{eq:linear_inverse}
    \vy=\mA\vx+\vepsilon,
\end{align}
where $\mA\in\mathbb{R}^{M\times N}$ is the measurement matrix and $\vepsilon\sim\mathcal{N}(0,\sigma_\epsilon^2\mI)$ denotes additive white Gaussian noise with variance $\sigma_\epsilon^2$. Here, $\mI$ is the $M\times M$ identity matrix. This inverse problem is often ill-posed due to insufficient measurements, and measurement noise further complicates recovery. Prior information about the signal is therefore needed to regularize the reconstruction. We assume that $\vx$ follows a prior distribution $p(\vx)$ learned by a generative diffusion model~\cite{song2021scorebased}. Reconstruction can then be performed by sampling from the posterior distribution $p(\vx|\vy)\propto p(\vy|\vx)p(\vx)$, where $p(\vy|\vx)$ is the measurement likelihood specified by the model in \eqref{eq:linear_inverse}.

\subsection{Diffusion Model}

The forward diffusion process gradually transforms a clean signal into noise and can be modeled by the It\^{o} stochastic differential equation (SDE)~\cite{song2021scorebased}:
\begin{align}
    \sd\vx_t=f(\vx_t,t)\sd t+g(t)\sd\vw_t,
\end{align}
where $f:\mathbb{R}^N\times[0,1]\rightarrow\mathbb{R}^N$ is the drift coefficient, $g:[0,1]\rightarrow\mathbb{R}$ is the diffusion coefficient, and $\vw_t$ is an $N$-dimensional standard Wiener process (Brownian motion). The corresponding reverse diffusion process is governed by the reverse-time SDE~\cite{anderson1982reverse}:
\begin{align}
\label{eq:reverse_sde_prior}
    \sd\vx_t\!=\!\left[f(\vx_t,t)\!-\!g(t)^2\nabla_{\vx_t}\log p_t(\vx_t)\right]\!\sd t\!+\!g(t)\sd\bar{\vw}_t,
\end{align}
where $p_t$ denotes the marginal density of $\vx_t$ and $\bar{\vw}_t$ is a standard Wiener process in reverse time. This SDE is integrated from $t=1$ to $t=0$. The time-dependent score function $\nabla_{\vx_t}\log p_t(\vx_t)$ is approximated by a neural network $s_{\theta}(\vx_t,t)$ trained using a weighted denoising score matching objective~\cite{vincent2011connection,song2021scorebased}:
\begin{align}
\begin{split}
    \theta^*=\arg\min_{\theta}\;&
    \mathbb{E}_t\Bigl[\lambda_t
    \mathbb{E}_{\vx_0}\mathbb{E}_{\vx_t|\vx_0}
    \bigl[\|s_{\theta}(\vx_t,t)\\
    &\quad-\nabla_{\vx_t}\log p_{t|0}(\vx_t|\vx_0)\|_2^2\bigr]\Bigr],
\end{split}
\end{align}
where $t$ is sampled uniformly from $[0,1]$, $\lambda_t$ is a weighting function, and the signal samples are drawn according to $\vx_0\sim p(\vx_0)$ and $\vx_t\sim p_{t|0}(\vx_t|\vx_0)$. Here, $p_{t|0}$ denotes the transition density of the forward diffusion process.

In the discrete-time setting, diffusion models progressively corrupt clean training data by adding noise according to a prescribed schedule and learn to reverse this process through iterative denoising. We use the subscript $i$ to denote the discrete step. Two representative approaches, Denoising Diffusion Probabilistic Models (DDPM)~\cite{sohl2015deep,ho2020denoising} and Score Matching with Langevin Dynamics (SMLD)~\cite{song2020improved}, differ in their forward diffusion processes~\cite{song2021scorebased}:
\begin{itemize}
    \item Starting from the clean signal $\vx_0$, DDPM rescales the signal and adds Gaussian noise at each step:
    \begin{align}
        \vx_i=\sqrt{1-\beta_i}\vx_{i-1}+\sqrt{\beta_i}\vz_{i-1},\quad i=1,\ldots,I,
    \end{align}
    where the noise vectors $\vz_{i-1}$ are independent samples from $\mathcal{N}(\vzero,\mI)$. The noise schedule $\{\beta_i\}_{i=1}^{I}$, with $0<\beta_i<1$, is chosen so that the terminal state $\vx_I$ is approximately distributed as $\mathcal{N}(\vzero,\mI)$. With $\alpha_i=1-\beta_i$ and $\overline{\alpha}_i=\prod_{j=1}^i\alpha_j$, the corrupted signal at any fixed step $i$ can be sampled directly as
    \begin{align}
        \label{eq:forward_diffusion_xi_x0}
        \vx_i=\sqrt{\overline{\alpha}_i}\vx_0+\sqrt{1-\overline{\alpha}_i}\vz,\quad\vz\sim\mathcal{N}(\vzero,\mI).
    \end{align}

    \item SMLD adds independent Gaussian noise increments directly to the signal:
    \begin{align}
        \vx_i=\vx_{i-1}+\sqrt{\sigma_i^2-\sigma_{i-1}^2}\vz_{i-1},\quad i=1,\ldots,I,
    \end{align}
    where the noise schedule $\{\sigma_i\}_{i=0}^{I}$ satisfies $0=\sigma_0<\sigma_1<\cdots<\sigma_I$. The terminal noise level $\sigma_I$ is chosen to be sufficiently large relative to the signal scale so that $\vx_I$ is approximately distributed as $\mathcal{N}(\vzero,\sigma_I^2\mI)$. 
\end{itemize}

\subsection{Posterior Sampling via the Diffusion Model}
\label{subsec:posterior_sampling_prior_work}

The reverse-time SDE in \eqref{eq:reverse_sde_prior} generates samples from the prior distribution $p(\vx)$. To solve the linear inverse problem in \eqref{eq:linear_inverse}, it can be adapted to sample from the posterior distribution $p(\vx|\vy)$. As discussed earlier, training a dedicated neural network to approximate the posterior score $\nabla_{\vx_t}\log p_t(\vx_t|\vy)$ for every measurement setting and imaging task is impractical. Instead, we apply Bayes' theorem to decompose the posterior score as $\nabla_{\vx_t}\log p_t(\vx_t|\vy)=\nabla_{\vx_t}\log p_t(\vx_t)+\nabla_{\vx_t}\log p_t(\vy|\vx_t)$. Substituting this decomposition yields the reverse-time SDE for posterior sampling~\cite{song2021scorebased}:
\begin{align}
\label{eq:reverse_sde_posterior}
\begin{split}
    \sd\vx_t={}&\Bigl[f(\vx_t,t)-g(t)^2
    \bigl(\nabla_{\vx_t}\log p_t(\vx_t)\\
    &\quad+\nabla_{\vx_t}\log p_t(\vy|\vx_t)\bigr)\Bigr]\sd t
    +g(t)\sd\bar{\vw}_t.
\end{split}
\end{align}
The prior score $\nabla_{\vx_t}\log p_t(\vx_t)$ is approximated by the pretrained score network $s_{\theta}(\vx_t,t)$. The main challenge lies in evaluating the likelihood score $\nabla_{\vx_t}\log p_t(\vy|\vx_t)$, since the likelihood involves marginalizing over the clean signal~\cite{chung2023diffusion}:
\begin{align}
\label{eq:measurement_likelihood}
    p_t(\vy|\vx_t)=\textstyle\int p(\vy|\vx_0)\,p(\vx_0|\vx_t)\,\sd\vx_0.
\end{align}
Here, $p(\vy|\vx_0)=\mathcal{N}(\vy;\mA\vx_0,\sigma_\epsilon^2\mI)$ is the Gaussian measurement likelihood specified by \eqref{eq:linear_inverse}, whereas the conditional distribution $p(\vx_0|\vx_t)$ is generally intractable.

\begin{figure*}[htbp]
  \centering
  \vspace{-1em}
  \begin{subfigure}{0.3\linewidth}
    \includegraphics[height=0.8333\linewidth]{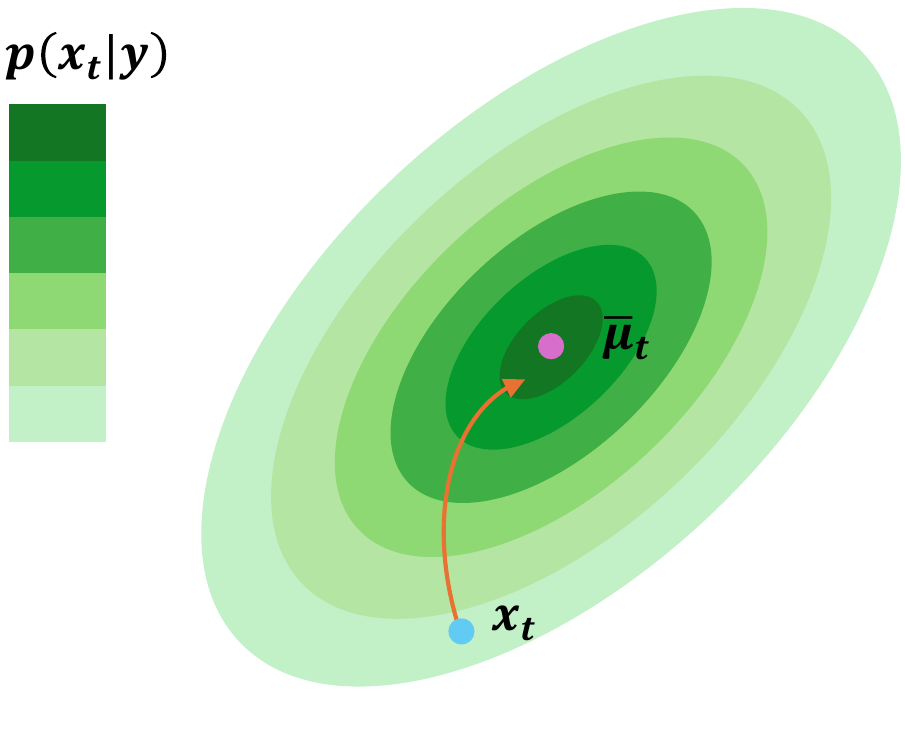}
    \caption{Continuous reverse-time SDE.}
    \label{fig:continuous_rt_sde_c}
  \end{subfigure}
  \hspace{2em}
  \begin{subfigure}{0.25\linewidth}
    \includegraphics[height=\linewidth]{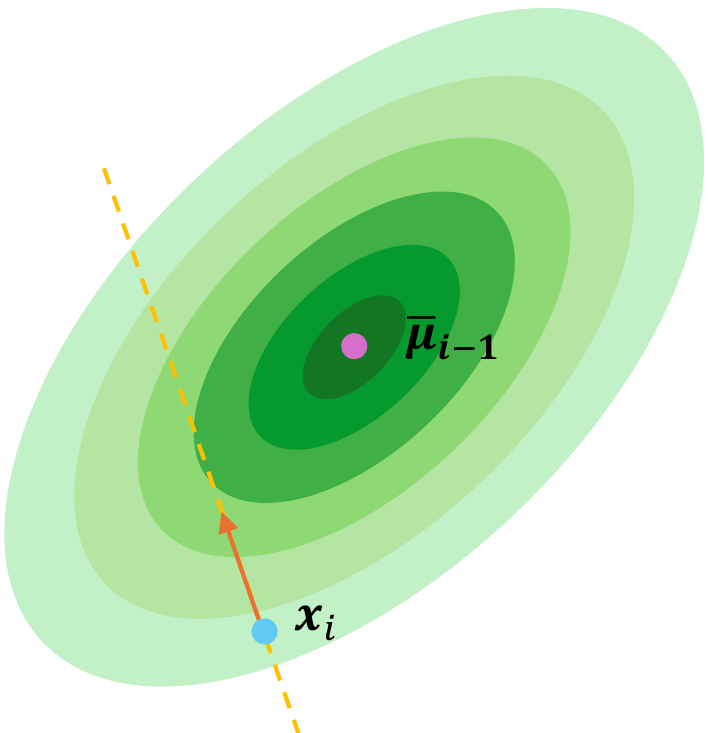}
    \caption{Conventional discrete update.}
    \label{fig:discrete_rt_sde_dps_c}
  \end{subfigure}
  \hspace{2em}
  \begin{subfigure}{0.25\linewidth}
    \includegraphics[height=\linewidth]{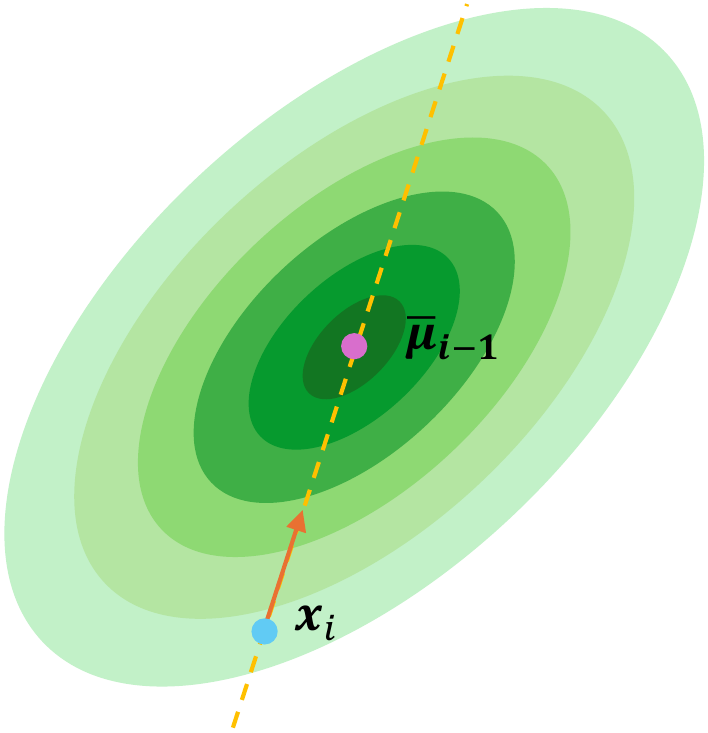}
    \caption{Proposed discrete update.}
    \label{fig:discrete_rt_sde_pps_c}
  \end{subfigure}
  \caption{Schematic comparison of continuous and discrete guidance in diffusion posterior sampling. For illustration, the posterior density is held fixed and the stochastic term $g(t)\sd\bar{\vw}_t$ is omitted. Darker green indicates higher posterior density; blue and magenta points denote the current samples and posterior means, respectively. (a) The continuous guidance direction changes as the state evolves, producing a curved path toward higher posterior density. (b) A conventional discrete update follows a locally determined direction that does not point toward the posterior mode. (c) The proposed discrete update directs the update toward $\overline{\vmu}_{i-1}$, the mean and mode of the Gaussian posterior approximation.}
  \vspace{-1em}
  \label{fig:rt_sde}
\end{figure*}

\subsection{Related Work} 

Several approaches approximate $p(\vx_0|\vx_t)$ to obtain an analytically tractable measurement likelihood $p_t(\vy|\vx_t)$. In DPS, Chung \etal~\cite{chung2023diffusion} approximate the denoising posterior by a Dirac delta centered at the posterior mean $\widehat{\vx}_0(t)=\mathbb{E}[\vx_0|\vx_t]$, yielding $p(\vx_0|\vx_t)\approx\delta(\vx_0-\widehat{\vx}_0(t))$. This approximation effectively sets the posterior variance to zero, neglecting posterior uncertainty and potentially making DPS sensitive to measurement noise and the choice of guidance step size. To account for posterior uncertainty, $\Pi$GDM~\cite{song2023pseudoinverse} uses the Gaussian approximation $p(\vx_0|\vx_t)\approx\mathcal{N}(\vx_0;\widehat{\vx}_0(t),u_t^2\mI)$, where $u_t^2=\frac{\sigma_t^2}{\sigma_t^2+1}$ is derived under the simplifying assumption of a standard Gaussian data prior, $\vx_0\sim\mathcal{N}(\vzero,\mI)$. Zhang \etal~\cite{zhang2025improving} proposed Decoupled Annealing Posterior Sampling (DAPS), which also adopts an isotropic Gaussian approximation centered at $\widehat{\vx}_0(t)$, but chooses $u_t^2$ heuristically. Peng \etal~\cite{peng2024improving} and Rissanen \etal~\cite{rissanen2024free} both further proposed more accurate estimations of the posterior covariance. Boys \etal~\cite{boys2023tweedie} proposed another Gaussian approximation obtained by minimizing the KL divergence between the true posterior and its approximation. These Gaussian approximations use Tweedie's denoiser $\widehat{\vx}_0(t)$ as the posterior mean. For methods that differentiate through this estimate when evaluating $\nabla_{\vx_t}\log p_t(\vy|\vx_t)$, the computation involves the denoiser Jacobian $\frac{\partial\widehat{\vx}_0(t)}{\partial\vx_t}$ and typically requires backpropagation through the pretrained network.

To avoid the memory and computational costs of backpropagating through the denoiser, Meng and Kabashima~\cite{meng2022diffusion} proposed DMPS, which assumes a flat, uninformative data prior and uses the approximation $p(\vx_0|\vx_t)\approx\mathcal{N}(\vx_0;\frac{\vx_t}{\sqrt{\overline{\alpha}_t}},\frac{1-\overline{\alpha}_t}{\overline{\alpha}_t}\mI)$. CoDPS, proposed by Yismaw \etal~\cite{yismaw2005gaussian}, retains $\widehat{\vx}_0(t)$ as the posterior mean while reducing computational cost by using a closed-form expression for the denoiser Jacobian derived under the isotropic Gaussian prior assumption $\vx_0\sim\mathcal{N}(\vzero,\sigma_0^2\mI)$.
\section{Progressive Posterior Sampling}
\label{sec:pps}

A challenge in implementing the reverse-time SDE in \eqref{eq:reverse_sde_posterior} for posterior sampling is the discrepancy between continuous dynamics and discrete updates. In continuous time, the drift responds to the evolving state, allowing the guidance direction to change along the trajectory, as illustrated in \cref{fig:continuous_rt_sde_c}. In practice, however, sampling follows a finite noise schedule $\{\beta_i\}_{i=1}^I$. A common two-step formulation first performs an unconditional sampling step using the pretrained diffusion model and then applies a gradient ascent correction to increase the measurement log-likelihood $\log p_{i-1}(\vy|\vx_{i-1})$ within the local tangent space of the noisy image manifold. This correction relies on local gradient information. Even when the unconditional transition is modeled by an isotropic Gaussian, conditioning on $\vy$ can yield an anisotropic posterior $p_{i-1}(\vx_{i-1}|\vy)$. In such a landscape, the local correction does not point directly toward the posterior mode, as illustrated in \cref{fig:discrete_rt_sde_dps_c}. Because each transition from step $i$ to step $i-1$ uses a finite update, it may not capture changes in the guidance direction within that interval. Adjusting the guidance step size alone changes only the distance traveled along the same local direction.

To address this limitation, we approximate $p_{i-1}(\vx_{i-1}|\vy)$ by a Gaussian distribution whose parameters can be computed in closed form. The mean of this approximation, $\overline{\vmu}_{i-1}$, coincides with its mode and therefore provides an explicit target for the guidance update. Our proposed progressive posterior sampling (PPS) method retains the two-step structure. First, we perform an unconditional sampling step using the pretrained diffusion model. Second, we move the sample toward $\overline{\vmu}_{i-1}$, as illustrated in \cref{fig:discrete_rt_sde_pps_c}, and project the updated sample back onto the noisy image manifold.

\subsection{Approximated Posterior Distribution}

As discussed above, the mean of the Gaussian posterior approximation coincides with its mode and provides a target for the guidance update. This section develops the discrete formulation under the DDPM framework; the corresponding SMLD formulation is provided in Appendix~\ref{sm:sec:smld}. By Bayes' theorem, $p_{i-1}(\vx_{i-1}|\vy)\propto p_{i-1}(\vy|\vx_{i-1})p_{i-1}(\vx_{i-1})$. We therefore first introduce approximations to the prior and measurement likelihood and then derive the resulting Gaussian posterior approximation.

\noindent\textbf{I. Prior Distribution.}
We use the Gaussian DDPM reverse transition as a local approximation to the prior~\cite{song2021scorebased}:
\begin{align}
\label{eq:prior_approx}
\begin{split}
    p_{i-1}(\vx_{i-1})
    &\approx p_{i-1}\big(\vx_{i-1}|\vx_i,\widehat{\vx}_0(i)\big)\\
    &=\mathcal{N}(\widehat{\vmu}_{i-1},\widehat{\sigma}_{i-1}^2\mI).
\end{split}
\end{align}
Here, the denoised estimate $\widehat{\vx}_0(i)=\mathbb{E}[\vx_0|\vx_i]$ is approximated using Tweedie's formula~\cite{efron2011tweedie}:
\begin{align}
\label{eq:tweedie_denoiser}
    \widehat{\vx}_0(i)
    \approx\textstyle\frac{1}{\sqrt{\overline{\alpha}_i}}
    \left(\vx_i+(1-\overline{\alpha}_i)s_{\theta}(\vx_i,i)\right),
\end{align}
where $s_{\theta}(\vx_i,i)$ is the pretrained score network. The local prior mean and variance are given by~\cite{ho2020denoising}
\begin{align}
\label{eq:local_prior_mean}
    \widehat{\vmu}_{i-1}
    &=\textstyle\frac{\sqrt{\alpha_i}(1-\overline{\alpha}_{i-1})}
    {1-\overline{\alpha}_i}\vx_i
    +\textstyle\frac{\sqrt{\overline{\alpha}_{i-1}}\beta_i}
    {1-\overline{\alpha}_i}\widehat{\vx}_0(i),\\
\label{eq:local_prior_covariance}
    \widehat{\sigma}_{i-1}^2
    &=\textstyle\frac{1-\overline{\alpha}_{i-1}}
    {1-\overline{\alpha}_i}\beta_i.
\end{align}

\noindent\textbf{II. Measurement Likelihood.}
\label{para:Measurement_Likelihood}
As shown in \eqref{eq:measurement_likelihood}, evaluating the measurement likelihood $p_{i-1}(\vy|\vx_{i-1})$ requires marginalization over the generally intractable denoising posterior $p(\vx_0|\vx_{i-1})$. Motivated by previous work on Gaussian approximations~\cite{meng2022diffusion,song2023pseudoinverse,boys2023tweedie,peng2024improving,yismaw2005gaussian}, we introduce
\begin{align}
\label{eq:p_x0_x_im1}
    p(\vx_0|\vx_{i-1})
    \approx q(\vx_0|\vx_{i-1})\!=\!\mathcal{N}\Big(
    \textstyle\frac{\vx_{i-1}}{\sqrt{\overline{\alpha}_{i-1}}},
    \widetilde{\sigma}_{i-1}^2\mI
    \Big),
\end{align}
where $\widetilde{\sigma}_{i-1}^2$ controls the variance of the approximation.

Our objective is to preserve a linear dependence on $\vx_{i-1}$ in the approximate measurement model, allowing the subsequent Gaussian posterior approximation to be derived in closed form. We therefore use the linear mean $\vx_{i-1}/\sqrt{\overline{\alpha}_{i-1}}$ in place of the denoised estimates produced by pretrained networks in other methods~\cite{song2023pseudoinverse,boys2023tweedie,peng2024improving,yismaw2005gaussian}. Meng and Kabashima~\cite{meng2022diffusion} use the same mean but derive the variance under a flat data prior. Here, we investigate three choices for $\widetilde{\sigma}_{i-1}^2$, with detailed derivations provided in Appendix~\ref{sm:sec:sigma_tilde_derivations}.

1) \textit{Dirac approximation.}
We use a Dirac distribution centered at $\vx_{i-1}/\sqrt{\overline{\alpha}_{i-1}}$, resulting in $\widetilde{\sigma}_{i-1}^2=0$.

2) \textit{Averaging under the local Gaussian prior.}
For each fixed $\vx_{i-1}$, we derive a conditional variance estimate $\widetilde{\sigma}_{i-1}^2(\vx_{i-1})$ using a criterion based on the KL divergence between $p(\vx_0|\vx_{i-1})$ and $q(\vx_0|\vx_{i-1})$. We then average this estimate under the local Gaussian prior $p_{i-1}(\vx_{i-1}|\vx_i,\widehat{\vx}_0(i))$ in \eqref{eq:prior_approx}, yielding $\widetilde{\sigma}_{i-1}^2=\mathbb{E}_{\vx_{i-1}}\left[\widetilde{\sigma}_{i-1}^2(\vx_{i-1})\right]$.

3) \textit{Averaging under the exact marginal prior.}
Alternatively, we average the conditional variance criterion under the exact marginal distribution $p_{i-1}(\vx_{i-1})$. Although this marginal is not explicitly available, the expectation can be evaluated using the forward diffusion relation in \eqref{eq:forward_diffusion_xi_x0}, giving $\widetilde{\sigma}_{i-1}^2=(1-\overline{\alpha}_{i-1})/\overline{\alpha}_{i-1}$. This expression follows from the forward diffusion model without requiring a particular form for the data prior $p(\vx_0)$. It coincides with the variance derived by~\cite{meng2022diffusion} under an uninformative data prior.

\noindent\textbf{Remark.} These three choices represent different approximation strategies. The first prioritizes simplicity, whereas the other two are motivated by KL minimization and use different distributions to average the conditional variance estimate. As shown later in Section~\ref{subsec:ablation_study}, the first strategy achieves the best performance among these three choices in our experiments, despite its simplicity.

Evaluating the measurement likelihood also requires the measurement noise variance $\sigma_\epsilon^2$. We estimate it from the residual associated with Tweedie's denoised signal:
\begin{align}
\label{eq:measurement_noise_variance}
    \widehat{\sigma}_\epsilon^2
    =\textstyle\frac{1}{M}\|\vy-\mA\widehat{\vx}_0(i)\|_2^2.
\end{align}
Combining this estimate with \eqref{eq:p_x0_x_im1},\eqref{eq:measurement_likelihood} yields the Gaussian likelihood approximation
\begin{align}
    p_{i-1}(\vy|\vx_{i-1})
    \!\approx\!\mathcal{N}\Big(\!
    \mA\textstyle\frac{\vx_{i-1}}{\sqrt{\overline{\alpha}_{i-1}}},
    \widetilde{\sigma}_{i-1}^2\mA\mA^T
    +\widehat{\sigma}_\epsilon^2\mI\!\Big).
\end{align}

\noindent\textbf{III. Posterior Distribution.}
Combining the local Gaussian prior with the approximate measurement likelihood yields a Gaussian posterior approximation at step $i-1$:
\begin{align}
\label{eq:posterior}
    p_{i-1}(\vx_{i-1}|\vy)
    \approx\mathcal{N}(\overline{\vmu}_{i-1},\overline{\mSigma}_{i-1}),
\end{align}
whose mean and covariance are given by
\begin{align}
\label{eq:posterior_mean}
    \overline{\vmu}_{i-1}
    &=\widehat{\vmu}_{i-1}
    \!+\!\widehat{\sigma}_{i-1}^2\mA^T\mB_{i-1}^{-1}\!
    \left(\!
    \sqrt{\overline{\alpha}_{i-1}}\vy
    \!-\!\mA\widehat{\vmu}_{i-1}
    \!\right)\\
    \mB_{i-1}
    &=\left(
    \widehat{\sigma}_{i-1}^2
    +\overline{\alpha}_{i-1}\widetilde{\sigma}_{i-1}^2
    \right)\mA\mA^T
    +\overline{\alpha}_{i-1}\widehat{\sigma}_\epsilon^2\mI\\
\label{eq:posterior_variance}
    \overline{\mSigma}_{i-1}
    &=\widehat{\sigma}_{i-1}^2\mI
    -\widehat{\sigma}_{i-1}^4
    \mA^T\mB_{i-1}^{-1}\mA.
\end{align}
Detailed derivations are provided in Appendix~\ref{sm:sec:derivation}. Although this Gaussian approximation can be sampled directly, the resulting samples may deviate from the noisy image manifold associated with the diffusion model. The ablation study in Section~\ref{subsec:ablation_study} shows that direct sampling performs worse than the proposed guided update. We therefore use $\overline{\vmu}_{i-1}$, the mean and mode of the Gaussian approximation, to define the guidance direction in PPS.

\subsection{Manifold Projection}

\begin{figure}[t]
    \centering
    \includegraphics[width=0.87\linewidth]{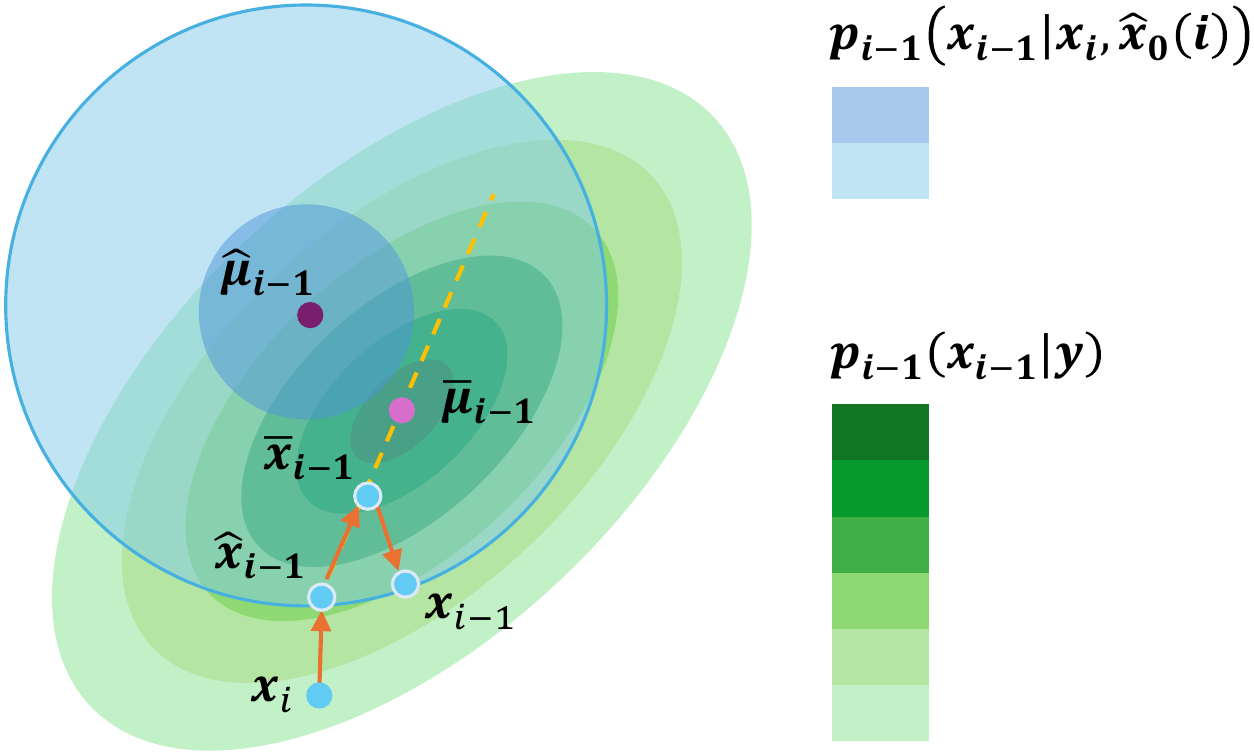}
    \caption{\textbf{PPS with manifold projection.} Blue and green contours depict the local Gaussian prior and posterior approximation, respectively. An unconditional transition from $\vx_i$ produces a sample $\widehat{\vx}_{i-1}$ from the local Gaussian prior. The sample is moved toward the mean $\overline{\vmu}_{i-1}$ of the Gaussian posterior approximation, yielding $\overline{\vx}_{i-1}$, and then projected radially onto the sphere centered at $\widehat{\vmu}_{i-1}$ with radius $\|\widehat{\vx}_{i-1}-\widehat{\vmu}_{i-1}\|_2$, producing $\vx_{i-1}$.}
    \vspace{-1em}
    \label{fig:manifold_projection}
\end{figure}

Our progressive posterior sampling (PPS) method consists of two steps. First, we draw a sample $\widehat{\vx}_{i-1}$ from the local Gaussian prior in \eqref{eq:prior_approx}:
\begin{align}
    \widehat{\vx}_{i-1}
    \sim\mathcal{N}\left(
    \widehat{\vmu}_{i-1},
    \widehat{\sigma}_{i-1}^2\mI
    \right).
\end{align}
Second, we guide the sample toward the posterior mean $\overline{\vmu}_{i-1}$, which coincides with the mode of the Gaussian posterior approximation:
\begin{align}
\label{eq:move_high_prob}
    \overline{\vx}_{i-1}
    =
    \widehat{\vx}_{i-1}
    +\rho\left(
    \overline{\vmu}_{i-1}-\widehat{\vx}_{i-1}
    \right),
\end{align}
where $\rho\in[0,1]$ controls the strength of the guidance.

This update may move the sample away from the local noisy manifold of the diffusion model. We shall approximate the local manifold geometry by a spherical level set based on the Gaussian prior $p_{i-1}(\vx_{i-1}|\vx_i,\widehat{\vx}_0)$. We then project the guided sample back onto this sphere to enforce the corresponding geometric constraint. In particular, we have the following Theorem \ref{thm:sa_main} on the local spherical approximation of the exact reverse distribution $p_{i-1}(\vx_{i-1}|\vx_i)$, with detailed proofs given in Appendix~\ref{sm:sec:theorem_proof}. We use a different notation $\{\mathbf X_i\}_{i=0}^I$ to denote samples drawn from forward distributions or ``exact'' reverse diffusion distributions.

\begin{theorem}[]
\label{thm:sa_main}
Consider the DDPM forward process with $\mathbb E\|\mathbf X_0\|_2^2<\infty$ and $0<\beta_j<1$. Fix $i\in\{2,\ldots,I\}$, let $\widehat{\mathbf X}_0(i)=\mathbb E[\mathbf X_0\mid\mathbf X_i]$, and define the spherical level set
\begin{align}
    \mathcal S_{i-1}(\mathbf X_i)
    =
    \{
        \vx:
        \|\vx-\widehat{\vmu}_{i-1}(\mathbf X_i)\|_2
        =
        \widehat{\sigma}_{i-1}\sqrt N
    \}.
\end{align}
Draw $\mathbf X_i\sim p_i(\cdot)$, where $p_i$ is the forward marginal, and an exact reverse sample $\mathbf X_{i-1}\sim p_{i-1}(\cdot\mid\mathbf X_i)$. Then
\begin{align}
\label{eq:sa_main_bound}
    \frac{
        \left(
            \mathbb E\!\left[
                \dist\bigl(
                    \mathbf X_{i-1},
                    \mathcal S_{i-1}(\mathbf X_i)
                \bigr)^2
            \right]
        \right)^{1/2}
    }{
        \widehat{\sigma}_{i-1}\sqrt N
    }
    \le
    \sqrt{\frac{2}{N}}
    +
    \sqrt{
        \frac{\beta_i}
        {\alpha_i(1-\overline{\alpha}_{i-1})}
    }.
\end{align}
\end{theorem}
\vspace{-0.2em}

Under an exact conditional-mean denoiser given by Tweedie's formula, Theorem~\ref{thm:sa_main} bounds the averaged distance from exact reverse-diffusion samples $\mathbf X_{i-1}$ to spheres centered at their corresponding conditional reverse means $\widehat{\vmu}_{i-1}$. Motivated by this result, we approximate the local geometry of the noisy image manifold by a sphere centered at $\widehat{\vmu}_{i-1}$. In practice, we determine its radius from the sample $\widehat{\vx}_{i-1}$ drawn from the local prior: $r_{i-1}=\|\widehat{\vx}_{i-1}-\widehat{\vmu}_{i-1}\|_2$.
As shown in \cref{fig:manifold_projection}, the radial projection of the guided sample $\overline{\vx}_{i-1}$ produces the denoised signal $\vx_{i-1}$:
\begin{align}
\label{eq:manifold_projection}
    \vx_{i-1}=\widehat{\vmu}_{i-1}+r_{i-1}\frac{\overline{\vx}_{i-1}-\widehat{\vmu}_{i-1}}{\|\overline{\vx}_{i-1}-\widehat{\vmu}_{i-1}\|_2}
\end{align}

The PPS approach is summarized in Algorithm~\ref{alg:pps}.

\begin{algorithm}[tbp]
\caption{PPS with Manifold Projection}
\label{alg:pps}
    \begin{algorithmic}[1]
    \Require $M,N,\vy,\mA,s_{\theta},\rho\in[0,1]$, and $\{\beta_i,\alpha_i,\overline{\alpha}_i\}_{i=1}^I$
    \State Set $\overline{\alpha}_0=1$ and draw $\vx_I\sim\mathcal{N}(\vzero,\mI)$.
    \For{$i=I,I-1,\ldots,1$}
        \State Compute the denoised estimate $\widehat{\vx}_0(i)$ using \eqref{eq:tweedie_denoiser}.
        \State Compute the local prior mean $\widehat{\vmu}_{i-1}$ and variance $\widehat{\sigma}_{i-1}^2$ using \eqref{eq:local_prior_mean} and \eqref{eq:local_prior_covariance}.
        \State Draw $\widehat{\vx}_{i-1}\sim\mathcal{N}(\widehat{\vmu}_{i-1},\widehat{\sigma}_{i-1}^2\mI)$.
        \State Set $r_{i-1}=\|\widehat{\vx}_{i-1}-\widehat{\vmu}_{i-1}\|_2$.
        \State Estimate the measurement noise variance $\widehat{\sigma}_{\epsilon}^2$ \eqref{eq:measurement_noise_variance}.
        \State Compute the posterior mean $\overline{\vmu}_{i-1}$ using \eqref{eq:posterior_mean}.
        \State Compute the guided sample $\overline{\vx}_{i-1}$ using \eqref{eq:move_high_prob}.
        \State Obtain $\vx_{i-1}$ by applying the projection in \eqref{eq:manifold_projection}.
    \EndFor
    \State \Return $\vx_0$.
    \end{algorithmic}
\end{algorithm}

\section{Experiments}
\label{sec:Experiments}

\subsection{Experimental Setup}
\label{subsec:Experimental setup}

We evaluate PPS on four linear inverse problems at three measurement noise levels and compare it with eight baselines for zero-shot image restoration, including DPS~\cite{chung2023diffusion}, DMPS~\cite{meng2022diffusion}, MPGD~\cite{he2024manifold}, DAPS~\cite{zhang2025improving}, CoDPS~\cite{yismaw2005gaussian}, the method of Peng \etal~\cite{peng2024improving}, DDRM~\cite{kawar2022denoising}, and DDNM~\cite{wang2023zeroshot}. We use the pretrained unconditional diffusion models released by Chung \etal~\cite{chung2023diffusion} for FFHQ and by Dhariwal and Nichol~\cite{dhariwal2021diffusion} for ImageNet. All methods use the same diffusion checkpoint. Like the four methods (CoDPS, MPGD, DDRM and DDNM), our PPS also adopts a DDIM-based sampling scheme~\cite{song2021ddim}. Specifically, we use stochastic DDIM sampling with 100 reverse steps and $\eta=0.85$ as recommended, requiring 100 function evaluations (NFEs) of the diffusion network per sample. 

\textbf{Datasets and Metrics.}
We evaluate on FFHQ~\cite{karras2019style} and ImageNet~\cite{deng2009imagenet,russakovsky2015imagenet} at a resolution of $256\times256$. For each dataset, we use 100 test images and a disjoint set of 100 images for hyperparameter tuning. The FFHQ test and tuning sets contain images indexed 60000--60099 and 61000--61099, respectively. For ImageNet, the test set contains the first 100 images of the official validation split(\texttt{ILSVRC2012\_val}); the tuning set contains images indexed 1000--1099 from the same split. ImageNet images are resized and center-cropped to $256\times256$, and all images pixel values are normalized to $[0,1]$. 

We report peak signal-to-noise ratio (PSNR) and structural similarity (SSIM) to assess pixel fidelity and structural agreement, respectively. Because these metrics do not fully capture perceptual quality, we also report Learned Perceptual Image Patch Similarity (LPIPS)~\cite{zhang2018unreasonable}, which measures differences in learned feature representations. LPIPS is evaluated using the VGG-16~\cite{simonyan2015very} backbone.

\textbf{Measurement Operators.}
We consider the following four restoration tasks: (i) random inpainting, with 30\%--70\% of pixels masked; (ii) Gaussian deblurring (GD), with a $61\times61$ kernel and standard
deviation 3.0; (iii) motion deblurring (MD), with randomly generated $61\times61$ kernels and intensity parameter 0.5, using the motion-blur kernel generator of Levi Borodenko;\footnote{\url{https://github.com/LeviBorodenko/motionblur}} and (iv) $4\times$ super-resolution(SR) using bicubic downsampling. For each task, we add independent Gaussian measurement noise with $\sigma_\epsilon\in\{0.05,0.15,0.3\}$. We set the random seed to 1234 for all experiments. Further details of the operators and their implementation are provided in Appendix~\ref{sm:sec:Experimental_Details}.

\textbf{Baselines.}
The aforementioned eight baselines cover several strategies for incorporating measurements into diffusion sampling. Implementation details, sampling schedules, and hyperparameter settings are provided in Appendix~\ref{sm:sec:Experimental_Details}. The comparison includes the method--task combinations reported in \Cref{tab:main_results}. We omit the SVD-based implementations of DDRM, DDNM, and DMPS from the MD comparison because an efficient SVD decomposition of the measurement operator is unavailable in their original experimental setup. For the remaining tasks, we provide DDRM and DDNM with the ground-truth measurement noise variance required by their sampling procedures.


\begin{table*}[t]
\centering
\vspace{-1em}
\setlength{\tabcolsep}{3pt}
\resizebox{\textwidth}{!}{%
\begin{tabular}{clccccccccc|ccccccccc}
\toprule
\multirow{3}{*}{Task} & \multirow{3}{*}{Method} & \multicolumn{9}{c|}{FFHQ} & \multicolumn{9}{c}{ImageNet} \\
\cmidrule(lr){3-11}\cmidrule(lr){12-20}
 &  & \multicolumn{3}{c}{$\sigma_\epsilon=0.05$} & \multicolumn{3}{c}{$\sigma_\epsilon=0.15$} & \multicolumn{3}{c|}{$\sigma_\epsilon=0.3$} & \multicolumn{3}{c}{$\sigma_\epsilon=0.05$} & \multicolumn{3}{c}{$\sigma_\epsilon=0.15$} & \multicolumn{3}{c}{$\sigma_\epsilon=0.3$} \\
\cmidrule(lr){3-5}\cmidrule(lr){6-8}\cmidrule(lr){9-11}\cmidrule(lr){12-14}\cmidrule(lr){15-17}\cmidrule(lr){18-20}
 &  & PSNR$\uparrow$ & SSIM$\uparrow$ & LPIPS$\downarrow$ & PSNR$\uparrow$ & SSIM$\uparrow$ & LPIPS$\downarrow$ & PSNR$\uparrow$ & SSIM$\uparrow$ & LPIPS$\downarrow$ & PSNR$\uparrow$ & SSIM$\uparrow$ & LPIPS$\downarrow$ & PSNR$\uparrow$ & SSIM$\uparrow$ & LPIPS$\downarrow$ & PSNR$\uparrow$ & SSIM$\uparrow$ & LPIPS$\downarrow$ \\
\midrule
\multirow{6}{*}{\begin{tabular}[c]{@{}c@{}}Motion\\deblurring\end{tabular}} & DPS & 24.29 & 0.677 & 0.299 & 24.66 & 0.688 & 0.289 & 23.54 & 0.652 & 0.315 & 22.63 & 0.558 & 0.435 & 21.98 & 0.529 & 0.445 & 21.40 & \underline{0.507} & 0.461 \\
 & MPGD & 26.30 & 0.716 & 0.278 & 23.45 & 0.632 & 0.324 & 22.02 & 0.570 & 0.368 & 22.43 & 0.532 & 0.393 & 20.52 & 0.425 & 0.466 & 19.04 & 0.349 & 0.520 \\
 & DAPS & \underline{28.88} & 0.753 & 0.244 & \underline{27.15} & \underline{0.753} & \underline{0.273} & 15.19 & 0.117 & 0.781 & \underline{26.79} & \underline{0.716} & \underline{0.324} & \underline{25.06} & \underline{0.649} & \underline{0.393} & 12.22 & 0.065 & 0.771 \\
 & CoDPS & 27.27 & 0.749 & 0.256 & 25.05 & 0.688 & 0.310 & 23.33 & 0.635 & 0.357 & 24.06 & 0.535 & 0.419 & 22.39 & 0.489 & 0.471 & 20.74 & 0.446 & 0.522 \\
 & Peng \etal & 28.41 & \underline{0.788} & \underline{0.221} & 25.88 & 0.710 & 0.278 & \underline{24.40} & \underline{0.658} & \underline{0.311} & 26.22 & 0.663 & 0.343 & 23.84 & 0.554 & 0.419 & \underline{22.42} & 0.492 & \underline{0.453} \\
\cmidrule{2-20}
 & \textbf{Ours} & \textbf{30.06} & \textbf{0.847} & \textbf{0.193} & \textbf{27.27} & \textbf{0.774} & \textbf{0.250} & \textbf{25.36} & \textbf{0.710} & \textbf{0.280} & \textbf{27.98} & \textbf{0.777} & \textbf{0.251} & \textbf{25.37} & \textbf{0.677} & \textbf{0.356} & \textbf{23.63} & \textbf{0.580} & \textbf{0.412} \\
\midrule
\multirow{8}{*}{\begin{tabular}[c]{@{}c@{}}Inpainting\end{tabular}} & DPS & 29.92 & 0.838 & 0.214 & 29.30 & 0.824 & 0.214 & 27.18 & 0.768 & 0.256 & 27.66 & 0.743 & 0.325 & 27.57 & 0.753 & 0.303 & 25.66 & 0.689 & 0.357 \\
 & MPGD & 28.59 & 0.794 & 0.209 & 25.61 & 0.628 & 0.313 & 22.29 & 0.606 & 0.366 & 22.96 & 0.597 & 0.371 & 21.23 & 0.474 & 0.448 & 19.37 & 0.365 & 0.524 \\
 & DAPS & 31.83 & 0.850 & 0.167 & 28.97 & 0.729 & 0.244 & 20.29 & 0.289 & 0.554 & 30.00 & 0.822 & 0.185 & 25.10 & 0.556 & 0.346 & 17.91 & 0.231 & 0.577 \\
 & CoDPS & 30.55 & 0.861 & 0.155 & 27.46 & 0.736 & 0.259 & 23.36 & 0.580 & 0.385 & 25.19 & 0.632 & 0.322 & 19.65 & 0.299 & 0.520 & 16.87 & 0.184 & 0.644 \\
 & Peng \etal & \underline{32.78} & \underline{0.905} & \underline{0.120} & 30.16 & 0.840 & \underline{0.183} & \underline{27.70} & 0.769 & \underline{0.241} & \underline{30.73} & \underline{0.866} & \underline{0.157} & \underline{28.43} & 0.780 & \underline{0.241} & \underline{26.07} & 0.676 & \underline{0.324} \\
 & DDRM & 29.59 & 0.848 & 0.207 & 29.31 & 0.840 & 0.211 & 27.69 & \underline{0.798} & 0.260 & 27.33 & 0.773 & 0.282 & 27.13 & 0.765 & 0.286 & 25.89 & \underline{0.713} & 0.329 \\
 & DDNM & 32.64 & \underline{0.905} & 0.137 & \textbf{30.78} & \textbf{0.867} & 0.198 & \textbf{28.92} & \textbf{0.825} & 0.246 & 30.26 & 0.859 & 0.171 & \textbf{28.93} & \textbf{0.817} & \textbf{0.230} & \textbf{27.39} & \textbf{0.764} & \textbf{0.303} \\
\cmidrule{2-20}
 & \textbf{Ours} & \textbf{34.46} & \textbf{0.938} & \textbf{0.111} & \underline{30.63} & \underline{0.853} & \textbf{0.176} & 27.44 & 0.785 & \textbf{0.239} & \textbf{31.43} & \textbf{0.892} & \textbf{0.155} & 27.79 & \underline{0.789} & 0.264 & 25.33 & 0.661 & 0.366 \\
\midrule
\multirow{9}{*}{\begin{tabular}[c]{@{}c@{}}Super-\\resolution\end{tabular}} & DPS & 24.24 & 0.675 & 0.313 & 25.46 & 0.709 & \textbf{0.280} & 23.73 & 0.657 & \textbf{0.318} & 22.55 & 0.554 & 0.465 & 23.83 & 0.606 & \textbf{0.393} & 22.27 & 0.545 & \textbf{0.435} \\
 & DMPS & 28.03 & 0.792 & \textbf{0.212} & 25.42 & 0.713 & 0.286 & 24.24 & 0.659 & 0.325 & 26.58 & 0.717 & \underline{0.331} & 23.94 & 0.619 & 0.443 & 22.74 & 0.571 & 0.477 \\
 & MPGD & 25.42 & 0.712 & 0.274 & 23.16 & 0.627 & 0.345 & 21.06 & 0.543 & 0.411 & 22.65 & 0.544 & 0.391 & 20.57 & 0.436 & 0.480 & 18.39 & 0.333 & 0.556 \\
 & DAPS & 28.28 & 0.750 & 0.275 & 26.54 & 0.740 & 0.299 & 16.78 & 0.220 & 0.731 & 26.12 & 0.686 & 0.369 & 24.64 & 0.633 & 0.426 & 15.48 & 0.189 & 0.718 \\
 & CoDPS & 27.84 & 0.775 & 0.270 & 24.37 & 0.674 & 0.343 & 21.43 & 0.543 & 0.469 & 25.45 & 0.655 & 0.370 & 21.64 & 0.486 & 0.502 & 18.83 & 0.307 & 0.635 \\
 & Peng \etal & 27.34 & 0.761 & 0.240 & 25.29 & 0.692 & 0.294 & 23.50 & 0.630 & 0.336 & 25.06 & 0.611 & 0.375 & 23.30 & 0.525 & 0.434 & 21.63 & 0.453 & 0.474 \\
 & DDRM & 28.72 & 0.816 & 0.245 & \underline{26.95} & \underline{0.770} & 0.292 & \underline{25.24} & \underline{0.727} & 0.331 & \underline{26.75} & \underline{0.730} & 0.332 & \underline{25.26} & \underline{0.675} & \underline{0.395} & \underline{23.79} & \underline{0.625} & 0.453 \\
 & DDNM & \textbf{29.12} & \textbf{0.824} & \underline{0.238} & \textbf{27.19} & \textbf{0.776} & 0.286 & \textbf{25.32} & \textbf{0.730} & 0.324 & \textbf{27.08} & \textbf{0.741} & \textbf{0.328} & \textbf{25.52} & \textbf{0.686} & 0.397 & \textbf{23.92} & \textbf{0.634} & \underline{0.452} \\
\cmidrule{2-20}
 & \textbf{Ours} & \underline{28.78} & \underline{0.818} & 0.241 & 26.64 & 0.758 & \underline{0.284} & 24.71 & 0.701 & \underline{0.321} & 26.65 & 0.721 & 0.344 & 24.87 & 0.647 & 0.401 & 23.23 & 0.595 & 0.453 \\
\midrule
\multirow{9}{*}{\begin{tabular}[c]{@{}c@{}}Gaussian\\deblurring\end{tabular}} & DPS & 25.39 & 0.701 & 0.269 & 25.57 & 0.706 & \underline{0.263} & 24.74 & 0.681 & \underline{0.285} & 23.47 & 0.580 & 0.390 & 22.57 & 0.542 & 0.422 & 22.23 & 0.528 & 0.430 \\
 & DMPS & 27.96 & 0.784 & \textbf{0.204} & 26.56 & 0.742 & \textbf{0.255} & 25.68 & 0.714 & \textbf{0.272} & 26.53 & 0.707 & \textbf{0.324} & 25.11 & 0.652 & 0.410 & 24.44 & 0.625 & 0.420 \\
 & MPGD & 26.41 & 0.720 & 0.258 & 24.55 & 0.655 & 0.303 & 22.88 & 0.613 & 0.330 & 22.74 & 0.527 & 0.374 & 21.51 & 0.457 & 0.434 & 20.53 & 0.444 & 0.452 \\
 & DAPS & 28.45 & 0.753 & 0.267 & 27.54 & 0.759 & 0.281 & 20.24 & 0.343 & 0.655 & 26.06 & 0.656 & 0.369 & 25.18 & 0.640 & 0.416 & 17.01 & 0.223 & 0.684 \\
 & CoDPS & 26.95 & 0.745 & 0.284 & 26.29 & 0.729 & 0.286 & 25.02 & 0.686 & 0.323 & 24.42 & 0.560 & 0.416 & 23.79 & 0.555 & 0.444 & 22.71 & 0.512 & 0.488 \\
 & Peng \etal & 27.53 & 0.762 & \underline{0.222} & 26.11 & 0.713 & 0.264 & 25.11 & 0.676 & 0.288 & 25.24 & 0.618 & 0.352 & 23.84 & 0.544 & 0.405 & 22.97 & 0.504 & 0.424 \\
 & DDRM & \underline{28.91} & \underline{0.818} & 0.246 & \underline{27.74} & \underline{0.786} & 0.275 & \underline{26.71} & \underline{0.759} & 0.299 & \underline{26.98} & \textbf{0.734} & \underline{0.330} & \underline{25.98} & \textbf{0.692} & \underline{0.377} & \underline{25.08} & \textbf{0.659} & \underline{0.416} \\
 & DDNM & \textbf{29.51} & \textbf{0.830} & 0.264 & \textbf{28.14} & \textbf{0.795} & 0.296 & \textbf{26.97} & \textbf{0.766} & 0.320 & \textbf{27.00} & \underline{0.731} & 0.349 & \textbf{25.99} & \underline{0.690} & 0.396 & \textbf{25.16} & \underline{0.658} & 0.432 \\
\cmidrule{2-20}
 & \textbf{Ours} & 28.79 & 0.815 & 0.239 & 27.39 & 0.774 & \underline{0.263} & 26.24 & 0.740 & \underline{0.285} & 26.93 & 0.729 & 0.337 & 25.62 & 0.672 & \textbf{0.362} & 24.70 & 0.639 & \textbf{0.403} \\
\bottomrule
\end{tabular}}
\vspace{-1em}
\caption{Quantitative comparison on both datasets under $\sigma_\epsilon\in\{0.05, 0.15, 0.3\}$. \textbf{Bold}: best; \underline{underline}: second best.}
\label{tab:main_results}
\vspace{-1em}
\end{table*}

\subsection{Results}
\label{subsec:Results}

\Cref{tab:main_results} reports quantitative results on FFHQ and ImageNet at three measurement noise levels. \cref{fig:qualitative} presents qualitative comparisons at the highest noise level, $\sigma_\epsilon=0.3$. Additional results are provided in Appendix~\ref{sm:sec:Additional_Results}. We report PPS results using $\widetilde{\sigma}_{i-1}^2=0$, which performs best among the evaluated variance choices. The ablation study in \cref{subsec:ablation_study} provides a detailed comparison of these choices.

\textbf{Motion deblurring.}
PPS achieves the highest PSNR and SSIM and the lowest LPIPS among the methods on both datasets at every tested noise level. At $\sigma_\epsilon=0.05$, PPS improves PSNR over DPS by 5.77~dB on FFHQ and 5.35~dB on ImageNet. Relative to DAPS, which has the second-highest PSNR in these two settings, the gains are 1.18~dB and 1.19~dB, respectively. These improvements in reconstruction fidelity are accompanied by better perceptual similarity: PPS achieves LPIPS scores of 0.193 on FFHQ and 0.251 on ImageNet, compared with the next-best scores of 0.221 from Peng \etal and 0.324 from DAPS, respectively.

At $\sigma_\epsilon=0.3$, PPS achieves 25.36~dB PSNR and 0.280 LPIPS on FFHQ, compared with 15.19~dB and 0.781 for DAPS. On ImageNet, PPS achieves 23.63~dB PSNR and 0.412 LPIPS at the same noise level.

\textbf{Inpainting.}
At $\sigma_\epsilon=0.05$, PPS achieves the best PSNR, SSIM, and LPIPS on both datasets. It reaches 34.46~dB PSNR on FFHQ and 31.43~dB on ImageNet, exceeding the next-best PSNR results from Peng \etal by 1.68~dB and 0.70~dB, respectively. At the two higher noise levels, PPS continues to achieve the lowest LPIPS on FFHQ, although DDNM obtains higher PSNR and SSIM. On ImageNet, DDNM also achieves lower LPIPS at these higher noise levels, indicating that the relative advantages of PPS depend on the task and measurement noise level.

\textbf{Super resolution and Gaussian deblurring.}
DDRM and DDNM achieve higher PSNR in most of these comparisons, while PPS remains competitive and often achieves lower LPIPS at the two higher noise levels. At $\sigma_\epsilon=0.3$, the PSNR of PPS is within 0.73~dB of the higher score achieved by DDRM and DDNM in each of the SR and GD settings. Across the three tasks other than MD, PPS obtains lower LPIPS than both DDRM and DDNM in four of the six dataset--task combinations at this noise level. These results show that the relative ranking depends on whether reconstruction fidelity or perceptual similarity is emphasized.

\begin{figure*}[htbp]
    \centering
    \setlength{\imgw}{\dimexpr\textwidth-\labelw-3pt\relax}

    \sbox{\ffhqbox}{\begin{minipage}[b]{\imgw}
        \includegraphics[width=\linewidth]
        {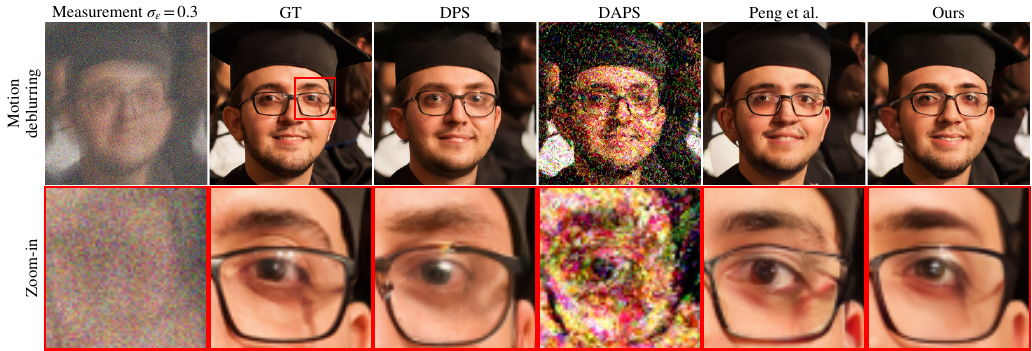}

        \includegraphics[width=\linewidth]
        {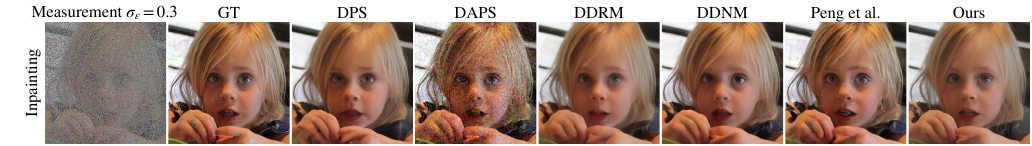}

        \includegraphics[width=\linewidth]
        {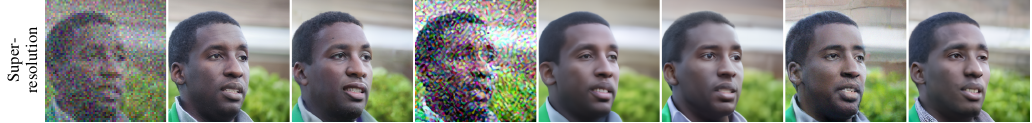}

        \includegraphics[width=\linewidth]
        {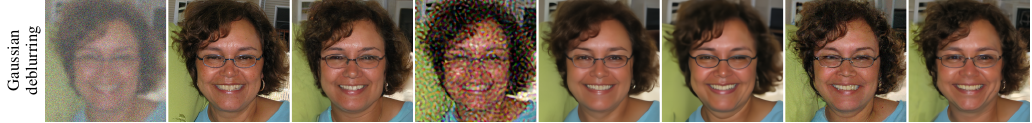}
    \end{minipage}}
    \dsetlabel{FFHQ}{\ffhqbox}\hspace{3pt}\usebox{\ffhqbox}

    \vspace{3pt}

    \sbox{\inetbox}{\begin{minipage}[b]{\imgw}
        \includegraphics[width=\linewidth]
        {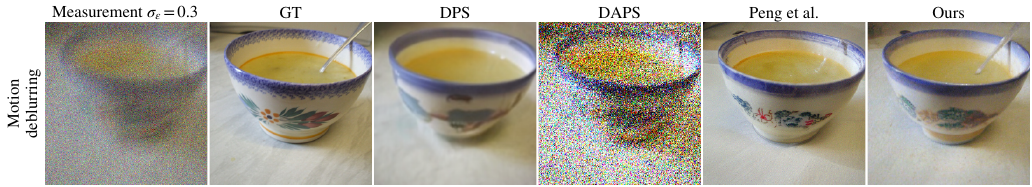}

        \includegraphics[width=\linewidth]
        {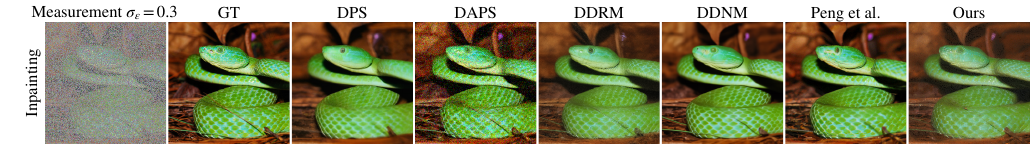}

        \includegraphics[width=\linewidth]
        {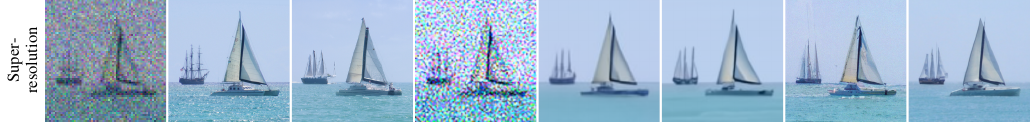}

        \includegraphics[width=\linewidth]
        {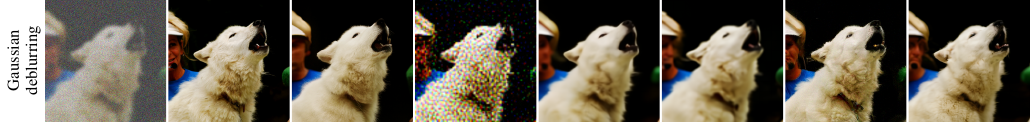}
    \end{minipage}}
    \dsetlabel{ImageNet}{\inetbox}\hspace{3pt}\usebox{\inetbox}

    \caption{Qualitative comparisons on FFHQ and ImageNet under
    additive Gaussian measurement noise with $\sigma_\epsilon=0.3$.
    Within each dataset, the task groups show MD,
    inpainting, $4\times$ SR, and GD,
    from top to bottom.}
    \label{fig:qualitative}
\end{figure*}

\textbf{Qualitative Comparisons.}
The enlarged regions in \cref{fig:qualitative} illustrate differences in fine details. In the FFHQ MD example, PPS preserves the dark eyeglass frame, eyebrow, and iris more faithfully. DPS produces a thinner frame, introduces artifacts near the hinge, and changes the iris color. The method of Peng \etal recovers the overall structure but introduces artifacts along the upper rim of the frame and a reddish streak beneath the eye. In the ImageNet MD example, PPS recovers the metal spoon that is absent from the compared baseline reconstructions and better preserves the colors and arrangement of the floral pattern.

In the inpainting, SR, and GD examples, DDRM and DDNM preserve the overall structure but smooth some fine details, including facial features, sea-surface texture, and animal fur. The method of Peng \etal produces sharper details in some examples but also introduces graininess textures that do not match the GT. For example, it adds structures to the defocused background in the FFHQ SR example and leaves visible background noise in the sailboat example. PPS preserves the main structures while recovering more natural textures in these examples.

\textbf{Sensitivity to Measurement Noise.}
DAPS performs well at the lowest noise level but degrades substantially at $\sigma_\epsilon=0.3$ under the evaluated configurations. Increasing $\sigma_\epsilon$ from 0.15 to 0.3 reduces its PSNR by 7.19--12.84~dB across the eight dataset--task combinations. At $\sigma_\epsilon=0.3$, DAPS has the lowest PSNR in seven of these eight settings, with ImageNet inpainting being the exception. Its average PSNR reduction from
$\sigma_\epsilon=0.05$ to 0.3 is 11.41~dB. The qualitative examples also show colored speckle artifacts at the highest noise level. These observations indicate sensitivity to measurement noise in the evaluated DAPS configurations.

\textbf{Computational Efficiency.}
We compare inference time on GD, which is evaluated for all methods, using the same hardware and software environment. As reported in Appendix~\ref{sm:sec:Additional_Results}, PPS takes 4.2~s per image on FFHQ and 7.2~s on ImageNet. Relative to DPS, DAPS, DMPS, and Peng \etal, the measured speedups of PPS range from $2.3\times$ to $13.1\times$, with larger gains on ImageNet. DDRM and DDNM are faster in this setting, whereas PPS incurs additional cost from the conjugate-gradient iterations used to compute its guidance update. PPS nevertheless combines competitive reconstruction quality with lower inference time than the four baselines listed above.

\subsection{Ablation Studies}
\label{subsec:ablation_study}

We compare two sampling strategies: direct posterior sampling and the proposed guidance update followed by manifold projection. For each strategy, we evaluate the three choices of $\widetilde{\sigma}_{i-1}^2$ introduced in \cref{para:Measurement_Likelihood}. As shown in Appendix~\ref{sm:sec:Additional_Results}, manifold projection improves PSNR over direct posterior sampling by 1.9~dB on FFHQ and 1.3~dB on ImageNet, with corresponding improvements in SSIM and LPIPS as well with any $\widetilde{\sigma}_{i-1}^2$ variant.

Secondly, $\widetilde{\sigma}_{i-1}^2=0$ performs best under both strategies. With direct posterior sampling, it improves PSNR over two other variance choices by 2.00~dB and 2.28~dB on FFHQ, and by 1.79~dB and 2.16~dB on ImageNet, respectively. With manifold projection, the gains are 0.46~dB and 0.54~dB on FFHQ, and 0.68~dB and 0.75~dB on ImageNet. These results support the use of manifold projection with the zero-variance approximation in our method.
\section{Conclusion}
\label{sec:conclusion}
We presented progressive posterior sampling (PPS) with manifold projection for noisy linear inverse problems. PPS combines posterior guidance with a geometric constraint from the diffusion prior, while automatically estimating the measurement noise variance and avoiding SVD of the measurement operator. Experiments show consistent gains in motion deblurring. Although improvements vary across other restoration tasks, PPS remains broadly applicable and competitive, with robust performance across all the evaluated noise levels. Future work could explore adaptive guidance step sizes and local approximations that capture anisotropic geometry to reduce manual tuning and improve reconstruction quality.

\clearpage

\section{LLM Usage Statement}
We used ChatGPT to proofread the text for grammar, wording, and clarity. 

\makeatother
\endgroup

{
    \small
    \bibliographystyle{ieeenat_fullname}
    \bibliography{main}
}


\clearpage


\onecolumn

\setcounter{page}{1}
\setcounter{section}{0}

\begin{center}
\textbf{\Large Denoising Diffusion with Progressive Posterior Sampling}\\
\vspace{1em}
{\Large Supplementary Material}
\vspace{5em}
\end{center}

\tableofcontents

\clearpage

\appendix

\section{Derivation of the variance $\widetilde{\sigma}_{i-1}^2$}
\label{sm:sec:sigma_tilde_derivations}

We derive the second and third choices of $\widetilde{\sigma}_{i-1}^2$ introduced in the main paper. For a fixed diffusion step, we consider the Gaussian family
\begin{align}
    q_v(\vx_0|\vx_{i-1})
    =\mathcal{N}\left(
    \vx_0;\frac{\vx_{i-1}}{\sqrt{\overline{\alpha}_{i-1}}},v\mI
    \right),\qquad v>0,
\end{align}
where $v$ is the variance parameter to be selected. We first characterize the variance $v$ that minimizes the forward KL divergence for a fixed $\vx_{i-1}$, and then consider a variance shared across candidate states at the same diffusion step. The KL arguments below concern nondegenerate conditional densities and assume the indicated divergences and moments are finite.

\begin{proposition}
\label{prop:sigma}
For a fixed $\vx_{i-1}$ and $\overline{\alpha}_{i-1}>0$, suppose that $p(\vx_0|\vx_{i-1})$ admits a density, $D_{\mathrm{KL}}(p(\cdot|\vx_{i-1})\|q_v(\cdot|\vx_{i-1}))$ is finite for some $v>0$, and
\begin{align}
    0<E(\vx_{i-1})
    \mathrel{:=}
    \mathbb{E}_{\vx_0|\vx_{i-1}}
    \left[
    \left\|\vx_0-\frac{\vx_{i-1}}{\sqrt{\overline{\alpha}_{i-1}}}\right\|_2^2
    \right]<\infty.
\end{align}
Then the unique minimizing variance is
\begin{align}
\begin{split}
    \widetilde{\sigma}_{i-1}^2(\vx_{i-1})
    &=
    \arg\min_{v>0}
    D_{\mathrm{KL}}\left(
    p(\cdot|\vx_{i-1})\|q_v(\cdot|\vx_{i-1})
    \right)\\
    &=\frac{1}{N}E(\vx_{i-1}).
\end{split}
\end{align}
\end{proposition}

\begin{proof}
For fixed $\vx_{i-1}$, the dependence of the KL divergence on $v$ is
\begin{align}
\begin{split}
    D_{\mathrm{KL}}(p\|q_v)
    &=\int p\log p\,\sd\vx_0
      -\int p\log q_v\,\sd\vx_0\\
    &=C(\vx_{i-1})+\frac{N}{2}\log v
      +\frac{E(\vx_{i-1})}{2v},
\end{split}
\end{align}
where $C(\vx_{i-1})$ is independent of $v$. Differentiation gives
\begin{align}
    \frac{\partial}{\partial v}D_{\mathrm{KL}}(p\|q_v)
    =\frac{Nv-E(\vx_{i-1})}{2v^2}.
\end{align}
The derivative is negative for $0<v<E(\vx_{i-1})/N$ and positive for $v>E(\vx_{i-1})/N$. Hence $v=E(\vx_{i-1})/N$ is the unique global
minimizer.
\end{proof}

\paragraph{A variance shared across candidate states.}
Let $ h(\cdot)$ be a chosen distribution of $\vx_{i-1}$, and define the averaged objective
\begin{align}
    J_ h(v)
    =
    \mathbb{E}_{\vx_{i-1}\sim h}
    \left[
    D_{\mathrm{KL}}\left(
    p(\cdot|\vx_{i-1})\|q_v(\cdot|\vx_{i-1})
    \right)
    \right].
\end{align}
Assuming this objective is finite for some $v>0$ and the averaged squared error is positive and finite, we obtain
\begin{align}
    J_ h(v)
    =C_ h+\frac{N}{2}\log v
    +\frac{1}{2v}
    \mathbb{E}_{\vx_{i-1}\sim h}[E(\vx_{i-1})].
\end{align}
Therefore,
\begin{align}
\label{sm:eq:shared_variance}
\begin{split}
    \arg\min_{v>0}J_ h(v)
    &=\frac{1}{N}
    \mathbb{E}_{\vx_{i-1}\sim h}[E(\vx_{i-1})]\\
    &=
    \mathbb{E}_{\vx_{i-1}\sim h}
    \left[
    \widetilde{\sigma}_{i-1}^2(\vx_{i-1})
    \right].
\end{split}
\end{align}
Thus, for this Gaussian family, averaging the conditional optimal variances yields the optimal shared variance for the averaged KL objective.

\paragraph{Second choice: local conditional approximation.}
To obtain a variance adapted to the current iterate $\vx_i$, first
consider the exact conditional weighting
$ h(\vx_{i-1})=p(\vx_{i-1}|\vx_i)$ under the forward DDPM model.
The Markov property gives
$p(\vx_0|\vx_{i-1},\vx_i)=p(\vx_0|\vx_{i-1})$.
Consequently, the optimal shared variance for this conditional weighting
can be written as
\begin{align}
\label{sm:eq:conditional_variance_target}
\begin{split}
    \nu_i^\star(\vx_i)
    &\mathrel{:=}
    \mathbb{E}_{\vx_{i-1}|\vx_i}
    \left[
    \widetilde{\sigma}_{i-1}^2(\vx_{i-1})
    \right]\\
    &=
    \frac{1}{N}
    \mathbb{E}_{(\vx_0,\vx_{i-1})|\vx_i}
    \left[
    \left\|\vx_0-\frac{\vx_{i-1}}{\sqrt{\overline{\alpha}_{i-1}}}\right\|_2^2
    \right].
\end{split}
\end{align}
Both expectations in this identity are taken under the same forward
joint distribution conditioned on $\vx_i$.

The conditional joint distribution factors as
\begin{align}
\begin{split}
    p(\vx_0,\vx_{i-1}|\vx_i)
    =p(\vx_0|\vx_i)\,
    p(\vx_{i-1}|\vx_i,\vx_0).
\end{split}
\end{align}
The second factor is the tractable DDPM Gaussian transition distribution conditioned on the clean signal~\cite{ho2020denoising}. To obtain a computable local rule, we approximate the first factor by a point mass at the denoised
estimate:
\begin{align}
    p(\vx_0|\vx_i)
    \approx
    \delta\big(\vx_0-\widehat{\vx}_0(i)\big).
\end{align}
This substitution is applied to the moment expression in \eqref{sm:eq:conditional_variance_target}, rather than to the KL divergence itself. It induces the local Gaussian approximation
\begin{align}
\begin{split}
     h_{i-1}(\vx_{i-1}|\vx_i)
    &\mathrel{:=}
    \left.
    p(\vx_{i-1}|\vx_i,\vx_0)
    \right|_{\vx_0=\widehat{\vx}_0(i)}\\
    &=\mathcal{N}\left(
    \vx_{i-1};
    \widehat{\vmu}_{i-1},
    \widehat{\sigma}_{i-1}^2\mI
    \right),
\end{split}
\end{align}
with the parameters in \eqref{eq:prior_approx}.

We approximate $\nu_i^\star(\vx_i)$ by the following local variance:
\begin{align}
\label{sm:eq:sigma_tilde}
\begin{split}
    \widetilde{\sigma}_{i-1}^2
    &\mathrel{:=}
    \frac{1}{N}
    \mathbb{E}_{\vx_{i-1}\sim h_{i-1}(\cdot|\vx_i)}
    \left[
    \left\|
    \widehat{\vx}_0(i)-\frac{\vx_{i-1}}{\sqrt{\overline{\alpha}_{i-1}}}
    \right\|_2^2
    \right]\\
    &=
    \frac{1}{N}
    \left\|
    \widehat{\vx}_0(i)
    -\frac{\widehat{\vmu}_{i-1}}{\sqrt{\overline{\alpha}_{i-1}}}
    \right\|_2^2
    +\frac{\widehat{\sigma}_{i-1}^2}{\overline{\alpha}_{i-1}}.
\end{split}
\end{align}
Expanding the squared norm gives the expression used in the main paper:
\begin{align}
\begin{split}
    \widetilde{\sigma}_{i-1}^2
    =\frac{1}{N}\Biggl(
    &\|\widehat{\vx}_0(i)\|_2^2
    +\left\|
    \frac{\widehat{\vmu}_{i-1}}{\sqrt{\overline{\alpha}_{i-1}}}
    \right\|_2^2
    +\frac{N\widehat{\sigma}_{i-1}^2}{\overline{\alpha}_{i-1}}-\frac{2\widehat{\vmu}_{i-1}^T
    \widehat{\vx}_0(i)}{\sqrt{\overline{\alpha}_{i-1}}}
    \Biggr).
\end{split}
\end{align}

\paragraph{Third choice: exact marginal averaging.}
We next choose $ h(\vx_{i-1})=p_{i-1}(\vx_{i-1})$, the exact noisy-data marginal under the forward diffusion model. The shared variance is then
\begin{align}
\begin{split}
    \widetilde{\sigma}_{i-1}^2
    &=
    \mathbb{E}_{\vx_{i-1}}
    \left[
    \widetilde{\sigma}_{i-1}^2(\vx_{i-1})
    \right]\\
    &=
    \frac{1}{N}
    \mathbb{E}_{(\vx_0,\vx_{i-1})}
    \left[
    \left\|\vx_0-\frac{\vx_{i-1}}{\sqrt{\overline{\alpha}_{i-1}}}\right\|_2^2
    \right].
\end{split}
\end{align}
Here, the joint expectation is taken under the original forward model. By \eqref{eq:forward_diffusion_xi_x0},
\begin{align}
    \vx_{i-1}
    =\sqrt{\overline{\alpha}_{i-1}}\vx_0+\sqrt{1-\overline{\alpha}_{i-1}}\vz,
    \qquad
    \vz\sim\mathcal{N}(\vzero,\mI),
\end{align}
where $\vz$ is independent of $\vx_0$. Hence
\begin{align}
\label{sm:eq:sigma_tilde_global}
\begin{split}
    \widetilde{\sigma}_{i-1}^2
    &=
    \frac{1-\overline{\alpha}_{i-1}}{\overline{\alpha}_{i-1}N}\,
    \mathbb{E}_{\vz}\big[\|\vz\|_2^2\big]\\
    &=\frac{1-\overline{\alpha}_{i-1}}
    {\overline{\alpha}_{i-1}}.
\end{split}
\end{align}
This moment identity follows from the specified Gaussian forward corruption model and does not require a particular form for the data prior. Under the above finiteness assumptions and $0<\overline{\alpha}_{i-1}<1$, it is also the unique optimal shared variance for the forward KL objective averaged over the exact noisy-data marginal. 

\section{Derivation of the Gaussian Posterior Approximation}
\label{sm:sec:derivation}

At step $i-1$, we combine the local Gaussian prior with the approximate measurement likelihood. Throughout this derivation, the current iterate $\vx_i$ and the estimated variance parameters are treated as fixed. We first define the following:
\begin{align}
    \mR
    \mathrel{:=}
    \widetilde{\sigma}_{i-1}^2\mA\mA^T
    +\widehat{\sigma}_{\epsilon}^2\mI,
\end{align}
where $\widehat{\sigma}_\epsilon^2$ is the estimated measurement noised variance obtained in \eqref{eq:measurement_noise_variance}.

The approximate model is
\begin{align}
    &\textnormal{Local prior approximation:}\quad p_{i-1}(\vx_{i-1})
    \approx p_{i-1}\big(\vx_{i-1}|\vx_i,\widehat{\vx}_0(i)\big)=
    \mathcal{N}\left(
    \vx_{i-1};
    \widehat{\vmu}_{i-1},
    \widehat{\sigma}_{i-1}^2\mI
    \right),\\
    &\textnormal{Likelihood approximation:}\quad p_{i-1}(\vy|\vx_{i-1})
    \approx
    \mathcal{N}\left(
    \vy;\frac{\mA\vx_{i-1}}{\sqrt{\overline{\alpha}_{i-1}}},\mR
    \right).
\end{align}
Assume $\overline{\alpha}_{i-1}>0$, $\widehat{\sigma}_{i-1}^2>0$, and $\mR\succ0$.
In particular, $\widehat{\sigma}_{\epsilon}^2>0$ is sufficient to ensure $\mR\succ0$ when $\widetilde{\sigma}_{i-1}^2\geq0$.

Bayes' theorem defines the normalized posterior of this approximate model as
\begin{align}
\begin{split}
    p_{i-1}(\vx_{i-1}|\vy)
    &=
    \frac{
    p_{i-1}(\vy|\vx_{i-1})p_{i-1}(\vx_{i-1})
    }{
    \int
    p_{i-1}(\vy|\vx)p_{i-1}(\vx)\,\sd\vx
    }\\
    &\approx \mathcal{N}\left(
    \vx_{i-1};
    \overline{\vmu}_{i-1},
    \overline{\mSigma}_{i-1}
    \right),
\end{split}
\end{align}
where dependence on the fixed $\vx_i$ and estimated parameters is suppressed in the notation for $p_{i-1}(\vx_{i-1}|\vy)\approx p_{i-1}(\vx_{i-1}|\vx_i,\vy)$.
We further have the posterior covariance and mean:
\begin{align}
\label{sm:eq:posterior_precision}
    \overline{\mSigma}_{i-1}
    &=
    \left(
    \frac{1}{\widehat{\sigma}_{i-1}^2}\mI
    +\frac{1}{\overline{\alpha}_{i-1}}\mA^T\mR^{-1}\mA
    \right)^{-1},\\
\label{sm:eq:posterior_mean_precision}
    \overline{\vmu}_{i-1}
    &=
    \overline{\mSigma}_{i-1}
    \left[
    \frac{\widehat{\vmu}_{i-1}}
    {\widehat{\sigma}_{i-1}^2}
    +\frac{1}{\sqrt{\overline{\alpha}_{i-1}}}\mA^T\mR^{-1}\vy
    \right].
\end{align}

\paragraph{Covariance.}
Applying the Woodbury matrix identity to
\eqref{sm:eq:posterior_precision} yields
\begin{align}
\label{sm:eq:posterior_covariance_woodbury}
\begin{split}
    \overline{\mSigma}_{i-1}
    &=
    \widehat{\sigma}_{i-1}^2\mI-
    \widehat{\sigma}_{i-1}^4
    \mA^T
    \left(
    \widehat{\sigma}_{i-1}^2\mA\mA^T+\overline{\alpha}_{i-1}\mR
    \right)^{-1}
    \mA\\
    &=
    \widehat{\sigma}_{i-1}^2\mI
    -\widehat{\sigma}_{i-1}^4
    \mA^T\mB_{i-1}^{-1}\mA,
\end{split}
\end{align}
where
\begin{align}
\label{sm:eq:B_definition}
\begin{split}
    \mB_{i-1}
    &\mathrel{:=}
    \widehat{\sigma}_{i-1}^2\mA\mA^T+\overline{\alpha}_{i-1}\mR\\
    &=
    \left(
    \widehat{\sigma}_{i-1}^2
    +\overline{\alpha}_{i-1}\widetilde{\sigma}_{i-1}^2
    \right)\mA\mA^T
    +\overline{\alpha}_{i-1}\widehat{\sigma}_{\epsilon}^2\mI.
\end{split}
\end{align}
The stated assumptions imply $\mB_{i-1}\succ0$.

\paragraph{Mean.}
From \eqref{sm:eq:B_definition}, we have
\begin{align}
\begin{split}
    \mI-
    \widehat{\sigma}_{i-1}^2
    \mB_{i-1}^{-1}\mA\mA^T
    &=
    \mB_{i-1}^{-1}
    \left(
    \mB_{i-1}
    -\widehat{\sigma}_{i-1}^2\mA\mA^T
    \right)\\
    &=\overline{\alpha}_{i-1}\mB_{i-1}^{-1}\mR.
\end{split}
\end{align}
Consequently,
\begin{align}
\begin{split}
    \overline{\mSigma}_{i-1}\mA^T\mR^{-1}
    &=
    \widehat{\sigma}_{i-1}^2\mA^T
    \left(
    \mI-\widehat{\sigma}_{i-1}^2
    \mB_{i-1}^{-1}\mA\mA^T
    \right)\mR^{-1}\\
    &=\overline{\alpha}_{i-1}\widehat{\sigma}_{i-1}^2
    \mA^T\mB_{i-1}^{-1}.
\end{split}
\end{align}
Substitution into \eqref{sm:eq:posterior_mean_precision} gives
\begin{align}
\label{sm:eq:posterior_mean_final}
\begin{split}
    \overline{\vmu}_{i-1}
    &=
    \frac{1}{\widehat{\sigma}_{i-1}^2}
    \overline{\mSigma}_{i-1}\widehat{\vmu}_{i-1}
    +\frac{1}{\sqrt{\overline{\alpha}_{i-1}}}
    \overline{\mSigma}_{i-1}\mA^T\mR^{-1}\vy\\
    &=
    \widehat{\vmu}_{i-1}
    +\widehat{\sigma}_{i-1}^2
    \mA^T\mB_{i-1}^{-1}
    \left(
    \sqrt{\overline{\alpha}_{i-1}}\vy-\mA\widehat{\vmu}_{i-1}
    \right).
\end{split}
\end{align}
These expressions are exact for the local Gaussian model and provide an approximation to the original posterior.

Only the posterior mean $\overline{\mu}_{i-1}$ is required for the PPS update. It can be evaluated by solving
\begin{align}
    \mB_{i-1}\vv =\sqrt{\overline{\alpha}_{i-1}}\vy-\mA\widehat{\vmu}_{i-1}
    \label{eq:smld_linear_solve}
\end{align}
using the conjugate gradient method and setting $\overline{\vmu}_{i-1}=\widehat{\vmu}_{i-1}+\widehat{\sigma}_{i-1}^2\mA^T\vv$. Thus, the formulation does not require an explicit SVD of $\mA$ or the construction of $\overline{\mSigma}_{i-1}$.

\section{Proof of Theorem~\ref{thm:sa_main}}
\label{sm:sec:theorem_proof}

To distinguish samples $\{\vx_i\}_{i=0}^I$ generated by the approximate reverse process from those associated with the exact diffusion process, we use $\{\mathbf X_i\}_{i=0}^I$ to denote random variables following the exact forward or reverse diffusion distributions. In particular, samples $\mathbf X_{i-1}$ drawn from the exact reverse distribution characterize the true local noisy-data distribution. The following theorem, Theorem~\ref{thm:sa_main_app}, establishes an upper bound on the normalized root-mean-square radial distance between these exact reverse samples and the spherical level set $\mathcal S_{i-1}(\mathbf X_i)$, providing theoretical justification for the proposed spherical approximation. We then present Corollary~\ref{cor:ddpm_spherical_stability}, which extends the analysis to spherical approximations constructed using approximate reverse states and a pretrained score function, explicitly accounting for the discrepancy between the exact and approximate sphere centers.

\setcounter{theorem}{0}
\begin{theorem}[]
\label{thm:sa_main_app}
Consider the DDPM forward process with $\mathbb E\|\mathbf X_0\|_2^2<\infty$ and $0<\beta_j<1$. Fix $i\in\{2,\ldots,I\}$, let $\widehat{\mathbf X}_0(i)=\mathbb E[\mathbf X_0\mid\mathbf X_i]$, and define the spherical level set
\begin{align}
    \mathcal S_{i-1}(\mathbf X_i)
    =
    \{\vx: \|\vx-\widehat{\vmu}_{i-1}(\mathbf X_i)\|_2 = \widehat{\sigma}_{i-1}\sqrt N
    \}.
\end{align}
Draw $\mathbf X_i\sim p_i(\cdot)$, where $p_i$ is the forward marginal, and an exact reverse sample $\mathbf X_{i-1}\sim p_{i-1}(\cdot\mid\mathbf X_i)$. Then
\begin{align}
\label{eq:sa_main_bound}
    \frac{
        \left(
            \mathbb E\!\left[
                \dist\bigl(
                    \mathbf X_{i-1},
                    \mathcal S_{i-1}(\mathbf X_i)
                \bigr)^2
            \right]
        \right)^{1/2}
    }{
        \widehat{\sigma}_{i-1}\sqrt N
    }
    \le
    \sqrt{\frac{2}{N}}
    +
    \sqrt{
        \frac{\beta_i}
        {\alpha_i(1-\overline{\alpha}_{i-1})}
    }.
\end{align}
The expectation is taken over both $\mathbf X_i$ and $\mathbf X_{i-1}$. 
\end{theorem}

\begin{proof}
For notational convenience, define
\begin{align}
    a_i&=\frac{\sqrt{\alpha_i}(1-\overline{\alpha}_{i-1})}{1-\overline{\alpha}_i},&\gamma_i&=\frac{\sqrt{\overline{\alpha}_{i-1}}\beta_i}{1-\overline{\alpha}_i}.
\end{align}
The exact DDPM local reverse distribution conditioned on the clean signal $\mathbf X_0$ is
\begin{align}
\label{eq:sa_bridge}
    p(\mathbf X_{i-1}\mid\mathbf X_i,\mathbf X_0)=\mathcal N\!\left(a_i\mathbf X_i+\gamma_i\mathbf X_0,\,\widehat{\sigma}_{i-1}^2\mI\right),
\end{align}
where
\begin{align}
    \widehat{\sigma}_{i-1}^2=\frac{1-\overline{\alpha}_{i-1}}{1-\overline{\alpha}_i}\beta_i.
\end{align}
Since the center of the spherical level set is
\begin{align}
    \widehat{\vmu}_{i-1}(\mathbf X_i)=a_i\mathbf X_i+\gamma_i\widehat{\mathbf X}_0(i),
\end{align}
an exact reverse sample $\mathbf X_{i-1}$ can be represented by
\begin{align}
\label{eq:sa_representation}
    \mathbf X_{i-1}=\widehat{\vmu}_{i-1}(\mathbf X_i)+\gamma_i\bigl(\mathbf X_0-\widehat{\mathbf X}_0(i)\bigr)+\widehat{\sigma}_{i-1}\mathbf Z,
\end{align}
where $\mathbf Z\sim\mathcal N(\mathbf 0,\mI)$ is independent of $(\mathbf X_0,\mathbf X_i)$.

Here $(\mathbf X_0,\mathbf X_i)$ follows its forward joint distribution. Equivalently, one may first draw $\mathbf X_i\sim p_i$, then $\mathbf X_0\sim p(\mathbf X_0\mid\mathbf X_i)$, and finally the independent Gaussian vector $\mathbf Z$. Marginalizing $\mathbf X_0$ in this construction gives the exact reverse distribution $p(\mathbf X_{i-1}\mid\mathbf X_i)$.

The distance from a point to a sphere is the absolute difference between its distance to the center and the sphere's radius. Therefore, the reverse triangle inequality and \eqref{eq:sa_representation} give

\begin{align}
\label{eq:sa_triangle}
\begin{split}
\dist\bigl(
\mathbf X_{i-1},
\mathcal S_{i-1}(\mathbf X_i)
\bigr) &=\left|\|\mathbf X_{i-1}-\widehat{\vmu}_{i-1}(\mathbf X_i)\|_2-\widehat{\sigma}_{i-1}\sqrt N\right|\\
&=
\left|
\left\|
\gamma_i(\mathbf X_0-\widehat{\mathbf X}_0(i))
+\widehat{\sigma}_{i-1}\mathbf Z
\right\|_2
-\widehat{\sigma}_{i-1}\sqrt N
\right|\\
&\overset{(a)}{\le}
\left|
\left\|
\gamma_i(\mathbf X_0-\widehat{\mathbf X}_0(i))
+\widehat{\sigma}_{i-1}\mathbf Z
\right\|_2
-\widehat{\sigma}_{i-1}\|\mathbf Z\|_2
\right|
+
\widehat{\sigma}_{i-1}
\left|\|\mathbf Z\|_2-\sqrt N\right|\\
&\overset{(b)}{\le}
\gamma_i\|\mathbf X_0-\widehat{\mathbf X}_0(i)\|_2
+
\widehat{\sigma}_{i-1}
\left|\|\mathbf Z\|_2-\sqrt N\right|,
\end{split}
\end{align}
where (a) follows from the triangle inequality $|u+v|\le |u|+|v|$, and (b) follows from the reverse triangle inequality $|\|\vx+\mathbf v\|_2-\|\mathbf v\|_2|\le\|\vx\|_2$.

We first bound the Gaussian radius fluctuation. Since $\|\mathbf Z\|_2^2\sim\chi_N^2$ follows the chi-squared distribution and thus has mean $N$ and variance $2N$, we have
\begin{align}
\label{eq:sa_gaussian_radius}
\begin{split}
    \mathbb E\!\left[\left(\|\mathbf Z\|_2-\sqrt N\right)^2\right]
    &=\mathbb E\!\left[\left(\frac{\|\mathbf Z\|_2^2-N}{\|\mathbf Z\|_2+\sqrt N}\right)^2\right]\\
    &= \mathbb E\!\left[\frac{\left(\|\mathbf Z\|_2^2-N\right)^2}{\left(\|\mathbf Z\|_2+\sqrt N\right)^2}\right]\\
    &\le \frac{\mathbb E[\left(\|\mathbf Z\|_2^2-N\right)^2]}{N}\\
    &=2.
\end{split}
\end{align}

Taking root-mean-square norms of \eqref{eq:sa_triangle} and applying Minkowski's inequality yields
\begin{align}
\label{eq:sa_intermediate_bound}
\begin{split}
    &\left(\mathbb E\!\left[\dist\bigl(\mathbf X_{i-1},\,\mathcal S_{i-1}(\mathbf X_i)\bigr)^2\right]\right)^{1/2} \le \sqrt{2}\widehat{\sigma}_{i-1} + \gamma_i\left(\mathbb E\left\|\mathbf X_0-\widehat{\mathbf X}_0(i)\right\|_2^2\right)^{1/2}.
\end{split}
\end{align}

Next, because $\widehat{\mathbf X}_0(i)=\mathbb E[\mathbf X_0\mid\mathbf X_i]$, the conditional mean minimizes the mean-squared estimation error. In particular, for any square-integrable estimator $h(\mathbf X_i)$,
\begin{align}
    \mathbb E\left[\|\mathbf X_0-h(\mathbf X_i)\|_2^2\right]=\mathbb E\left[\left\|\mathbf X_0-\widehat{\mathbf X}_0(i)\right\|_2^2\right] + \mathbb E\left[\left\|\widehat{\mathbf X}_0(i)-h(\mathbf X_i)\right\|_2^2\right].
\end{align}
The cross term vanishes because $\mathbb E\left[\mathbf X_0-\widehat{\mathbf X}_0(i) \mid\mathbf X_i\right]=\mathbf 0$.

Choosing $h(\mathbf X_i)=\mathbf X_i/\sqrt{\overline{\alpha}_i}$ and using the forward representation
\begin{align}
    \mathbf X_i = \sqrt{\overline{\alpha}_i}\mathbf X_0 + \sqrt{1-\overline{\alpha}_i}\vz, \qquad \vz\sim\mathcal N(\mathbf 0,\mI),
\end{align}
where $\vz$ is independent of $\mathbf X_0$, we obtain
\begin{align}
\label{eq:sa_mmse_bound}
\begin{split}
    \mathbb E\left[\left\|\mathbf X_0-\widehat{\mathbf X}_0(i)\right\|_2^2\right] & \le \mathbb E\left[\left\|\mathbf X_0-\frac{\mathbf X_i}{\sqrt{\overline{\alpha}_i}}\right\|_2^2\right]\\
    &=N\frac{1-\overline{\alpha}_i}{\overline{\alpha}_i}.
\end{split}
\end{align}

Substituting \eqref{eq:sa_mmse_bound} into \eqref{eq:sa_intermediate_bound} and dividing by $\widehat{\sigma}_{i-1}\sqrt N>0$ gives
\begin{align}
\begin{split}
    \frac{
        \left(
            \mathbb E\!\left[
                \dist\bigl(
                    \mathbf X_{i-1},
                    \mathcal S_{i-1}(\mathbf X_i)
                \bigr)^2
            \right]
        \right)^{1/2}
    }{
        \widehat{\sigma}_{i-1}\sqrt N
    }
    \le
    \sqrt{\frac{2}{N}}
    +
    \frac{\gamma_i}{\widehat{\sigma}_{i-1}}
    \sqrt{
        \frac{
            1-\overline{\alpha}_i
        }{
            \overline{\alpha}_i
        }
    }.
\end{split}
\end{align}
Finally, using $\overline{\alpha}_i=\alpha_i\overline{\alpha}_{i-1}$,
\begin{align}
\begin{split}
    \frac{\gamma_i^2}{\widehat{\sigma}_{i-1}^2}
    \frac{
        1-\overline{\alpha}_i
    }{
        \overline{\alpha}_i
    }
    &=
    \frac{
        \overline{\alpha}_{i-1}\beta_i
    }{
        (1-\overline{\alpha}_{i-1})
        \overline{\alpha}_i
    }
    \\
    &=
    \frac{
        \beta_i
    }{
        \alpha_i(1-\overline{\alpha}_{i-1})
    }.
\end{split}
\end{align}
This proves \eqref{eq:sa_main_bound}.
\end{proof}


\subsection{Stability of the Spherical Approximation under DDPM}
\label{sec:ddpm_spherical_stability}


Theorem~\ref{thm:sa_main} establishes the spherical approximation under the exact forward marginal and an exact conditional-mean denoiser. To account for practical sampling, we consider reference spheres centered at the local prior mean computed from intermediate PPS states $\vx_i$ and the pretrained score function $s_\theta(\vx_i,i)$, while retaining the reference radius $\widehat{\sigma}_{i-1}\sqrt{N}$. The following corollary bounds the normalized root-mean-square radial distance from exact reverse samples to these reference spheres in terms of the discrepancy between the exact and approximate centers. Since measurement guidance may shift the PPS intermediate-state distribution away from the unconditional diffusion marginal, the guarantee depends on the resulting center discrepancy.

\begin{corollary}[Stability of the spherical approximation]
\label{cor:ddpm_spherical_stability}
Consider the DDPM framework under the assumptions of Theorem~\ref{thm:sa_main}. Let $\mathbf X_i\sim p_i(\mathbf X_i)$ and $\vx_i\sim \widehat{p}_i(\vx_i|\vx_{i+1},\widehat{\vx}_0(i+1))$, where $p_i$ is the exact forward marginal and $\widehat{p}_i$ is the distribution of intermediate states $\vx_i$ generated by the approximate reverse process. Conditional on $\mathbf X_i$, we draw $\mathbf X_{i-1}$ from the exact reverse distribution $p_{i-1}(\cdot\mid\mathbf X_i)$ independently of $\vx_i$.

Define the exact and approximate local reverse means by
\begin{align}
\label{eq:ddpm_exact_reverse_mean}
\widehat{\vmu}_{i-1}(\mathbf X_i)
&=\frac{1}{\sqrt{\alpha_i}}
\left(\mathbf X_i+\beta_i\cdot \nabla\log p_i(\mathbf X_i)\right),\\
\label{eq:ddpm_approx_reverse_mean}
\widehat{\vmu}^\theta_{i-1}(\vx_i)
&=\frac{1}{\sqrt{\alpha_i}}
\left(\vx_i+\beta_i\cdot s_\theta(\vx_i,i)\right),
\end{align}
where $\alpha_i=1-\beta_i$, $\overline{\alpha}_i=\prod_{j=1}^{i}\alpha_j$, and $s_\theta(\cdot)$ is the pretrained score function. 

Define the spherical level set constructed by the approximate reverse process as
\begin{align}
\label{eq:ddpm_practical_sphere}
\mathcal S_{i-1}(\vx_i)
=\left\{\vx:\
\|\vx-\widehat{\vmu}^\theta_{i-1}(\vx_i)\|_2
=\widehat\sigma_{i-1}\sqrt N\right\},
\end{align}
where
\begin{align}
\widehat\sigma_{i-1}^{2}
=\frac{1-\overline{\alpha}_{i-1}}{1-\overline{\alpha}_i}\,\beta_i.
\end{align}
Then, we have
\begin{align}
\label{eq:ddpm_spherical_stability}
\frac{\left(\mathbb E\left[ \operatorname{dist}\bigl(\mathbf X_{i-1}, \mathcal S_{i-1}(\vx_i)\bigr)^2\right]\right)^{1/2}}{\widehat\sigma_{i-1}\sqrt N}
\le\sqrt{\frac{2}{N}}
+\sqrt{\frac{\beta_i}{\alpha_i(1-\overline{\alpha}_{i-1})}}
+\frac{\left(\mathbb E\left[ \|\widehat{\vmu}_{i-1}(\mathbf X_i) -\widehat{\vmu}^\theta_{i-1}(\vx_i)\|_2^2 \right]\right)^{1/2}}{\widehat\sigma_{i-1}\sqrt N}.
\end{align}
The expectations are taken over the joint distribution of $(\mathbf X_{i-1},\mathbf X_i,\vx_i)$.
\end{corollary}

\begin{proof}
Define the ideal sphere
\begin{align}
\mathcal S_{i-1}(\mathbf X_i)
=\left\{\vx:\
\|\vx-\widehat{\vmu}_{i-1}(\mathbf X_i)\|_2
=\widehat\sigma_{i-1}\sqrt N\right\}.
\end{align}

For any point $\vx$ and two spheres with the same radius $r$ but different centers $\va$ and $\vb$, the reverse triangle inequality gives
\begin{align}
\begin{split}
&\dist\bigl(\vx,\mathcal S(\vb,r)\bigr)\\
&=
\left|\left\|\vx-\vb\right\|_2-r\right|\\
&=
\left|
\bigl(\|\vx-\vb\|_2
-\|\vx-\va\|_2\bigr)
+
\bigl(\|\vx-\va\|_2-r\bigr)
\right|\\
&\le
\left|
\|\vx-\vb\|_2
-\|\vx-\va\|_2
\right|+
\left|
\|\vx-\va\|_2-r
\right|\\
&\le
\left\|\va-\vb\right\|_2 + \dist\bigl(\vx,\mathcal S(\va, r)\bigr).
\end{split}
\end{align}

Applying this inequality with $\vx=\mathbf X_{i-1}$ and the two sphere centers yields
\begin{align}
\begin{split}
&\operatorname{dist}\bigl(\mathbf X_{i-1},
\mathcal S_{i-1}(\vx_i)\bigr)
\le\operatorname{dist}\bigl(\mathbf X_{i-1},
\mathcal S_{i-1}(\mathbf X_i)\bigr)
+\|\widehat{\vmu}_{i-1}(\mathbf X_i)
-\widehat{\vmu}^\theta_{i-1}(\vx_i)\|_2.
\end{split}
\end{align}
Taking root-mean-square norms and applying Minkowski's inequality, we have
\begin{align}
\label{eq:ddpm_stability_minkowski}
\begin{split}
\left(\mathbb E\left[ \operatorname{dist}\bigl(\mathbf X_{i-1}, \mathcal S_{i-1}(\vx_i)\bigr)^2\right]\right)^{1/2}&\le
\left(\mathbb E\left[
\operatorname{dist}\bigl(\mathbf X_{i-1},
\mathcal S_{i-1}(\mathbf X_i)\bigr)^2
\right]\right)^{1/2}+\left(\mathbb E\left[ \|\widehat{\vmu}_{i-1}(\mathbf X_i) -\widehat{\vmu}^\theta_{i-1}(\vx_i)\|_2^2 \right]\right)^{1/2}.
\end{split}
\end{align}
By construction, $(\mathbf X_{i-1},\mathbf X_i)$ has the exact DDPM joint distribution. Thus, Theorem~\ref{thm:sa_main} implies
\begin{align}
\begin{split}
&\left(\mathbb E\left[
\operatorname{dist}\bigl(\mathbf X_{i-1},
\mathcal S_{i-1}(\mathbf X_i)\bigr)^2
\right]\right)^{1/2}
\le\widehat\sigma_{i-1}\sqrt N
\left(\sqrt{\frac{2}{N}}+
\sqrt{\frac{\beta_i}{\alpha_i(1-\overline{\alpha}_{i-1})}}\right).
\end{split}
\end{align}
Substituting into \eqref{eq:ddpm_stability_minkowski} and dividing by $\widehat\sigma_{i-1}\sqrt N>0$ proves
\eqref{eq:ddpm_spherical_stability}.
\end{proof}

The final term in \eqref{eq:ddpm_spherical_stability} measures the discrepancy between the centers of the ideal and approximate spherical level sets. This discrepancy arises from two sources: 1) the difference between the exact and approximate intermediate states, and 2) the error introduced by the pretrained score network.

\section{PPS under the SMLD Framework}
\label{sm:sec:smld}

In this section we derive progressive posterior sampling (PPS) under the SMLD framework. We use the same measurement model as in the main paper, $\vy=\mA\vx_0+\boldsymbol\epsilon$, where $\boldsymbol\epsilon\sim\mathcal N(\vzero,\sigma_\epsilon^2\mI)$. Here, $\sigma_\epsilon^2$ denotes the measurement noise variance, whereas $\sigma_i^2$ denotes the accumulated variance of the forward
diffusion noise. Define
\begin{align}
    \Delta_i=\sigma_i^2-\sigma_{i-1}^2>0,
    \qquad i=1,\ldots,I.
    \label{eq:smld_increment}
\end{align}
The SMLD forward process and its marginal perturbation kernel are
\begin{align}
    \vx_i
    &=\vx_{i-1}+\sqrt{\Delta_i}\vz_{i-1},
    \qquad \vz_{i-1}\sim\mathcal N(\vzero,\mI),
    \label{eq:smld_forward}\\
    p(\vx_i\mid\vx_0)
    &=\mathcal N(\vx_i;\vx_0,\sigma_i^2\mI),
    \label{eq:smld_forward_marginal}
\end{align}
where the noise increments are mutually independent and independent of $\vx_0$, and $0=\sigma_0<\sigma_1<\cdots<\sigma_I$.

Let $s_\theta(\vx_i,i)$ approximate the score $\nabla_{\vx_i}\log p_i(\vx_i)$ at noise level $\sigma_i$. Tweedie's formula gives the denoised estimate
\cite{efron2011tweedie,song2021scorebased}
\begin{align}
    \widehat{\vx}_0(i)
    =\vx_i+\sigma_i^2s_\theta(\vx_i,i).
    \label{eq:smld_tweedie}
\end{align}
With an exact score, this estimate equals $\mathbb E[\vx_0\mid\vx_i]$. The derivation below first considers $i\geq2$; the final step to $\sigma_0=0$ is treated separately.

\begingroup
\makeatletter
\let\addcontentsline\@gobblethree

\subsection{Gaussian Posterior Approximation}

\makeatletter
\endgroup

\paragraph{I. Local Prior Distribution.}

Conditioned on $\vx_0$, the pair $(\vx_{i-1},\vx_i)$ is jointly Gaussian, with variances $\sigma_{i-1}^2\mI$ and $\sigma_i^2\mI$ respectively, and cross-covariance $\sigma_{i-1}^2\mI$. Gaussian conditioning therefore gives the exact transition distribution
\begin{align}
\begin{split}
    &p(\vx_{i-1}\mid\vx_i,\vx_0)\\
    &\quad=\mathcal N\!\left(
        \vx_{i-1}\left |
        \frac{\sigma_{i-1}^2}{\sigma_i^2}\vx_i
        +\frac{\Delta_i}{\sigma_i^2}\vx_0,
        \frac{\sigma_{i-1}^2\Delta_i}{\sigma_i^2}\mI\right.
    \right).
\end{split}
\label{eq:smld_bridge}
\end{align}

Replacing the unknown clean signal in this distribution by $\widehat{\vx}_0(i)$ yields the local Gaussian prior approximation
\begin{align}
\begin{split}
    h_{i-1}(\vx_{i-1}\mid\vx_i)
    &\mathrel{:=}\left.p(\vx_{i-1}\mid\vx_i,\vx_0)
      \right|_{\vx_0=\widehat{\vx}_0(i)}\\
    &=\mathcal N\!\left(
        \vx_{i-1}\left|\widehat{\vmu}_{i-1},
        \widehat{\sigma}_{i-1}^2\mI\right.
    \right),
\end{split}
\label{eq:smld_local_prior}
\end{align}
where
\begin{align}
    \widehat{\vmu}_{i-1}
    &=\frac{\sigma_{i-1}^2}{\sigma_i^2}\vx_i
      +\frac{\Delta_i}{\sigma_i^2}\widehat{\vx}_0(i)
      \notag\\
    &=\vx_i+\Delta_i s_\theta(\vx_i,i),
    \label{eq:smld_prior_mean}\\
    \widehat{\sigma}_{i-1}^2
    &=\frac{\sigma_{i-1}^2\Delta_i}{\sigma_i^2}.
    \label{eq:smld_prior_variance}
\end{align}
These are the mean and variance of the ancestral sampler for SMLD \cite{song2021scorebased}. With an exact conditional-mean denoiser, $\widehat{\vmu}_{i-1}$ is also the exact reverse conditional mean. The Gaussian distribution in \eqref{eq:smld_local_prior} remains an approximation because it replaces the uncertainty in $\vx_0$ by a single estimate.

\paragraph{II. Conditional Measurement Distribution.}

As in the DDPM formulation, we approximate the intractable conditional distribution of the clean signal by an isotropic Gaussian:
\begin{align}
    p(\vx_0\mid\vx_{i-1})
    \approx q(\vx_0\mid\vx_{i-1})
    =\mathcal N\!\left(
        \vx_0\left|\vx_{i-1},\widetilde{\sigma}_{i-1}^2\mI\right.
    \right).
    \label{eq:smld_clean_approximation}
\end{align}
The mean is $\vx_{i-1}$ because the SMLD forward process adds noise without rescaling the signal. The three variance choices considered in the main paper take the following forms.

\begin{enumerate}
    \item \textbf{Point-mass approximation.}
    Setting
    \begin{align}
        \widetilde{\sigma}_{i-1}^2=0
        \label{eq:smld_variance_zero}
    \end{align}
    corresponds to
    $q(\vx_0\mid\vx_{i-1})=\delta(\vx_0-\vx_{i-1})$.

    \item \textbf{Local moment approximation.}
    We replace the clean signal by its current estimate
    $\widehat{\vx}_0(i)$ and evaluate the squared discrepancy under
    the local prior in \eqref{eq:smld_local_prior}:
    \begin{align}
    \begin{split}
        \widetilde{\sigma}_{i-1}^2
        &=\frac{1}{N}\mathbb E_{\vx_{i-1}\sim h_{i-1}(\cdot|\vx_i)}
          \!\left[
            \left\|\widehat{\vx}_0(i)-\vx_{i-1}\right\|_2^2
          \right]\\
        &=\frac{1}{N}
          \left\|\widehat{\vx}_0(i)
                 -\widehat{\vmu}_{i-1}\right\|_2^2
          +\widehat{\sigma}_{i-1}^2.
    \end{split}
    \label{eq:smld_variance_local}
    \end{align}
    The expectation is taken with $\vx_i$ fixed. This choice includes the additional plug-in approximation of the clean signal by $\widehat{\vx}_0(i)$.

    \item \textbf{Global moment approximation.}
    Averaging the squared discrepancy under the exact forward joint     distribution gives
    \begin{align}
    \begin{split}
        \widetilde{\sigma}_{i-1}^2
        &=\frac{1}{N}\mathbb E_{(\vx_0,\vx_{i-1})}
          \!\left[\|\vx_0-\vx_{i-1}\|_2^2\right]\\
        &=\sigma_{i-1}^2,
    \end{split}
    \label{eq:smld_variance_global}
    \end{align}
    since
    $\vx_{i-1}=\vx_0+\sigma_{i-1}\vz$ with $\vz\sim\mathcal N(\vzero,\mI)$ being independent of $\vx_0$. This identity does not require a particular distribution for the clean signal.
\end{enumerate}

We estimate the measurement noise variance $\widehat{\sigma}_\epsilon^2$ from the current denoised signal $\widehat{\vx}_0(i)$ using
\begin{align}
    \widehat{\sigma}_\epsilon^2
    =\frac{1}{M}\left\|\vy-\mA\widehat{\vx}_0(i)\right\|_2^2.
    \label{eq:smld_measurement_variance}
\end{align}
The dependence of $\widehat{\sigma}_\epsilon^2$ on $i$ is suppressed
for brevity. Both this estimate and the selected
$\widetilde{\sigma}_{i-1}^2$ are held fixed during the current local
update. Integrating the measurement model against
\eqref{eq:smld_clean_approximation} then gives the approximate likelihood
\begin{align}
\begin{split}
    p_{i-1}(\vy\mid\vx_{i-1})&\approx\int
        \mathcal N(\vy\mid\mA\vx_0,\widehat{\sigma}_\epsilon^2\mI)
        \cdot q(\vx_0\mid\vx_{i-1})\,\mathrm d\vx_0\\
    &=\mathcal N\!\left(
        \vy;\mA\vx_{i-1},\mR_{i-1}
    \right),
\end{split}
\label{eq:smld_likelihood}
\end{align}
where
\begin{align}
    \mR_{i-1}
    =\widetilde{\sigma}_{i-1}^2\mA\mA^T
     +\widehat{\sigma}_\epsilon^2\mI.
    \label{eq:smld_likelihood_covariance}
\end{align}

\paragraph{III. Local Posterior Distribution.}

For the current solution $\vx_i$, combining the local prior and approximate likelihood yields a Gaussian approximation to the posterior distribution $p_{i-1}(\vx_{i-1}\mid\vy)$:
\begin{align}
\begin{split}
    p_{i-1}(\vx_{i-1}|\vy)&\approx p_{i-1}(\vx_{i-1}\mid\vx_i,\vy)\\
    &=\mathcal N\!\left(
        \vx_{i-1};\overline{\vmu}_{i-1},
        \overline{\mSigma}_{i-1}
    \right),
\end{split}
\label{eq:smld_posterior}
\end{align}
where dependence on the fixed $\vx_i$ and estimated parameters is suppressed in the notation for $p_{i-1}$.
Define
\begin{align}
\begin{split}
    \mB_{i-1}
    &=\widehat{\sigma}_{i-1}^2\mA\mA^T+\mR_{i-1}\\
    &=\left(\widehat{\sigma}_{i-1}^2
           +\widetilde{\sigma}_{i-1}^2\right)\mA\mA^T
      +\widehat{\sigma}_\epsilon^2\mI.
\end{split}
\label{eq:smld_B}
\end{align}
The posterior mean and covariance are
\begin{align}
\begin{split}
    \overline{\vmu}_{i-1}
    ={}&\widehat{\vmu}_{i-1}
       +\widehat{\sigma}_{i-1}^2\mA^T\mB_{i-1}^{-1}\cdot
       \left(\vy-\mA\widehat{\vmu}_{i-1}\right),
\end{split}
\label{eq:smld_posterior_mean}\\
    \overline{\mSigma}_{i-1}
    ={}&\widehat{\sigma}_{i-1}^2\mI
       -\widehat{\sigma}_{i-1}^4
        \mA^T\mB_{i-1}^{-1}\mA.
    \label{eq:smld_posterior_covariance}
\end{align}

To obtain these expressions, observe that the local Gaussian model has cross-covariance $\operatorname{Cov}(\vx_{i-1},\vy)=\widehat{\sigma}_{i-1}^2\mA^T$ and measurement covariance $\operatorname{Cov}(\vy)=\mB_{i-1}$. The result follows directly from Gaussian conditioning. These expressions assume that $\mB_{i-1}$ is nonsingular, which is guaranteed when $\widehat{\sigma}_\epsilon^2>0$. A small positive floor may be applied to the residual variance estimate when needed.

Only the posterior mean is required for the PPS update. It can be evaluated by solving
\begin{align}
    \mB_{i-1}\vv
    =\vy-\mA\widehat{\vmu}_{i-1}
    \label{eq:smld_linear_solve}
\end{align}
using the conjugate gradient method and setting $\overline{\vmu}_{i-1}=\widehat{\vmu}_{i-1}+\widehat{\sigma}_{i-1}^2\mA^T\vv$. Thus, the formulation does not require an explicit SVD of $\mA$ or the construction of $\overline{\mSigma}_{i-1}$.

\begingroup
\makeatletter
\let\addcontentsline\@gobblethree

\subsection{Manifold Projection}
\label{sm:sec:smld_spherical}

\makeatletter
\endgroup

We next establish an SMLD counterpart of Theorem~\ref{thm:sa_main}. To distinguish samples $\{\vx_i\}_{i=0}^I$ generated by the approximate reverse process from those associated with the exact diffusion process, we use $\{\mathbf X_i\}_{i=0}^I$ to denote random variables following the exact forward or reverse diffusion distributions. The following theorem establishes the corresponding spherical approximation bound under SMLD, with a proof analogous to that of the DDPM case. We finally present Corollary~\ref{cor:smld_spherical_stability}, which bounds the additional error from approximate reverse states and pre-trained score function through the discrepancy between the sphere centers.

\begin{theorem}[Spherical approximation under SMLD]
\label{thm:sa_smld}
Consider the SMLD forward process with $\mathbb E\|\mathbf X_0\|_2^2<\infty$. Fix $i\in\{2,\ldots,I\}$, let $\widehat{\mathbf X}_0(i)=\mathbb E[\mathbf X_0\mid\mathbf X_i]$, and define the spherical level set
\begin{align}
\label{eq:sa_smld_sphere}
    \mathcal S_{i-1}(\mathbf X_i)
    =
    \left\{
    \vx:
    \|\vx-\widehat{\vmu}_{i-1}(\mathbf X_i)\|_2
    =\widehat{\sigma}_{i-1}\sqrt N
    \right\}.
\end{align}
Draw $\mathbf X_i\sim p_i(\cdot)$, where $p_i$ is the forward marginal, and an exact reverse sample $\mathbf X_{i-1}\sim p_{i-1}(\cdot|\mathbf X_i)$. Then
\begin{align}
\label{eq:sa_smld_bound}
\begin{split}
    &\frac{
    \left(
    \mathbb E\left[
    \dist\bigl(
    \mathbf X_{i-1},
    \mathcal S_{i-1}(\mathbf X_i)
    \bigr)^2
    \right]
    \right)^{1/2}
    }{
    \widehat{\sigma}_{i-1}\sqrt N
    }\le
    \sqrt{\frac{2}{N}}
    +
    \sqrt{
    \frac{\sigma_i^2-\sigma_{i-1}^2}
    {\sigma_{i-1}^2}
    }.
\end{split}
\end{align}
The expectation is taken over both $\mathbf X_i$ and $\mathbf X_{i-1}$. 
\end{theorem}

\begin{proof}
Under the SMLD framework, we have the local exact reverse distribution of $\mathbf X_{i-1}$ conditioned on $\mathbf X_i,\mathbf X_0$
\begin{align}
\label{eq:sa_smld_bridge}
\begin{split}
    &p_{i-1}(\mathbf X_{i-1}|\mathbf X_i,\mathbf X_0)=\mathcal N\left(
    \frac{\sigma_{i-1}^2}{\sigma_i^2}\mathbf X_i
    +\frac{\Delta_i}{\sigma_i^2}\mathbf X_0,
    \widehat{\sigma}_{i-1}^2\mI
    \right).
\end{split}
\end{align}
Let $\gamma_i=\Delta_i/\sigma_i^2$. An exact reverse sample can then be represented as
\begin{align}
\label{eq:sa_smld_representation}
\begin{split}
    \mathbf X_{i-1}
    =&\widehat{\vmu}_{i-1}(\mathbf X_i)
    +\gamma_i\left(
    \mathbf X_0-\widehat{\mathbf X}_0(i)
    \right)+\widehat{\sigma}_{i-1}\mathbf Z,
\end{split}
\end{align}
where $\mathbf Z\sim\mathcal N(\vzero,\mI)$ is independent of $(\mathbf X_0,\mathbf X_i)$, and the center of the spherical level set is
\begin{align}
    \widehat{\vmu}_{i-1}(\mathbf X_i)=\frac{\sigma_{i-1}^2}{\sigma_i^2}\mathbf X_i+\gamma_i\widehat{\mathbf X}_0(i)
\end{align}

The distance from a point to a sphere equals the absolute difference between its distance to the center and the sphere's radius. Using \eqref{eq:sa_smld_representation},
we obtain
\begin{align}
\label{eq:sa_smld_triangle}
\begin{split}
\dist\bigl(
\mathbf X_{i-1},
\mathcal S_{i-1}(\mathbf X_i)
\bigr) &=\left|\|\mathbf X_{i-1}-\widehat{\vmu}_{i-1}(\mX_i)\|_2-\widehat{\sigma}_{i-1}\sqrt N\right|\\
&=
\left|
\left\|
\gamma_i(\mathbf X_0-\widehat{\mathbf X}_0(i))
+\widehat{\sigma}_{i-1}\mathbf Z
\right\|_2
-\widehat{\sigma}_{i-1}\sqrt N
\right|\\
&\overset{(a)}{\le}
\left|
\left\|
\gamma_i(\mathbf X_0-\widehat{\mathbf X}_0(i))
+\widehat{\sigma}_{i-1}\mathbf Z
\right\|_2
-\widehat{\sigma}_{i-1}\|\mathbf Z\|_2
\right|
+
\widehat{\sigma}_{i-1}
\left|\|\mathbf Z\|_2-\sqrt N\right|\\
&\overset{(b)}{\le}
\gamma_i\|\mathbf X_0-\widehat{\mathbf X}_0(i)\|_2
+
\widehat{\sigma}_{i-1}
\left|\|\mathbf Z\|_2-\sqrt N\right|,
\end{split}
\end{align}
where (a) follows from the triangle inequality $|u+v|\le |u|+|v|$, and (b) follows from the reverse triangle inequality $|\|\vx+\mathbf v\|_2-\|\mathbf v\|_2|\le\|\vx\|_2$.

Since $\|\mathbf Z\|_2^2\sim\chi_N^2$ follows the chi-squared distribution and thus has mean $N$ and variance $2N$, we have
\begin{align}
\begin{split}
    \mathbb E\!\left[\left(\|\mathbf Z\|_2-\sqrt N\right)^2\right]
    &=\mathbb E\!\left[\left(\frac{\|\mathbf Z\|_2^2-N}{\|\mathbf Z\|_2+\sqrt N}\right)^2\right]\\
    &= \mathbb E\!\left[\frac{\left(\|\mathbf Z\|_2^2-N\right)^2}{\left(\|\mathbf Z\|_2+\sqrt N\right)^2}\right]\\
    &\le \frac{\mathbb E[\left(\|\mathbf Z\|_2^2-N\right)^2]}{N}\\
    &=2.
\end{split}
\end{align}

Taking root-mean-square norms of \eqref{eq:sa_smld_triangle} and applying Minkowski's inequality gives
\begin{align}
\label{eq:sa_smld_intermediate}
\begin{split}
    &\left(
    \mathbb E\left[
    \dist\bigl(
    \mathbf X_{i-1},
    \mathcal S_{i-1}(\mathbf X_i)
    \bigr)^2
    \right]
    \right)^{1/2}
    \le
    \gamma_i
    \left(
    \mathbb E
    \|\mathbf X_0-\widehat{\mathbf X}_0(i)\|_2^2
    \right)^{1/2}
    +
    \sqrt{2}\widehat{\sigma}_{i-1}.
\end{split}
\end{align}

Because $\widehat{\mathbf X}_0(i)=\mathbb E[\mathbf X_0|\mathbf X_i]$ minimizes the mean-squared estimation error, we have
\begin{align}
\label{eq:sa_smld_mmse}
\begin{split}
    &\mathbb E
    \|\mathbf X_0-\widehat{\mathbf X}_0(i)\|_2^2
    \le
    \mathbb E
    \|\mathbf X_0-\mathbf X_i\|_2^2
    =N\sigma_i^2,
\end{split}
\end{align}
where the last equality follows from the SMLD forward process.

Combining \eqref{eq:sa_smld_intermediate} and \eqref{eq:sa_smld_mmse} and dividing by $\widehat{\sigma}_{i-1}\sqrt N>0$, we obtain
\begin{align}
\begin{split}
    \frac{
    \left(
    \mathbb E\left[
    \dist\bigl(
    \mathbf X_{i-1},
    \mathcal S_{i-1}(\mathbf X_i)
    \bigr)^2
    \right]
    \right)^{1/2}
    }{
    \widehat{\sigma}_{i-1}\sqrt N
    }
    &\le
    \sqrt{\frac{2}{N}}
    +
    \frac{\gamma_i\sigma_i}
    {\widehat{\sigma}_{i-1}}\\
    &=
    \sqrt{\frac{2}{N}}
    +
    \sqrt{
    \frac{\Delta_i}{\sigma_{i-1}^2}
    },
\end{split}
\end{align}
where we used
\begin{align}
    \gamma_i=\frac{\Delta_i}{\sigma_i^2},
    \qquad
    \widehat{\sigma}_{i-1}^2
    =\frac{\sigma_{i-1}^2\Delta_i}{\sigma_i^2}.
\end{align}
\end{proof}

\begingroup
\makeatletter
\let\addcontentsline\@gobblethree

\subsubsection{Spherical Approximation with Approximate Reverse States}\
\label{sec:smld_approximate_states}

\makeatletter
\endgroup

Theorem~\ref{thm:sa_smld} analyzes the spherical approximation under the exact forward marginal and an exact conditional-mean denoiser. For practical PPS states $\vx_i$, we construct a reference sphere with radius $\widehat{\sigma}_{i-1}\sqrt{N}$ and a center computed using the pretrained score function $s_\theta(\vx_i,i)$. Corollary~\ref{cor:smld_spherical_stability} quantifies how the discrepancy between the exact and approximate centers affects the normalized root-mean-square radial distance from exact reverse samples to this reference sphere.

\begin{corollary}[Stability of the spherical approximation]
\label{cor:smld_spherical_stability}
Consider the SMLD framework under the assumptions of Theorem~\ref{thm:sa_smld}. Let $\mathbf X_i\sim p_i(\mathbf X_i)$ and $\vx_i\sim \widehat{p}_i(\vx_i|\vx_{i+1},\widehat{\vx}_0(i+1))$, where $p_i$ is the exact forward marginal and $\widehat{p}_i$ is the distribution of intermediate states $\vx_i$ generated by the approximate reverse process. Conditional on $\mathbf X_i$, we draw $\mathbf X_{i-1}$ from the exact reverse distribution $p_{i-1}(\cdot\mid\mathbf X_i)$ independently of $\vx_i$.

Define the exact and approximate local reverse means as
\begin{align}
\label{eq:exact_reverse_mean_smld}
\widehat{\vmu}_{i-1}(\mathbf X_i)
&=\mathbf X_i+\Delta_i\cdot\nabla\log p_i(\mathbf X_i),\\
\label{eq:approx_reverse_mean_smld}
\widehat{\vmu}^\theta_{i-1}(\vx_i)
&=
\vx_i+\Delta_i\cdot s_\theta(\vx_i,i),
\end{align}
where $\Delta_i=\sigma_i^2-\sigma_{i-1}^2$ and $s_\theta(\vx_i,i)$ is the pretrained score function.

The spherical level set constructed approximately by PPS is
\begin{align}
\mathcal S_{i-1}(\vx_i)
=
\left\{
\vx:
\|\vx-\widehat{\vmu}^\theta_{i-1}(\vx_i)\|_2
=
\widehat{\sigma}_{i-1}\sqrt N
\right\},
\end{align}
where $\widehat{\sigma}_{i-1}^2=\frac{\sigma_{i-1}^2\Delta_i}{\sigma_i^2}$.

We then have,
\begin{align}
\label{eq:smld_spherical_stability}
\begin{split}
&\frac{
\left(
\mathbb E\left[
\dist\bigl(
\mathbf X_{i-1},\ 
\mathcal S_{i-1}(\vx_i)
\bigr)^2
\right]
\right)^{1/2}
}{
\widehat{\sigma}_{i-1}\sqrt N
}
\le
\sqrt{\frac{2}{N}}
+
\sqrt{\frac{\Delta_i}{\sigma_{i-1}^2}}
+
\frac{
\left(
\mathbb E\left[
\left\|\widehat{\vmu}_{i-1}(\mathbf X_i)
-\widehat{\vmu}^\theta_{i-1}(\vx_i)\right\|_2^2
\right]
\right)^{1/2}
}{
\widehat{\sigma}_{i-1}\sqrt N
}.
\end{split}
\end{align}
The expectation is taken over the joint distribution of $(\mathbf X_{i-1},\mathbf X_i,\vx_i)$.
\end{corollary}

\begin{proof}
Define the ideal spherical approximation centered at the exact local reverse mean $\widehat{\vmu}_{i-1}(\mathbf X_i)$
\begin{align}
\mathcal S_{i-1}(\mathbf X_i)
=
\left\{
\vx:
\|\vx-\widehat{\vmu}_{i-1}(\mathbf X_i)\|_2
=
\widehat{\sigma}_{i-1}\sqrt N
\right\}.
\end{align}

For any point $\vx$ and two spheres with the same radius $r$ but different centers $\va$ and $\vb$, the reverse triangle inequality gives
\begin{align}
\begin{split}
&\dist\bigl(\vx,\mathcal S(\vb,r)\bigr)\\
&=
\left|\left\|\vx-\vb\right\|_2-r\right|\\
&=
\left|
\bigl(\|\vx-\vb\|_2
-\|\vx-\va\|_2\bigr)
+
\bigl(\|\vx-\va\|_2-r\bigr)
\right|\\
&\le
\left|
\|\vx-\vb\|_2
-\|\vx-\va\|_2
\right|+
\left|
\|\vx-\va\|_2-r
\right|\\
&\le
\left\|\va-\vb\right\|_2 + \dist\bigl(\vx,\mathcal S(\va, r)\bigr).
\end{split}
\end{align}

Applying this inequality to the ideal and approximate spheres yields\
\begin{align}
\begin{split}
&\dist\bigl(\mathbf X_{i-1}, \mathcal S_{i-1}(\vx_i)\bigr)
\le
\dist\bigl(\mathbf X_{i-1}, \mathcal S_{i-1}(\mathbf X_i)\bigr)
+
\|\widehat{\vmu}_{i-1}(\mathbf X_i)-\widehat{\vmu}^\theta_{i-1}(\vx_i)\|_2.
\end{split}\
\end{align}

Taking root-mean-square norms and applying Minkowski's inequality, we obtain
\begin{align}
\begin{split}
&\left(
\mathbb E\left[
\dist\bigl(
\mathbf X_{i-1},
\mathcal S_{i-1}(\vx_i)
\bigr)^2
\right]
\right)^{1/2}
\le
\left(
\mathbb E\left[
\dist\bigl(
\mathbf X_{i-1},
\mathcal S_{i-1}(\mathbf X_i)
\bigr)^2
\right]
\right)^{1/2}
+
\left(
\mathbb E\left[
\|\widehat{\vmu}_{i-1}(\mathbf X_i)
-\widehat{\vmu}^\theta_{i-1}(\vx_i)\|_2^2
\right]
\right)^{1/2}.
\end{split}
\end{align}

Theorem~\ref{thm:sa_smld} gives
\begin{align}
\begin{split}
&\left(
\mathbb E\left[
\dist\bigl(
\mathbf X_{i-1},
\mathcal S_{i-1}(\mathbf X_i)
\bigr)^2
\right]
\right)^{1/2}
\le
\widehat{\sigma}_{i-1}\sqrt N
\left(
\sqrt{\frac{2}{N}}
+
\sqrt{\frac{\Delta_i}{\sigma_{i-1}^2}}
\right).
\end{split}
\end{align}

Substituting this bound and dividing by $\widehat{\sigma}_{i-1}\sqrt N>0$ proves \eqref{eq:smld_spherical_stability}.
\end{proof}

\begingroup
\makeatletter
\let\addcontentsline\@gobblethree

\subsection{Progressive Posterior Guidance and Manifold Projection.}

\makeatletter
\endgroup

We first draw a sample from the local prior,
\begin{align}
    \widehat{\vx}_{i-1}
    =\widehat{\vmu}_{i-1}
     +\widehat{\sigma}_{i-1}\vz,
    \qquad \vz\sim\mathcal N(\vzero,\mI),
    \label{eq:smld_prior_sample}
\end{align}
and then move it toward the approximate posterior mean using a damped update:
\begin{align}
    \overline{\vx}_{i-1}
    =\widehat{\vx}_{i-1}
     +\rho\left(\overline{\vmu}_{i-1}
                 -\widehat{\vx}_{i-1}\right),
    \qquad 0<\rho<1.
    \label{eq:smld_guidance}
\end{align}

The local Gaussian prior is isotropic, so its density level sets are spheres centered at $\widehat{\vmu}_{i-1}$. The preceding theorem and corollary characterize this geometry using the reference radius $\widehat{\sigma}_{i-1}\sqrt{N}$. In the PPS update, we use the radius of the initial local-prior sample:
\begin{align}
    r_{i-1}
    =
    \|\widehat{\vx}_{i-1}-\widehat{\vmu}_{i-1}\|_2.
    \label{eq:smld_radius}
\end{align}
Since $r_{i-1}^2/\widehat{\sigma}_{i-1}^2\sim\chi_N^2$, the relative radius $r_{i-1}/(\widehat{\sigma}_{i-1}\sqrt{N})$ concentrates around one as $N$ increases. Thus, the sphere used by PPS shares the reference sphere's center and has a radius that is close to the reference radius in relative terms with high probability in high dimensions. We project the guided sample onto this sphere:
\begin{align}
    \vx_{i-1}
    =\widehat{\vmu}_{i-1}
     +r_{i-1}
      \frac{\overline{\vx}_{i-1}-\widehat{\vmu}_{i-1}}
           {\left\|\overline{\vx}_{i-1}
                   -\widehat{\vmu}_{i-1}\right\|_2}.
    \label{eq:smld_projection}
\end{align}
This update preserves the initial sample's distance from the local prior mean and hence its density level under the local Gaussian prior. If the denominator in \eqref{eq:smld_projection} vanishes, we retain $\widehat{\vx}_{i-1}$, which already lies on the prescribed sphere.

\paragraph{Final Denoising Step.}

At $i=1$, the convention $\sigma_0=0$ gives $\widehat{\sigma}_0^2=0$ and
$\widehat{\vmu}_0=\widehat{\vx}_0(1)$. The local Gaussian approximation therefore collapses to a point mass, and the spherical projection is not applied. After performing the guided updates for $i=I,\ldots,2$, we return
\begin{align}
    \vx_0
    =\widehat{\vx}_0(1)
    =\vx_1+\sigma_1^2s_\theta(\vx_1,1).
    \label{eq:smld_final_denoise}
\end{align}
This is the deterministic final denoising step of the local ancestral sampling. Algorithm~\ref{alg:pps_smld} summarizes the procedure under the SMLD framework.

\begin{algorithm}[tbp]
\caption{PPS under the SMLD Framework}
\label{alg:pps_smld}
\begin{algorithmic}[1]
    \Require $\mA,\vy,s_\theta,\{\sigma_i\}_{i=0}^{I},\rho\in(0,1)$,
    and a variance-selection rule
    \State Initialize
    $\vx_I\sim\mathcal N(\vzero,\sigma_I^2\mI)$
    \For{$i=I,\ldots,2$}
        \State Compute $\widehat{\vx}_0(i)$ using
        \eqref{eq:smld_tweedie}
        \State Compute $\widehat{\vmu}_{i-1}$ and
        $\widehat{\sigma}_{i-1}^2$ using
        \eqref{eq:smld_prior_mean}--\eqref{eq:smld_prior_variance}
        \State Choose $\widetilde{\sigma}_{i-1}^2$ using
        \eqref{eq:smld_variance_zero},
        \eqref{eq:smld_variance_local}, or
        \eqref{eq:smld_variance_global}
        \State Estimate $\widehat{\sigma}_\epsilon^2$ using
        \eqref{eq:smld_measurement_variance}
        \State Obtain $\overline{\vmu}_{i-1}$ using
        \eqref{eq:smld_B} and \eqref{eq:smld_linear_solve}
        \State Draw $\widehat{\vx}_{i-1}$ using
        \eqref{eq:smld_prior_sample}
        \State Compute $\overline{\vx}_{i-1}$ using
        \eqref{eq:smld_guidance}
        \State Compute $r_{i-1}$ using
        \eqref{eq:smld_radius}
        \If{$\|\overline{\vx}_{i-1}
                -\widehat{\vmu}_{i-1}\|_2>0$}
            \State Compute $\vx_{i-1}$ using
            \eqref{eq:smld_projection}
        \Else
            \State Set $\vx_{i-1}=\widehat{\vx}_{i-1}$
        \EndIf
    \EndFor
    \State \Return $\vx_0=\vx_1+\sigma_1^2s_\theta(\vx_1,1)$
\end{algorithmic}
\end{algorithm}

\FloatBarrier

\section{Experimental Details}
\label{sm:sec:Experimental_Details}

\subsection{Hyperparameter Settings}

PPS has two hyperparameters: $\rho$ in \eqref{eq:move_high_prob} and $\eta$. We do not tune $\eta$ for any experiment, the value same as DDNM~\cite{wang2023zeroshot} and DDRM~\cite{kawar2022denoising}. Actually, PPS has only one hyperparameter. The best $\rho$ values are listed in below table. $\rho$ lies in a narrow range of $[0.5, 1]$ and follows a consistent trend: it decreases as the measurement noise $\sigma_\epsilon$ grows, and inpainting prefers a larger $\rho$ than other scenarios. It proves our approach does not require extensive tuning for each tasks and has enough robustness. 

\begin{table}[H]
\centering
\setlength{\tabcolsep}{3pt}
\begin{tabular}{lccccccc}
\toprule
\multirow{2}{*}{Task} & \multicolumn{6}{c}{$\rho$} & \multirow{3}{*}{$\eta$} \\
\cmidrule(lr){2-7}
 & \multicolumn{3}{c}{FFHQ} & \multicolumn{3}{c}{ImageNet} &  \\
\cmidrule(lr){2-4}\cmidrule(lr){5-7}
$\sigma_\epsilon$ & 0.05 & 0.15 & 0.3 & 0.05 & 0.15 & 0.3 &  \\
\midrule
Motion deblurring & 0.7 & 0.6 & 0.6 & 0.6 & 0.5 & 0.5 & \multirow{4}{*}{0.85} \\
Inpainting & 1.0 & 0.8 & 0.7 & 1.0 & 0.7 & 0.7 &  \\
Super-resolution & 0.7 & 0.6 & 0.6 & 0.7 & 0.6 & 0.5 &  \\
Gaussian deblurring & 0.6 & 0.6 & 0.6 & 0.7 & 0.6 & 0.5 &  \\
\bottomrule
\end{tabular}
\caption{Hyperparameters used for each task, dataset and noise level. $\eta=0.85$ is shared by all tasks.}
\label{tab:hyperparameters}
\end{table}

\FloatBarrier

Following DDIM~\cite{song2021ddim}, DDNM~\cite{wang2023zeroshot}, DDRM~\cite{kawar2022denoising} and ours adopt a non-Markovian sampling process. At each step $t$, given the rectified estimate $\hat{\mathbf{x}}_{0|t}$, the next sample is obtained by 
\begin{equation}
\mathbf{x}_{t-1} = \sqrt{\bar{\alpha}_{t-1}}\,\hat{\mathbf{x}}_{0|t}
+ \sqrt{1-\bar{\alpha}_{t-1}}\left(\eta\,\boldsymbol{\epsilon}
+ \sqrt{1-\eta^{2}}\,\boldsymbol{\epsilon}_{\theta}(\mathbf{x}_t, t)\right),
\quad \boldsymbol{\epsilon}\sim\mathcal{N}(\mathbf{0},\mathbf{I}),
\label{eq:sampling}
\end{equation}
here $\eta \in [0,1]$ controls the stochasticity of the sampling process.
Specifically, $\eta$ balances the contribution of freshly sampled Gaussian
noise $\boldsymbol{\epsilon}$ against the noise $\boldsymbol{\epsilon}_{\theta}$
predicted by the network, while keeping the total noise variance at
$1-\bar{\alpha}_{t-1}$. Setting $\eta=0$ yields a fully deterministic sampler,
whereas a larger $\eta$ injects more randomness.

\subsection{Baseline Details}

\textbf{DPS} We keep the official DPS setup (1000-step DDPM sampling) and tune only the guidance step size $\zeta$. For $\sigma_\epsilon=0.05$ we use the official values ($\zeta=0.3$ for SR and deblurring, $0.5$ for inpainting), which were tuned for this noise level. For $\sigma_\epsilon\in\{0.15, 0.3\}$ we re-tune $\zeta$ for each dataset and task. We search the grid $\zeta\in\{0.3, 0.5, 0.7, 1.0\}$. On FFHQ the selected value is $\zeta=1.0$ for all four tasks at both noise levels. On ImageNet we select $\zeta=1.0$ for SR and inpainting at both noise levels. For Gaussian deblurring and motion deblurring the selected values are $0.3$ and $0.5$ at $\sigma_\epsilon=0.15$, and $0.5$ and $0.7$ at $\sigma_\epsilon=0.3$.
\\
\\
\textbf{DMPS} We use the official implementation with the default 1000-step DDPM sampler. The only hyperparameter is the step size $\lambda$ that scales the likelihood update. For $\sigma_y=0.05$, we adopt the default value $\lambda=1.75$ from the official code for all tasks. For $\sigma_\epsilon\in\{0.15, 0.3\}$, we run a grid search over $\lambda\in\{0.5, 1.0, 1.75, 3.0\}$. We select $\lambda=1.75$ for FFHQ SR and $\lambda=1.0$ for FFHQ Gaussian deblurring and both ImageNet tasks. For $\sigma_y=0.15$, we use $\lambda=0.75$ for all tasks. 
\\
\\
\textbf{MPGD}
We use the MPGD-Z variant, which projects the guidance step onto the latent space of a pretrained VQGAN~\cite{esser2021taming} autoencoder for $0.3<t<0.5$ and takes it in pixel space otherwise, using the LDM~\cite{rombach2022latentdiffusion} CelebA-HQ autoencoder for FFHQ and the LDM ImageNet (VQ-f8) autoencoder for ImageNet, with 100 DDIM steps and $\eta=0.5$. The guidance scale is the only tuned hyperparameter: for each dataset, task and noise level we grid-search over a log-spaced grid from 0.25 to 40. Listing values for $\sigma_\epsilon=0.05/0.15/0.3$, this gives a scale of 5 for SR in all settings; 20/10/5 on FFHQ and 10/10/5 on ImageNet for Gaussian deblurring; 20/10/10 on FFHQ and 10 throughout on ImageNet for motion deblurring; and 5/5/2.5 on FFHQ and 2.5 throughout on ImageNet for inpainting, where the FFHQ SR value at $\sigma_\epsilon=0.05$ coincides with the authors' setting.
\\
\\
\textbf{DAPS}
We use the pixel-space DAPS-1K configuration of the official implementation and keep its sampler schedule fixed: 200 annealing steps on an EDM~\cite{Karras2022ElucidatingTD} schedule ($\sigma_{\max}=100$, $\sigma_{\min}=0.1$), 5 Euler ODE steps per annealing step, and 100 Langevin steps per annealing step. We tune the three Langevin hyperparameters: the likelihood scale $\tau$, the step size $\eta$, and the decay ratio $\delta$, where $\eta$ is linearly annealed to $\delta\eta$. Because the data-consistency update scales with $\eta/\tau^2$, we parametrize $\eta=c,\tau^2$ and search over $(\tau,c)$; the default setting of the paper ($\tau=0.01$, $\eta=10^{-4}$) corresponds to $c=1$. For $\sigma_\epsilon=0.05$ we search $\tau\in\{0.01,0.03,0.05\}$ and $c\in\{0.25,1\}$ with $\delta=0.01$. All tasks select $\tau=0.01$ and $\delta=0.01$, with $\eta=10^{-4}$ for SR, Gaussian deblurring and motion deblurring on FFHQ and for Gaussian deblurring on ImageNet, and $\eta=2.5\times10^{-5}$ otherwise. For $\sigma_\epsilon=0.15$ we search $\tau\in\{0.01,0.03,0.1\}$ and $c\in\{0.1,0.25\}$, plus $(\tau,c)=(0.01,1)$, with $\delta=0.01$. Every task on both datasets selects $\tau=0.01$, $\eta=10^{-5}$ and $\delta=0.01$. For $\sigma_\epsilon=0.3$ we search $\tau\in\{0.03,0.1,0.3\}$ and $c\in\{0.25,1\}$ with $\delta=0.01$, plus additional $c=4$ and $\delta=0.001$ runs for SR and Gaussian deblurring. On both datasets, SR and Gaussian deblurring select $\tau=0.03$, $\eta=2.25\times10^{-4}$ and $\delta=0.001$; inpainting selects $\tau=0.03$, $\eta=2.25\times10^{-4}$ and $\delta=0.01$; and motion deblurring selects $\tau=0.3$, $\eta=0.0225$ and $\delta=0.01$.
\\
\\
\textbf{CoDPS}
We use the official implementation with its DDIM variant (CoDPS+, 100 sampling steps) and tune only the guidance scale $\zeta$. We search $\zeta\in\{0.02, 0.05, 0.1, 0.15\}$ for motion deblurring, Gaussian deblurring, and SR, and $\zeta\in\{0.2, 0.4, 0.7, 1.0\}$ ($\sigma_\epsilon=0.05$) or $\zeta\in\{0.05, 0.1, 0.2, 0.4, 0.7, 1.0\}$ ($\sigma_\epsilon=0.15, 0.3$) for inpainting. The selected scales are as follows: for $\sigma_\epsilon=0.05$, $\zeta=0.05$ for both deblurring tasks and SR on both datasets, except ImageNet motion deblurring, where $\zeta=0.02$ and $\zeta=0.05$ performed on par and we kept the default $\zeta=0.04$, and $\zeta=0.2$ for inpainting on both datasets; for $\sigma_\epsilon=0.15$, $\zeta=0.05$ for both deblurring tasks and SR, and $\zeta=0.1$ (FFHQ) and $\zeta=0.2$ (ImageNet) for inpainting; for $\sigma_\epsilon=0.3$, $\zeta=0.05$ for both deblurring tasks, $\zeta=0.1$ for SR, and $\zeta=0.05$ for inpainting on both datasets.
\\
\\
\textbf{Peng et al.}
We use Type~I guidance with the Convert posterior covariance, with the 50-step Heun ODE sampler on the Karras noise schedule ($\sigma_t\in[0.01,80]$, 99 NFEs); the remaining sampler settings follow the official implementation. The only hyperparameter we tune is the covariance threshold $\tau$: at diffusion noise levels $\sigma_t<\tau$ the posterior covariance is converted from the learned reverse-process variances, whereas for $\sigma_t\geq\tau$ the analytic isotropic variance $\sigma_t^2/(1+\sigma_t^2)$ is used. For $\sigma_\epsilon=0.05$ we keep the default $\tau=0.2$ for all tasks. For $\sigma_\epsilon=0.3$ we search $\tau\in\{0.1,0.2,0.5,1,2,5\}$ separately for each dataset and task, which gives $\tau=2$ for inpainting and SR and $\tau=1$ for Gaussian and motion deblurring on both datasets. For $\sigma_\epsilon=0.15$ we keep $\tau=0.2$ for deblurring and SR, since at $\sigma_\epsilon=0.3$ tuning changed these tasks by only about $0.1$\,dB PSNR, and tune only inpainting over $\tau\in\{0.2,1,2\}$, extended to $\tau=5$ on ImageNet where $\tau=2$ was still the best; this gives $\tau=1$ on FFHQ and $\tau=2$ on ImageNet, where we keep $\tau=2$ because $\tau=5$ improved PSNR by less than $0.01$\,dB and LPIPS by $0.002$.
\\
\\
\textbf{DDRM}
We adopt the default hyperparameters recommended by the authors ($\eta=0.85$, $\eta_b=1$) for all tasks and noise levels, without task-specific tuning.
\\
\\
\textbf{DDNM}
It has only a single hyperparameters $\eta$, so we follow official default $\eta = 0.85$ for any task on both datasets.

\FloatBarrier

\section{Additional Experimental Results}
\label{sm:sec:Additional_Results}

\subsection{Additional Ablation Study 1: with projection vs. without projection}
\label{sm:sub:additional_ablation_study_proj}
Manifold projection is the key idea of PPS: at each step, the intermediate estimate $\overline{\vx}_{i-1}$ is projected back onto the data manifold
before sampling continues (~\eqref{eq:manifold_projection}). To isolate its contribution, we remove the projection and directly use the unprojected estimate (guided sample in \eqref{eq:move_high_prob}):
\begin{equation}
  \vx_{i-1} = \overline{\vx}_{i-1},
\end{equation}

while keeping all other components unchanged. To ensure a fair comparison, we re-tune $\rho$ for this variant in every setting on the same held-out split used for our method.


\begin{table}[H]
\centering
\setlength{\tabcolsep}{3pt}
\resizebox{\textwidth}{!}{%
\begin{tabular}{clccccccccc|ccccccccc}
\toprule
\multirow{3}{*}{Task} & \multirow{3}{*}{Variant} & \multicolumn{9}{c|}{FFHQ} & \multicolumn{9}{c}{ImageNet} \\
\cmidrule(lr){3-11}\cmidrule(lr){12-20}
 &  & \multicolumn{3}{c}{$\sigma_\epsilon=0.05$} & \multicolumn{3}{c}{$\sigma_\epsilon=0.15$} & \multicolumn{3}{c|}{$\sigma_\epsilon=0.3$} & \multicolumn{3}{c}{$\sigma_\epsilon=0.05$} & \multicolumn{3}{c}{$\sigma_\epsilon=0.15$} & \multicolumn{3}{c}{$\sigma_\epsilon=0.3$} \\
\cmidrule(lr){3-5}\cmidrule(lr){6-8}\cmidrule(lr){9-11}\cmidrule(lr){12-14}\cmidrule(lr){15-17}\cmidrule(lr){18-20}
 &  & PSNR$\uparrow$ & SSIM$\uparrow$ & LPIPS$\downarrow$ & PSNR$\uparrow$ & SSIM$\uparrow$ & LPIPS$\downarrow$ & PSNR$\uparrow$ & SSIM$\uparrow$ & LPIPS$\downarrow$ & PSNR$\uparrow$ & SSIM$\uparrow$ & LPIPS$\downarrow$ & PSNR$\uparrow$ & SSIM$\uparrow$ & LPIPS$\downarrow$ & PSNR$\uparrow$ & SSIM$\uparrow$ & LPIPS$\downarrow$ \\
\midrule
\multirow{3}{*}{\begin{tabular}[c]{@{}c@{}}Motion\\deblurring\end{tabular}} & w/o projection & 28.29 & 0.806 & 0.329 & 26.01 & 0.743 & 0.385 & 24.42 & 0.695 & 0.424 & 25.33 & 0.667 & 0.455 & 23.41 & 0.592 & 0.541 & 22.11 & 0.549 & 0.598 \\
 & \textbf{w/ projection} & 30.06 & 0.847 & 0.193 & 27.27 & 0.774 & 0.250 & 25.36 & 0.710 & 0.280 & 27.98 & 0.777 & 0.251 & 25.37 & 0.677 & 0.356 & 23.63 & 0.580 & 0.412 \\
 & {\bfseries\boldmath $\Delta$} & {\bfseries\boldmath +1.77} & {\bfseries\boldmath +0.041} & {\bfseries\boldmath $-$0.136} & {\bfseries\boldmath +1.26} & {\bfseries\boldmath +0.031} & {\bfseries\boldmath $-$0.135} & {\bfseries\boldmath +0.94} & {\bfseries\boldmath +0.015} & {\bfseries\boldmath $-$0.144} & {\bfseries\boldmath +2.65} & {\bfseries\boldmath +0.110} & {\bfseries\boldmath $-$0.204} & {\bfseries\boldmath +1.96} & {\bfseries\boldmath +0.085} & {\bfseries\boldmath $-$0.185} & {\bfseries\boldmath +1.52} & {\bfseries\boldmath +0.031} & {\bfseries\boldmath $-$0.186} \\
\midrule
\multirow{3}{*}{\begin{tabular}[c]{@{}c@{}}Inpainting\end{tabular}} & w/o projection & 30.72 & 0.852 & 0.285 & 27.60 & 0.771 & 0.368 & 25.07 & 0.706 & 0.433 & 26.01 & 0.700 & 0.428 & 23.69 & 0.625 & 0.526 & 21.72 & 0.564 & 0.619 \\
 & \textbf{w/ projection} & 34.46 & 0.938 & 0.111 & 30.63 & 0.853 & 0.176 & 27.44 & 0.785 & 0.239 & 31.43 & 0.892 & 0.155 & 27.79 & 0.789 & 0.264 & 25.33 & 0.661 & 0.366 \\
 & {\bfseries\boldmath $\Delta$} & {\bfseries\boldmath +3.74} & {\bfseries\boldmath +0.086} & {\bfseries\boldmath $-$0.174} & {\bfseries\boldmath +3.03} & {\bfseries\boldmath +0.082} & {\bfseries\boldmath $-$0.192} & {\bfseries\boldmath +2.37} & {\bfseries\boldmath +0.079} & {\bfseries\boldmath $-$0.194} & {\bfseries\boldmath +5.42} & {\bfseries\boldmath +0.192} & {\bfseries\boldmath $-$0.273} & {\bfseries\boldmath +4.10} & {\bfseries\boldmath +0.164} & {\bfseries\boldmath $-$0.262} & {\bfseries\boldmath +3.61} & {\bfseries\boldmath +0.097} & {\bfseries\boldmath $-$0.253} \\
\midrule
\multirow{3}{*}{\begin{tabular}[c]{@{}c@{}}Super-\\resolution\end{tabular}} & w/o projection & 25.33 & 0.731 & 0.402 & 22.98 & 0.669 & 0.456 & 20.86 & 0.613 & 0.509 & 23.13 & 0.569 & 0.562 & 21.75 & 0.534 & 0.612 & 20.23 & 0.502 & 0.679 \\
 & \textbf{w/ projection} & 28.78 & 0.818 & 0.241 & 26.64 & 0.758 & 0.284 & 24.71 & 0.701 & 0.321 & 26.65 & 0.721 & 0.344 & 24.87 & 0.647 & 0.401 & 23.23 & 0.595 & 0.453 \\
 & {\bfseries\boldmath $\Delta$} & {\bfseries\boldmath +3.45} & {\bfseries\boldmath +0.087} & {\bfseries\boldmath $-$0.161} & {\bfseries\boldmath +3.66} & {\bfseries\boldmath +0.089} & {\bfseries\boldmath $-$0.172} & {\bfseries\boldmath +3.85} & {\bfseries\boldmath +0.088} & {\bfseries\boldmath $-$0.188} & {\bfseries\boldmath +3.52} & {\bfseries\boldmath +0.152} & {\bfseries\boldmath $-$0.218} & {\bfseries\boldmath +3.12} & {\bfseries\boldmath +0.113} & {\bfseries\boldmath $-$0.211} & {\bfseries\boldmath +3.00} & {\bfseries\boldmath +0.093} & {\bfseries\boldmath $-$0.226} \\
\midrule
\multirow{3}{*}{\begin{tabular}[c]{@{}c@{}}Gaussian\\deblurring\end{tabular}} & w/o projection & 28.25 & 0.783 & 0.317 & 26.88 & 0.740 & 0.363 & 25.50 & 0.701 & 0.399 & 25.52 & 0.666 & 0.442 & 24.19 & 0.613 & 0.509 & 23.07 & 0.569 & 0.555 \\
 & \textbf{w/ projection} & 28.79 & 0.815 & 0.239 & 27.39 & 0.774 & 0.263 & 26.24 & 0.740 & 0.285 & 26.93 & 0.729 & 0.337 & 25.62 & 0.672 & 0.362 & 24.70 & 0.639 & 0.403 \\
 & {\bfseries\boldmath $\Delta$} & {\bfseries\boldmath +0.54} & {\bfseries\boldmath +0.032} & {\bfseries\boldmath $-$0.078} & {\bfseries\boldmath +0.51} & {\bfseries\boldmath +0.034} & {\bfseries\boldmath $-$0.100} & {\bfseries\boldmath +0.74} & {\bfseries\boldmath +0.039} & {\bfseries\boldmath $-$0.114} & {\bfseries\boldmath +1.41} & {\bfseries\boldmath +0.063} & {\bfseries\boldmath $-$0.105} & {\bfseries\boldmath +1.43} & {\bfseries\boldmath +0.059} & {\bfseries\boldmath $-$0.147} & {\bfseries\boldmath +1.63} & {\bfseries\boldmath +0.070} & {\bfseries\boldmath $-$0.152} \\
\bottomrule
\end{tabular}}
\caption{Ablation study on the manifold projection. We compare our full method (w/ projection) against the same sampler without projection on FFHQ and ImageNet (100 test images each) under $\sigma_\epsilon\in\{0.05, 0.15, 0.3\}$. $\Delta$: w/ projection minus w/o projection; positive $\Delta$PSNR/SSIM and negative $\Delta$LPIPS indicate improvement.}
\label{tab:ablation_proj}
\end{table}

As shown in above table, manifold projection consistently and substantially improves reconstruction quality: it improves all three metrics in every one of the 24 settings, with an average gain of 2.47\,dB PSNR and 0.080 SSIM, and a 38\% relative reduction in LPIPS. The improvement is not driven by a few outliers. The gains are largest on inpainting (up to 5.42\,dB on ImageNet) and SR (up to 3.85\,dB on FFHQ), where the measurements leave a large part of the image
unconstrained and the unprojected iterates are prone to drift away from the data manifold. Even on Gaussian deblurring, where the gap is smallest, the projection still yields consistent improvements of 0.51--1.63\,dB. These results confirm that manifold projection is the main source of the performance of PPS.

\begin{figure}[htbp]
    \centering
    \includegraphics[width=0.79\textwidth]{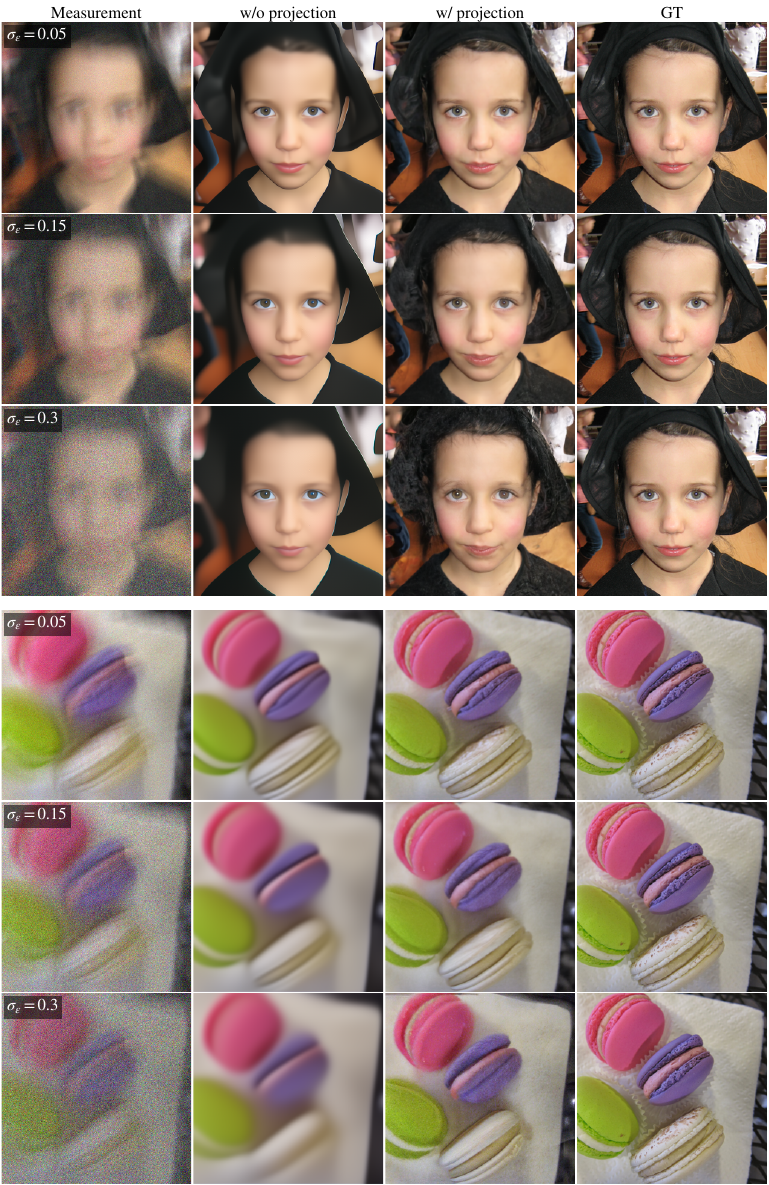}
    \caption{Qualitative comparison with and without manifold projection for motion deblurring. Top three rows: FFHQ. Buttom three rows: ImageNet}
    \label{fig:ablation_proj_MB}
\end{figure}

\begin{figure}[htbp]
    \centering
    \includegraphics[width=0.79\textwidth]{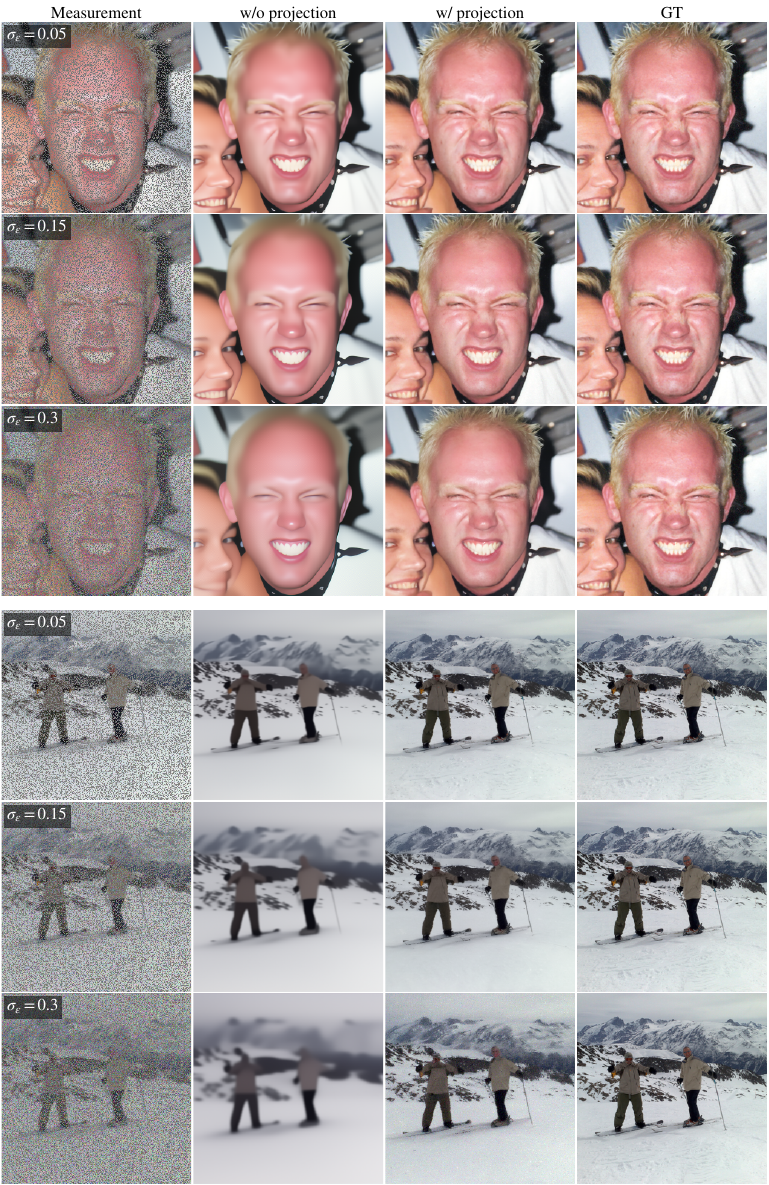}
    \caption{Qualitative comparison with and without manifold projection for motion inpainting. Top three rows: FFHQ. Buttom three rows: ImageNet}
    \label{fig:ablation_proj_IP}
\end{figure}

\begin{figure}[htbp]
    \centering
    \includegraphics[width=0.79\textwidth]{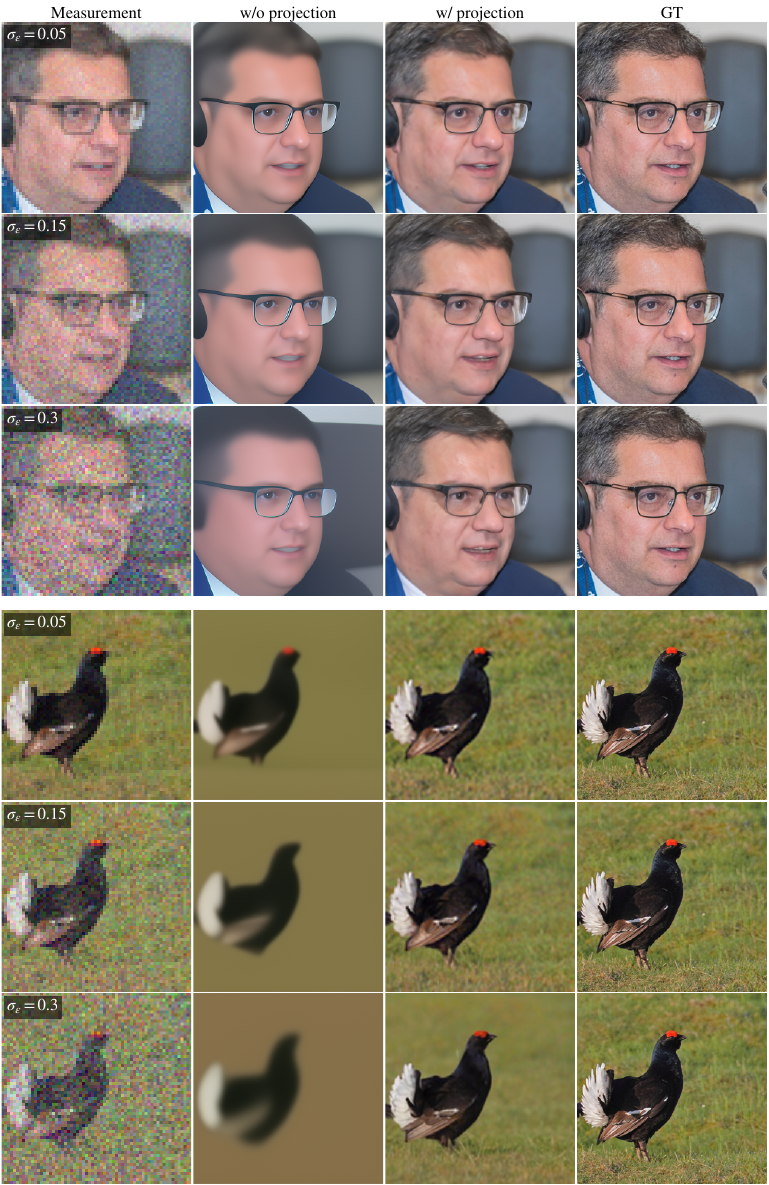}
    \caption{Qualitative comparison with and without manifold projection for SR. Top three rows: FFHQ. Buttom three rows: ImageNet}
    \label{fig:ablation_proj_SR}
\end{figure}

\begin{figure}[htbp]
    \centering
    \includegraphics[width=0.79\textwidth]{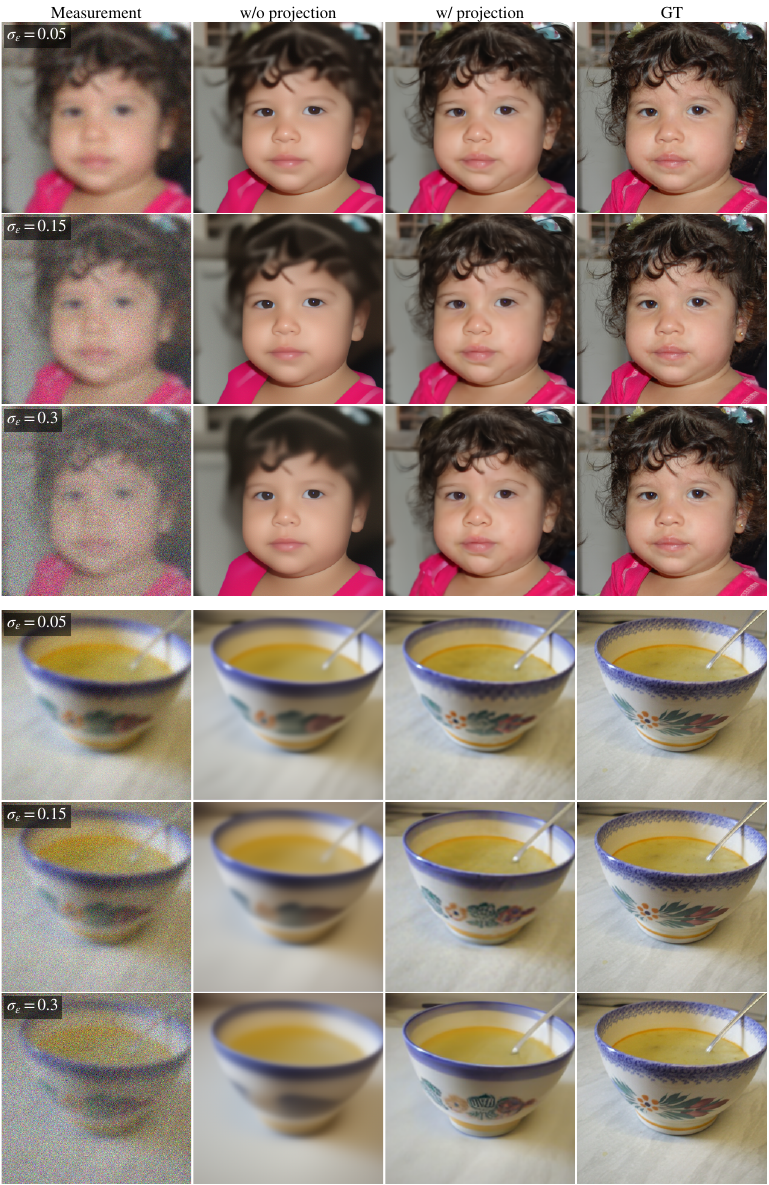}
    \caption{Qualitative comparison with and without manifold projection for Gaussian deblurring. Top three rows: FFHQ. Buttom three rows: ImageNet}
    \label{fig:ablation_proj_GB}
\end{figure}

For qualitative comparison, \cref{fig:ablation_proj_MB}--\ref{fig:ablation_proj_GB} present representative reconstructions obtained with and without manifold projection, highlighting the effect of the projection step. 
Across all settings, removing the manifold projection leads to a consistent failure mode:
the reconstructions recover the coarse structure and colors of the scene but are noticeably over-smoothed, and high-frequency details are largely lost, \eg, hair strands and skin texture on FFHQ, and fine textures such as grass, feathers, and decorative patterns on ImageNet.
This blurring becomes more pronounced as the measurement noise increases,
and at $\sigma_\epsilon = 0.3$ the outputs are close to a smooth, low-frequency approximation of the ground truth.
In contrast, manifold projection faithfully restores sharp edges and fine textures that match the ground truth at all noise levels, without amplifying the measurement noise. Since this behavior is shared across inverse problems with very different forward operators, we attribute it to the projection step itself rather than to a particular measurement operator.

\FloatBarrier

\subsection{Additional Ablation Study 2: inference strategies and the approximation variance}
\label{sm:sub:additional_ablation_study}


\begin{table}[H]
\centering
\setlength{\tabcolsep}{3pt}
\resizebox{\textwidth}{!}{%
\begin{tabular}{cclccccccccc|ccccccccc}
\toprule
\multirow{3}{*}{Task} & \multirow{3}{*}{Strategy} & \multirow{3}{*}{$\widetilde{\sigma}_{i-1}^2$} & \multicolumn{9}{c|}{FFHQ} & \multicolumn{9}{c}{ImageNet} \\
\cmidrule(lr){4-12}\cmidrule(lr){13-21}
 &  &  & \multicolumn{3}{c}{$\sigma_\epsilon=0.05$} & \multicolumn{3}{c}{$\sigma_\epsilon=0.15$} & \multicolumn{3}{c|}{$\sigma_\epsilon=0.3$} & \multicolumn{3}{c}{$\sigma_\epsilon=0.05$} & \multicolumn{3}{c}{$\sigma_\epsilon=0.15$} & \multicolumn{3}{c}{$\sigma_\epsilon=0.3$} \\
\cmidrule(lr){4-6}\cmidrule(lr){7-9}\cmidrule(lr){10-12}\cmidrule(lr){13-15}\cmidrule(lr){16-18}\cmidrule(lr){19-21}
 &  &  & PSNR$\uparrow$ & SSIM$\uparrow$ & LPIPS$\downarrow$ & PSNR$\uparrow$ & SSIM$\uparrow$ & LPIPS$\downarrow$ & PSNR$\uparrow$ & SSIM$\uparrow$ & LPIPS$\downarrow$ & PSNR$\uparrow$ & SSIM$\uparrow$ & LPIPS$\downarrow$ & PSNR$\uparrow$ & SSIM$\uparrow$ & LPIPS$\downarrow$ & PSNR$\uparrow$ & SSIM$\uparrow$ & LPIPS$\downarrow$ \\
\midrule
\multirow{6}{*}{\begin{tabular}[c]{@{}c@{}}Motion\\deblurring\end{tabular}} & \multirow{3}{*}{\begin{tabular}[c]{@{}c@{}}Direct posterior\\sampling\end{tabular}} & Local & 28.40 & 0.807 & 0.279 & 26.32 & 0.749 & 0.313 & 24.82 & 0.704 & 0.336 & 26.21 & 0.710 & 0.382 & 24.39 & 0.640 & 0.442 & 23.12 & \textbf{0.592} & 0.483 \\
 &  & Marginal & 28.14 & 0.801 & 0.284 & 26.15 & 0.744 & 0.317 & 24.70 & 0.700 & 0.339 & 25.82 & 0.698 & 0.395 & 24.17 & 0.633 & 0.451 & 22.99 & 0.587 & 0.490 \\
 &  & \textbf{Dirac} & 29.75 & 0.842 & 0.237 & \underline{27.24} & \textbf{0.777} & 0.276 & \textbf{25.57} & \textbf{0.729} & 0.300 & \underline{27.70} & \underline{0.768} & \underline{0.307} & \textbf{25.37} & \textbf{0.677} & \textbf{0.355} & \underline{23.53} & 0.566 & \underline{0.416} \\
\cmidrule{2-21}
 & \multirow{3}{*}{\begin{tabular}[c]{@{}c@{}}Manifold\\projection\end{tabular}} & Local & \underline{29.92} & \underline{0.845} & \underline{0.202} & 27.04 & 0.771 & \underline{0.275} & \underline{25.38} & \underline{0.721} & \underline{0.297} & 26.28 & 0.642 & 0.321 & \underline{24.40} & \underline{0.641} & 0.439 & 23.11 & \textbf{0.592} & 0.478 \\
 &  & Marginal & 29.90 & \underline{0.845} & 0.205 & 26.98 & 0.770 & 0.278 & 25.36 & 0.720 & 0.300 & 27.13 & 0.747 & 0.326 & 24.21 & 0.635 & 0.447 & 23.00 & \underline{0.589} & 0.486 \\
 &  & \textbf{Dirac} & \textbf{30.06} & \textbf{0.847} & \textbf{0.193} & \textbf{27.27} & \underline{0.774} & \textbf{0.250} & 25.36 & 0.710 & \textbf{0.280} & \textbf{27.98} & \textbf{0.777} & \textbf{0.251} & \textbf{25.37} & \textbf{0.677} & \underline{0.356} & \textbf{23.63} & 0.580 & \textbf{0.412} \\
\midrule
\multirow{6}{*}{\begin{tabular}[c]{@{}c@{}}Inpainting\end{tabular}} & \multirow{3}{*}{\begin{tabular}[c]{@{}c@{}}Direct posterior\\sampling\end{tabular}} & Local & 24.09 & 0.729 & 0.325 & 22.82 & 0.700 & 0.349 & 21.30 & 0.661 & 0.385 & 23.64 & 0.659 & 0.409 & 22.58 & 0.623 & 0.444 & 21.34 & 0.580 & 0.495 \\
 &  & Marginal & 23.07 & 0.709 & 0.339 & 21.97 & 0.684 & 0.362 & 20.60 & 0.647 & 0.396 & 22.36 & 0.626 & 0.436 & 21.48 & 0.597 & 0.469 & 20.38 & 0.559 & 0.517 \\
 &  & \textbf{Dirac} & 30.43 & \underline{0.850} & 0.230 & 28.04 & 0.800 & 0.273 & \underline{25.59} & 0.745 & 0.312 & 28.34 & \underline{0.800} & 0.276 & \underline{26.39} & 0.733 & 0.337 & \underline{24.54} & \underline{0.667} & 0.395 \\
\cmidrule{2-21}
 & \multirow{3}{*}{\begin{tabular}[c]{@{}c@{}}Manifold\\projection\end{tabular}} & Local & \underline{34.37} & \textbf{0.938} & \textbf{0.111} & \underline{29.28} & \underline{0.827} & \underline{0.213} & 25.54 & \underline{0.750} & \underline{0.285} & \underline{31.38} & \textbf{0.892} & \textbf{0.155} & 26.15 & \underline{0.738} & \underline{0.322} & 24.51 & \textbf{0.672} & \underline{0.369} \\
 &  & Marginal & 34.33 & \textbf{0.938} & \underline{0.112} & 29.01 & 0.821 & 0.219 & 25.11 & 0.742 & 0.291 & 31.34 & \textbf{0.892} & \underline{0.156} & 25.53 & 0.720 & 0.339 & 24.00 & 0.659 & 0.382 \\
 &  & \textbf{Dirac} & \textbf{34.46} & \textbf{0.938} & \textbf{0.111} & \textbf{30.63} & \textbf{0.853} & \textbf{0.176} & \textbf{27.44} & \textbf{0.785} & \textbf{0.239} & \textbf{31.43} & \textbf{0.892} & \textbf{0.155} & \textbf{27.79} & \textbf{0.789} & \textbf{0.264} & \textbf{25.33} & 0.661 & \textbf{0.366} \\
\midrule
\multirow{6}{*}{\begin{tabular}[c]{@{}c@{}}Super-\\resolution\end{tabular}} & \multirow{3}{*}{\begin{tabular}[c]{@{}c@{}}Direct posterior\\sampling\end{tabular}} & Local & 27.03 & 0.766 & 0.296 & 25.24 & 0.716 & 0.328 & 23.43 & 0.667 & 0.360 & 24.77 & 0.641 & 0.447 & 23.39 & 0.598 & 0.487 & 21.91 & 0.555 & 0.529 \\
 &  & Marginal & 26.96 & 0.764 & 0.297 & 25.18 & 0.714 & 0.329 & 23.39 & 0.665 & 0.361 & 24.69 & 0.638 & 0.450 & 23.33 & 0.596 & 0.489 & 21.85 & 0.553 & 0.530 \\
 &  & \textbf{Dirac} & 28.29 & 0.806 & 0.254 & \underline{26.42} & \underline{0.752} & \underline{0.294} & \underline{24.60} & \textbf{0.701} & \underline{0.328} & 26.58 & \underline{0.716} & 0.362 & \textbf{24.93} & \textbf{0.656} & \underline{0.408} & \textbf{23.30} & \textbf{0.597} & \textbf{0.453} \\
\cmidrule{2-21}
 & \multirow{3}{*}{\begin{tabular}[c]{@{}c@{}}Manifold\\projection\end{tabular}} & Local & \underline{28.46} & \underline{0.812} & \underline{0.243} & 26.20 & 0.745 & 0.298 & 24.40 & \underline{0.692} & 0.330 & \underline{26.63} & \textbf{0.721} & \textbf{0.341} & 24.52 & 0.640 & 0.425 & 21.90 & 0.555 & \underline{0.526} \\
 &  & Marginal & 28.45 & 0.811 & 0.244 & 26.18 & 0.744 & 0.298 & 24.38 & \underline{0.692} & 0.331 & 26.62 & \textbf{0.721} & \underline{0.342} & 24.50 & 0.639 & 0.426 & 21.85 & 0.554 & 0.528 \\
 &  & \textbf{Dirac} & \textbf{28.78} & \textbf{0.818} & \textbf{0.241} & \textbf{26.64} & \textbf{0.758} & \textbf{0.284} & \textbf{24.71} & \textbf{0.701} & \textbf{0.321} & \textbf{26.65} & \textbf{0.721} & 0.344 & \underline{24.87} & \underline{0.647} & \textbf{0.401} & \underline{23.23} & \underline{0.595} & \textbf{0.453} \\
\midrule
\multirow{6}{*}{\begin{tabular}[c]{@{}c@{}}Gaussian\\deblurring\end{tabular}} & \multirow{3}{*}{\begin{tabular}[c]{@{}c@{}}Direct posterior\\sampling\end{tabular}} & Local & 28.01 & 0.793 & 0.269 & 26.78 & 0.756 & 0.292 & 25.67 & 0.724 & 0.312 & 26.00 & 0.688 & 0.395 & 24.99 & 0.648 & 0.431 & 24.07 & 0.615 & 0.460 \\
 &  & Marginal & 27.97 & 0.792 & 0.270 & 26.75 & 0.756 & 0.293 & 25.64 & 0.723 & 0.313 & 25.97 & 0.687 & 0.397 & 24.96 & 0.647 & 0.432 & 24.05 & 0.614 & 0.461 \\
 &  & \textbf{Dirac} & \underline{28.51} & \underline{0.808} & \textbf{0.239} & \underline{27.24} & \underline{0.770} & \underline{0.266} & \underline{26.16} & \underline{0.738} & \underline{0.287} & 26.85 & 0.724 & 0.343 & \textbf{25.70} & \textbf{0.678} & \underline{0.375} & \underline{24.68} & \underline{0.638} & \textbf{0.400} \\
\cmidrule{2-21}
 & \multirow{3}{*}{\begin{tabular}[c]{@{}c@{}}Manifold\\projection\end{tabular}} & Local & 28.46 & 0.806 & \underline{0.247} & 27.17 & 0.768 & 0.271 & 26.07 & 0.736 & 0.290 & \underline{26.86} & \underline{0.725} & \textbf{0.327} & 25.50 & 0.669 & 0.389 & 24.06 & 0.615 & 0.459 \\
 &  & Marginal & 28.44 & 0.806 & 0.248 & 27.16 & 0.768 & 0.271 & 26.06 & 0.736 & 0.291 & \underline{26.86} & \underline{0.725} & \underline{0.328} & 25.49 & 0.669 & 0.390 & 24.04 & 0.614 & 0.460 \\
 &  & \textbf{Dirac} & \textbf{28.79} & \textbf{0.815} & \textbf{0.239} & \textbf{27.39} & \textbf{0.774} & \textbf{0.263} & \textbf{26.24} & \textbf{0.740} & \textbf{0.285} & \textbf{26.93} & \textbf{0.729} & 0.337 & \underline{25.62} & \underline{0.672} & \textbf{0.362} & \textbf{24.70} & \textbf{0.639} & \underline{0.403} \\
\bottomrule
\end{tabular}}
\caption{Ablation study on the inference combination on FFHQ and ImageNet (100 test images each) under $\sigma_\epsilon\in\{0.05, 0.15, 0.3\}$. \textbf{Bold}: best; \underline{underline}: second best for each task, dataset and noise level. Local: Averaging under the local Gaussian prior. Marginal: Averaging under the exact marginal prior. Dirac: Dirac approximation.}
\label{tab:ablation_strategy}
\end{table}

Above table ablates the two design choices of our sampling step: the inference strategy (direct posterior sampling \emph{vs.}\ manifold projection) and the three choices of $\widetilde{\sigma}_{i-1}^2$ in \cref{para:Measurement_Likelihood}.

\textbf{Manifold projection is crucial.} Replacing direct posterior sampling
with manifold projection improves PSNR by 1.61\,dB on average and reduces
LPIPS in 70 of 72 settings, with the largest gains on inpainting (up to
11.26\,dB). It also makes the sampler considerably less sensitive to the
choice of $\widetilde{\sigma}_{i-1}^2$.

\textbf{The Dirac approximation is the most effective.} Setting
$\widetilde{\sigma}_{i-1}^2$ to the Dirac variant outperforms the local and
marginal variants by 1.33\,dB PSNR on average, and achieves a higher PSNR
in 94 of 96 comparisons.

Combining both, our final approach (manifold projection with Dirac approximation) achieves the best result in 57 of the 72 combinations.

\subsection{Quantitative Comparison On 1000 Samples}
We evaluate our method and several representative baselines on 1,000 test samples. For a more comprehensive comparison, we additionally report the Fréchet Inception Distance (FID)~\cite{heusel2018ganstrainedtimescaleupdate}, computed using an ImageNet-pretrained Inception-v3 network~\cite{szegedy2015rethinking}. Unlike PSNR, SSIM and LPIPS, FID compares the feature distributions of generated and real images without requiring paired correspondence. Thus, it complements the reference-based metrics by quantifying the realism and diversity of the outputs at the distribution level. Secondly, FID estimates the mean and covariance of 2048-dimensional Inception features, a small sample size yields a highly rank-deficient covariance estimate (rank at most N-1) and leads to unstable computation. So, that is why we report FID on 1000 samples rather than 100 samples.


\begin{table}[H]
\centering
\setlength{\tabcolsep}{3.5pt}
\resizebox{\textwidth}{!}{%
\begin{tabular}{clcccc|cccc|cccc}
\toprule
\multirow{2}{*}{Task} & \multirow{2}{*}{Method} & \multicolumn{4}{c|}{$\sigma_\epsilon=0.05$} & \multicolumn{4}{c|}{$\sigma_\epsilon=0.15$} & \multicolumn{4}{c}{$\sigma_\epsilon=0.3$} \\
\cmidrule(lr){3-6}\cmidrule(lr){7-10}\cmidrule(lr){11-14}
 &  & PSNR$\uparrow$ & SSIM$\uparrow$ & LPIPS$\downarrow$ & FID$\downarrow$ & PSNR$\uparrow$ & SSIM$\uparrow$ & LPIPS$\downarrow$ & FID$\downarrow$ & PSNR$\uparrow$ & SSIM$\uparrow$ & LPIPS$\downarrow$ & FID$\downarrow$ \\
\midrule
\multicolumn{14}{c}{\textbf{FFHQ}} \\
\midrule
\multirow{3}{*}{\begin{tabular}[c]{@{}c@{}}Motion\\deblurring\end{tabular}} & MPGD & 26.10 & 0.712 & 0.280 & 40.32 & 23.28 & 0.627 & 0.327 & \underline{37.34} & 21.90 & 0.565 & 0.371 & \underline{43.12} \\
 & CoDPS & \underline{27.16} & \underline{0.750} & \underline{0.252} & \underline{32.30} & \underline{24.94} & \underline{0.689} & \underline{0.307} & 39.69 & \underline{23.25} & \underline{0.637} & \underline{0.353} & 48.22 \\
\cmidrule{2-14}
 & \textbf{Ours} & \textbf{29.89} & \textbf{0.845} & \textbf{0.194} & \textbf{27.06} & \textbf{27.08} & \textbf{0.771} & \textbf{0.253} & \textbf{35.32} & \textbf{25.19} & \textbf{0.707} & \textbf{0.283} & \textbf{31.32} \\
\midrule
\multirow{5}{*}{\begin{tabular}[c]{@{}c@{}}Inpainting\end{tabular}} & MPGD & 28.58 & 0.800 & 0.202 & 28.80 & 25.58 & 0.634 & 0.309 & 40.57 & 22.31 & 0.608 & 0.364 & \underline{39.41} \\
 & CoDPS & 30.54 & 0.865 & 0.150 & \underline{17.97} & 27.44 & 0.744 & 0.253 & 35.44 & 23.54 & 0.600 & 0.370 & 70.33 \\
 & DDRM & 29.67 & 0.853 & 0.200 & 36.02 & 29.38 & 0.845 & 0.204 & \underline{35.09} & \underline{27.74} & \underline{0.801} & 0.256 & 43.52 \\
 & DDNM & \underline{32.58} & \underline{0.908} & \underline{0.134} & 23.14 & \textbf{30.73} & \textbf{0.869} & \underline{0.195} & 35.44 & \textbf{28.87} & \textbf{0.827} & \underline{0.243} & 45.04 \\
\cmidrule{2-14}
 & \textbf{Ours} & \textbf{34.35} & \textbf{0.940} & \textbf{0.107} & \textbf{17.68} & \underline{30.67} & \underline{0.863} & \textbf{0.167} & \textbf{20.87} & 27.57 & 0.798 & \textbf{0.232} & \textbf{32.82} \\
\midrule
\multirow{5}{*}{\begin{tabular}[c]{@{}c@{}}Super-\\resolution\end{tabular}} & MPGD & 25.14 & 0.702 & 0.278 & \textbf{28.66} & 22.98 & 0.620 & 0.347 & \textbf{36.97} & 20.90 & 0.536 & 0.413 & \underline{48.13} \\
 & CoDPS & 27.61 & 0.772 & 0.273 & \underline{42.13} & 24.25 & 0.681 & 0.341 & 50.14 & 21.37 & 0.551 & 0.468 & 89.04 \\
 & DDRM & 28.48 & 0.811 & 0.248 & 48.29 & \underline{26.76} & \underline{0.766} & 0.295 & 58.01 & \underline{25.07} & \underline{0.724} & 0.333 & 64.97 \\
 & DDNM & \textbf{28.87} & \textbf{0.819} & \textbf{0.240} & 45.85 & \textbf{26.99} & \textbf{0.772} & \underline{0.288} & 53.70 & \textbf{25.15} & \textbf{0.726} & \underline{0.326} & 58.58 \\
\cmidrule{2-14}
 & \textbf{Ours} & \underline{28.53} & \underline{0.814} & \underline{0.244} & 42.53 & 26.46 & 0.755 & \textbf{0.287} & \underline{46.58} & 24.56 & 0.699 & \textbf{0.325} & \textbf{46.01} \\
\midrule
\multirow{5}{*}{\begin{tabular}[c]{@{}c@{}}Gaussian\\deblurring\end{tabular}} & MPGD & 26.16 & 0.713 & 0.260 & \textbf{31.54} & 24.35 & 0.650 & 0.305 & \textbf{35.61} & 22.72 & 0.605 & 0.333 & \textbf{33.09} \\
 & CoDPS & 26.78 & 0.743 & 0.283 & 47.38 & 26.10 & 0.726 & 0.286 & \underline{39.62} & 24.87 & 0.686 & 0.321 & \underline{44.44} \\
 & DDRM & \underline{28.67} & \underline{0.813} & \underline{0.248} & 49.02 & \underline{27.52} & \underline{0.781} & \underline{0.278} & 55.79 & \underline{26.51} & \underline{0.754} & \underline{0.302} & 60.11 \\
 & DDNM & \textbf{29.26} & \textbf{0.825} & 0.267 & 56.38 & \textbf{27.91} & \textbf{0.790} & 0.300 & 61.89 & \textbf{26.77} & \textbf{0.761} & 0.323 & 65.89 \\
\cmidrule{2-14}
 & \textbf{Ours} & 28.53 & 0.810 & \textbf{0.242} & \underline{45.70} & 27.17 & 0.769 & \textbf{0.266} & 45.53 & 26.04 & 0.735 & \textbf{0.288} & 45.36 \\
\midrule
\multicolumn{14}{c}{\textbf{ImageNet}} \\
\midrule
\multirow{3}{*}{\begin{tabular}[c]{@{}c@{}}Motion\\deblurring\end{tabular}} & MPGD & 21.93 & 0.533 & \underline{0.391} & \underline{51.31} & 20.02 & 0.427 & \underline{0.462} & \underline{61.51} & 18.58 & 0.348 & \underline{0.515} & \underline{72.37} \\
 & CoDPS & \underline{23.88} & \underline{0.547} & 0.417 & 60.03 & \underline{22.02} & \underline{0.491} & 0.474 & 81.65 & \underline{20.23} & \underline{0.432} & 0.528 & 103.40 \\
\cmidrule{2-14}
 & \textbf{Ours} & \textbf{27.93} & \textbf{0.793} & \textbf{0.239} & \textbf{26.32} & \textbf{25.07} & \textbf{0.688} & \textbf{0.344} & \textbf{44.07} & \textbf{23.30} & \textbf{0.590} & \textbf{0.405} & \textbf{54.38} \\
\midrule
\multirow{5}{*}{\begin{tabular}[c]{@{}c@{}}Inpainting\end{tabular}} & MPGD & 22.86 & 0.622 & 0.358 & 45.16 & 21.00 & 0.488 & 0.441 & 56.05 & 19.08 & 0.370 & 0.521 & 71.34 \\
 & CoDPS & 25.42 & 0.661 & 0.304 & 34.27 & 19.74 & 0.326 & 0.506 & 65.98 & 16.94 & 0.204 & 0.633 & 124.04 \\
 & DDRM & 27.43 & 0.796 & 0.259 & 38.23 & 27.23 & 0.788 & 0.262 & 37.67 & \underline{25.93} & \underline{0.737} & \underline{0.305} & \underline{42.65} \\
 & DDNM & \underline{30.57} & \underline{0.877} & \underline{0.155} & \underline{19.01} & \textbf{29.14} & \textbf{0.838} & \textbf{0.212} & \textbf{28.17} & \textbf{27.47} & \textbf{0.787} & \textbf{0.280} & \textbf{42.37} \\
\cmidrule{2-14}
 & \textbf{Ours} & \textbf{31.69} & \textbf{0.907} & \textbf{0.136} & \textbf{17.10} & \underline{27.97} & \underline{0.808} & \underline{0.239} & \underline{34.03} & 25.37 & 0.679 & 0.346 & 43.43 \\
\midrule
\multirow{5}{*}{\begin{tabular}[c]{@{}c@{}}Super-\\resolution\end{tabular}} & MPGD & 21.97 & 0.539 & 0.391 & \textbf{47.46} & 20.08 & 0.436 & 0.473 & \underline{59.91} & 17.99 & 0.330 & 0.553 & 79.72 \\
 & CoDPS & 25.00 & 0.660 & 0.369 & 54.03 & 21.26 & 0.479 & 0.504 & 88.66 & 18.58 & 0.308 & 0.636 & 143.30 \\
 & DDRM & \underline{26.27} & \underline{0.736} & \underline{0.323} & \underline{51.22} & \underline{24.85} & \underline{0.682} & \underline{0.382} & 62.15 & \underline{23.37} & \underline{0.629} & \underline{0.439} & \underline{74.55} \\
 & DDNM & \textbf{26.63} & \textbf{0.749} & \textbf{0.315} & 51.26 & \textbf{25.12} & \textbf{0.694} & \textbf{0.380} & 65.18 & \textbf{23.51} & \textbf{0.638} & \textbf{0.435} & 78.94 \\
\cmidrule{2-14}
 & \textbf{Ours} & 26.18 & 0.726 & 0.339 & 52.65 & 24.48 & 0.652 & 0.395 & \textbf{59.22} & 22.81 & 0.596 & 0.445 & \textbf{71.38} \\
\midrule
\multirow{5}{*}{\begin{tabular}[c]{@{}c@{}}Gaussian\\deblurring\end{tabular}} & MPGD & 22.10 & 0.525 & 0.372 & \textbf{47.99} & 20.95 & 0.461 & 0.430 & \textbf{54.31} & 19.89 & 0.434 & 0.449 & \textbf{57.47} \\
 & CoDPS & 24.00 & 0.561 & 0.417 & 63.41 & 23.31 & 0.553 & 0.446 & 71.45 & 22.18 & 0.505 & 0.491 & 84.97 \\
 & DDRM & \textbf{26.54} & \textbf{0.742} & \textbf{0.318} & \underline{50.95} & \textbf{25.47} & \textbf{0.697} & \underline{0.365} & 61.02 & \underline{24.55} & \textbf{0.661} & \underline{0.401} & 68.68 \\
 & DDNM & \underline{26.50} & \underline{0.738} & 0.340 & 58.49 & \underline{25.46} & \underline{0.695} & 0.385 & 66.79 & \textbf{24.60} & \underline{0.660} & 0.420 & 75.11 \\
\cmidrule{2-14}
 & \textbf{Ours} & 26.46 & 0.735 & \underline{0.329} & 54.74 & 25.15 & 0.678 & \textbf{0.356} & \underline{54.80} & 24.19 & 0.641 & \textbf{0.395} & \underline{61.09} \\
\bottomrule
\end{tabular}}
\caption{Quantitative comparison on FFHQ (60000-61000) and ImageNet (official validation set 0-1000) under measurement noise $\sigma_\epsilon\in\{0.05, 0.15, 0.3\}$. \textbf{Bold}: best; \underline{underline}: second best for each task, dataset and noise level.}
\label{tab:main_1000}
\end{table}

Compared with ~\cref{tab:main_results} (100 test samples), our approach still ranks first on all four metrics at every noise level on both datasets under motion deblurring. On the remaining tasks DDNM attains slightly higher PSNR/SSIM on SR and Gaussian deblurring (0.36/0.52\,dB on average on FFHQ/ImageNet), whereas our reconstructions are perceptually better: on FFHQ we obtain lower LPIPS in 8 of 9 and lower FID in all 9 same settings. Notably, methods that achieve a low FID on some tasks, such as MPGD, do so at a substantial cost in fidelity (3.8--5.3\,dB lower PSNR), whereas our method remains among the top performers on other both metrics. The main exception is only inpainting on ImageNet at $\sigma_\epsilon\geq0.15$.

Overall, for the five methods evaluated in both datasets, the metrics change only marginally (mean absolute difference of 0.26\,dB PSNR, 0.007 SSIM and 0.007 LPIPS), and the best-performing method is unchanged in 69 of 72 cells. The three exceptions are swaps between DDRM and DDNM on ImageNet and do not involve our method. This confirms that the conclusions drawn from the 100-image evaluation are statistically stable. It further proves PPS is robust enough in a larger test set.

\subsection{Comparison of Inference Speed}

\begin{table}[H]
\centering
\setlength{\tabcolsep}{8pt}
\begin{tabular}{lrr}
\toprule
\multirow{2}{*}{Method} & \multicolumn{2}{c}{Time (s/image)$\downarrow$} \\
\cmidrule(lr){2-3}
 & \multicolumn{1}{c}{FFHQ} & \multicolumn{1}{c}{ImageNet} \\
\midrule
DPS & 29.0 & 91.0 \\
DMPS & 23.7 & 94.0 \\
MPGD & 2.4 & 5.1 \\
DAPS & 32.0 & 61.0 \\
CoDPS & 2.0 & 4.7 \\
Peng et al. & 10.5 & 16.5 \\
DDRM & 0.5 & 1.1 \\
DDNM & 1.4 & 4.3 \\
\midrule
\textbf{Ours} & \textbf{4.2} & \textbf{7.2} \\
\bottomrule
\end{tabular}
\caption{Average inference time per image on FFHQ and ImageNet, measured on a single NVIDIA RTX PRO 6000 Blackwell Max-Q Workstation Edition GPU with batch size 1.}
\label{tab:speed}
\end{table}

Our method only requires 4.2\,s and 7.2\,s per image on FFHQ and ImageNet, respectively, which is $2.3$--$13.1\times$ faster than DPS, DMPS, DAPS and Peng et al., while also achieving higher PSNR than each of them in nearly all settings (\autoref{tab:main_results}). MPGD and CoDPS run faster than us, they save merely 1--3\,s per image but trail our method by 2.4--5.1\,dB PSNR on average. SVD-based methods (DDRM, DDNM) are the fastest, but they rely on a closed-form SVD of the forward operator and therefore do not apply to, \eg, motion deblurring. Overall, among methods that handle all four tasks, none is both faster and more accurate than ours, placing our approach on the Pareto front of speed and reconstruction quality.

\FloatBarrier

\subsection{Limitations}
We show some representative failure samples on ImageNet with heavy measurement noise ($\sigma_\varepsilon = 0.3$). On these samples, none of the compared methods recovers a faithful reconstruction, so the difficulty is not specific to our approach. We attribute it to two factors. First, ImageNet is far more diverse than FFHQ. FFHQ contains aligned, centered faces with a strong shared structure. ImageNet spans 1000 categories with arbitrary composition, scale, and illumination, and the target objects are often small or surrounded by cluttered backgrounds. Second, a single unconditional diffusion model must cover the entire ImageNet distribution, which makes it a much weaker prior than a face-specific model trained on FFHQ.

The failures nevertheless differ in nature and fall into three types:
\textbf{i. residual noise} where the measurement noise is not suppressed or is even amplified (DAPS, CoDPS); \textbf{ii. hallucination}, where sharp outputs are obtained by inventing content inconsistent with the GT (MPGD, Peng~\etal, and DPS in several cases); \textbf{iii. over-smoothing}, where reconstructions are blurry but lack fine texture (DDRM, DDNM, DMPS, and ours). Our method only belongs to the last category, but its errors remain largely consistent with the measurement: the global layout and object shapes are preserved, while fine details are lost. For inverse problems, we argue that hallucination are more harmful than blur.

\begin{figure}[htbp]
    \centering
    \includegraphics[width=\textwidth]{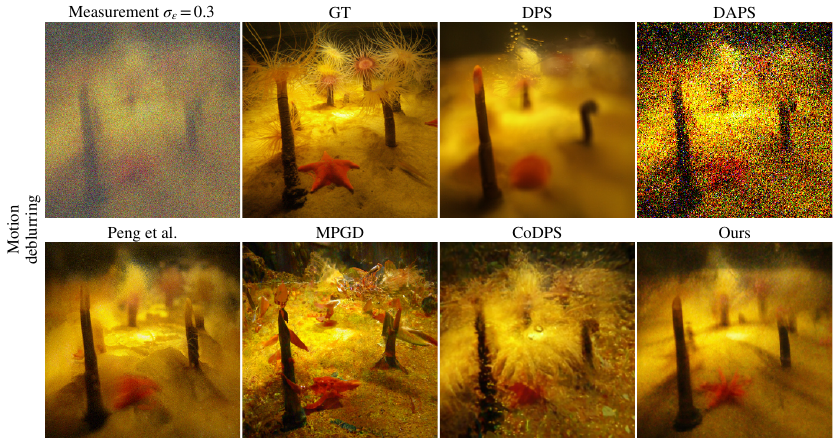}
    \caption{Failure case of motion deblurring on ImageNet ($\sigma_\varepsilon = 0.3$). DPS collapses the starfish into a round blob, removes the anemones, and adds spurious bubbles. DAPS is dominated by noise. MPGD fills the scene with leaf-like structures, and CoDPS covers the whole image with hair-like texture, losing the starfish. Peng~\etal turn the anemones into sand-like mounds. Our method loses the tentacles, but it recovers the stalks and a star-shaped starfish at the correct location.} 
\end{figure}

\begin{figure}[htbp]
    \centering
    \includegraphics[width=\textwidth]{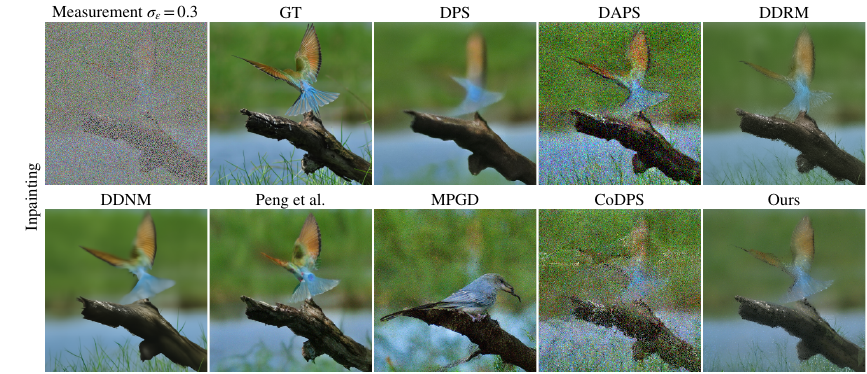}
    \caption{Failure case of inpainting on ImageNet ($\sigma_\varepsilon = 0.3$). MPGD replaces the bird with a different species, and Peng~\etal render a sharp, natural-looking bird, but the long left wing is truncated to a short stub. DAPS and CoDPS leave strong residual noise, and DDRM is noticeably grainy. DPS loses the bird's head and the wing structure. Our result is blurry, and the extended left wing is barely visible, but the pose, colors, and branch remain consistent with the GT.}
\end{figure}

\begin{figure}[htbp]
    \centering
    \includegraphics[width=\textwidth]{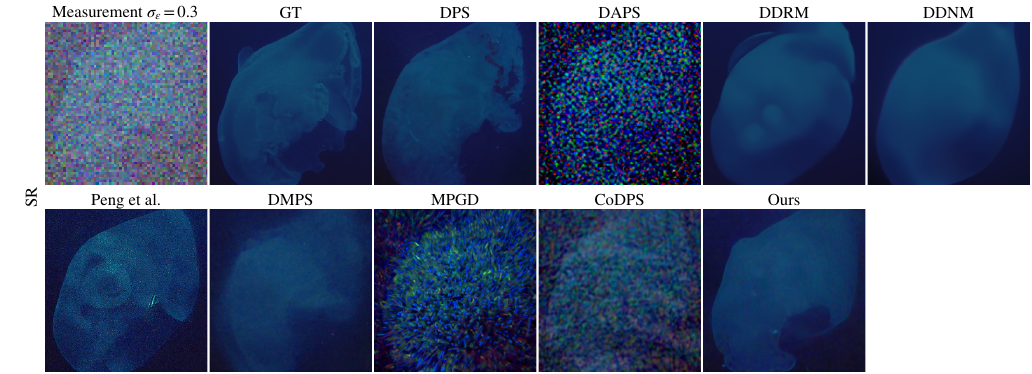}
    \caption{Failure case of SR on ImageNet ($\sigma_\varepsilon = 0.3$). This sample is one of the most challenging sample. It is a dark, low-contrast object with almost no texture. DAPS amplifies the measurement noise, MPGD hallucinates a spiky texture, and CoDPS leaves strong residual noise. Peng~\etal introduce a spurious ring structure and speckle artifacts. DPS recovers a plausible silhouette but adds dark stain-like artifacts on the right. DDRM, DDNM, and DMPS collapse into featureless blobs. Our result is over-smoothed and misplaces a dark region at the bottom, but it roughly preserves the object's
silhouette without introducing noise or false textures.}
\end{figure}

\begin{figure}[htbp]
    \centering
    \includegraphics[width=\textwidth]{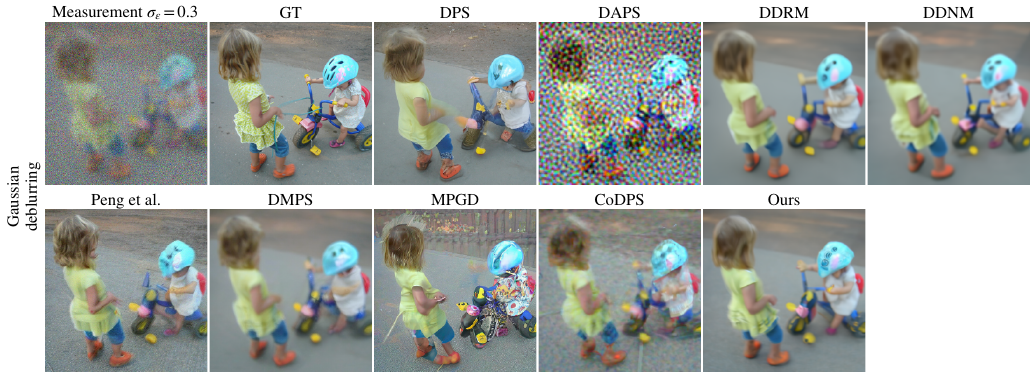}
    \caption{Failure case of Gaussian deblurring on ImageNet ($\sigma_\varepsilon = 0.3$). DPS produces a visually sharp result, but it turns the tricycle into a different object, smears the child's arm, and drops the cord. Peng~\etal distort the helmet and tricycle. MPGD hallucinates a fence in the background and patterned clothing. DAPS and CoDPS are corrupted by noise. DDRM, DDNM, DMPS, and our method are over-smoothed, but ours preserves the layout of both children and the tricycle.}
\end{figure}

We expect these failures to be alleviated mainly by a stronger prior model. Promising directions include larger diffusion models trained on more data, and class- or domain-specific priors, \eg, a model trained only on a given category, or conditioning the prior on a class label or text prompt inferred from the measurement. Since our method only requires a pretrained prior, it can directly benefit from such models.
\FloatBarrier

\subsection{More Qualitative Samples}

\begin{figure}[htbp]
    \centering
    \includegraphics[width=\textwidth]{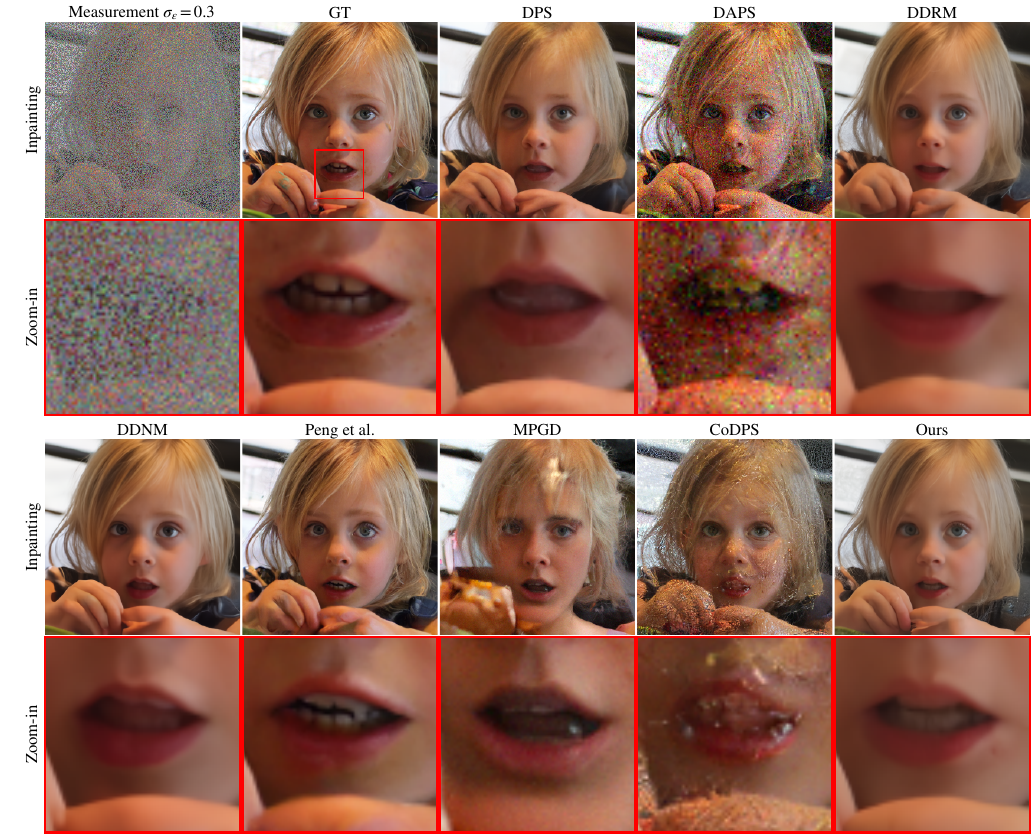}
    \caption{Qualitative comparison on FFHQ inpainting with $\sigma_\epsilon = 0.3$. DAPS and CoDPS leave strong residual noise and texture artifacts, while MPGD hallucinates content inconsistent with the measurement (\eg, an altered facial identity and spurious foreground objects). DDRM and DDNM over-smooth the mouth region and wash out the teeth, and Peng~\etal introduces jagged dark artifacts around the teeth. In contrast, our method produces a clean and natural mouth region with faithful lip contours, free of residual noise and hallucinated structures.}
    \label{fig:app_qualitative_FFHQ_IP03}
\end{figure}

\begin{figure}[htbp]
    \centering
    \includegraphics[width=\textwidth]{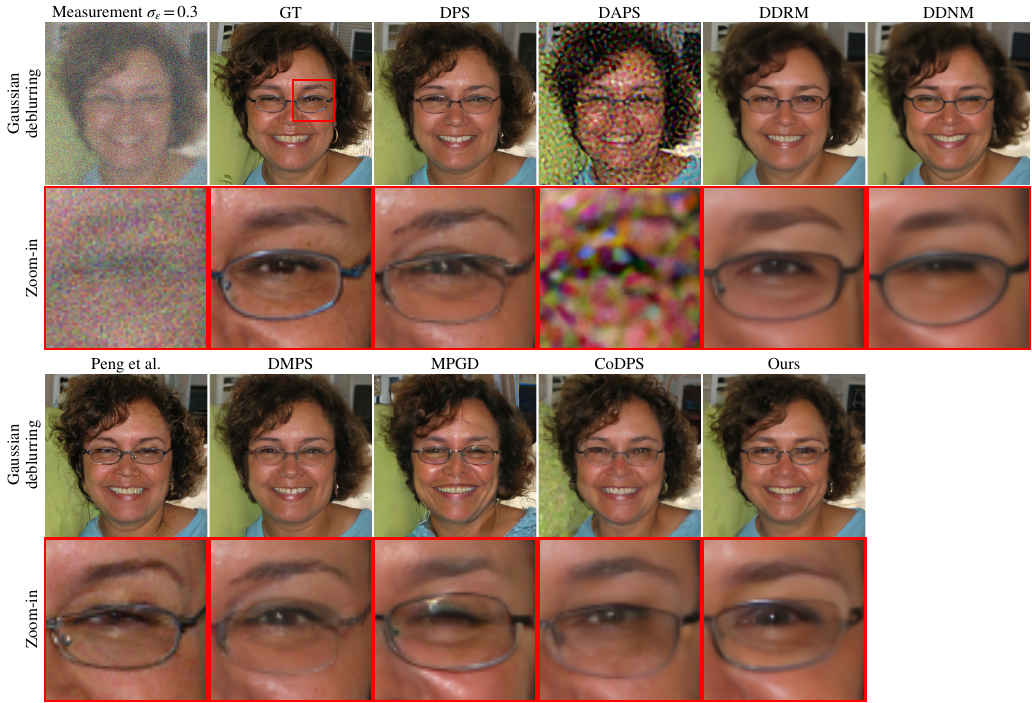}
    \caption{Qualitative comparison on FFHQ Gaussian deblurring with $\sigma_\epsilon = 0.3$. DAPS fails to suppress the noise and produces colored particles. DDRM, DDNM, DMPS, and CoDPS yield over-smoothed results in which the eyeglass frame and eye details are blurred or faded. DPS shows ghosted edges around the eyebrow, Peng~\etal introduces grainy textures and broken frame edges, and MPGD exhibits blocky artifacts and alters the background.
Our method recovers sharp eyeglass frames, a clear iris with its specular highlight, and clean skin, giving the result closest to the GT.}
    \label{fig:app_qualitative_FFHQ_GB03}
\end{figure}

\begin{figure}[htbp]
    \centering
    \includegraphics[width=\textwidth]{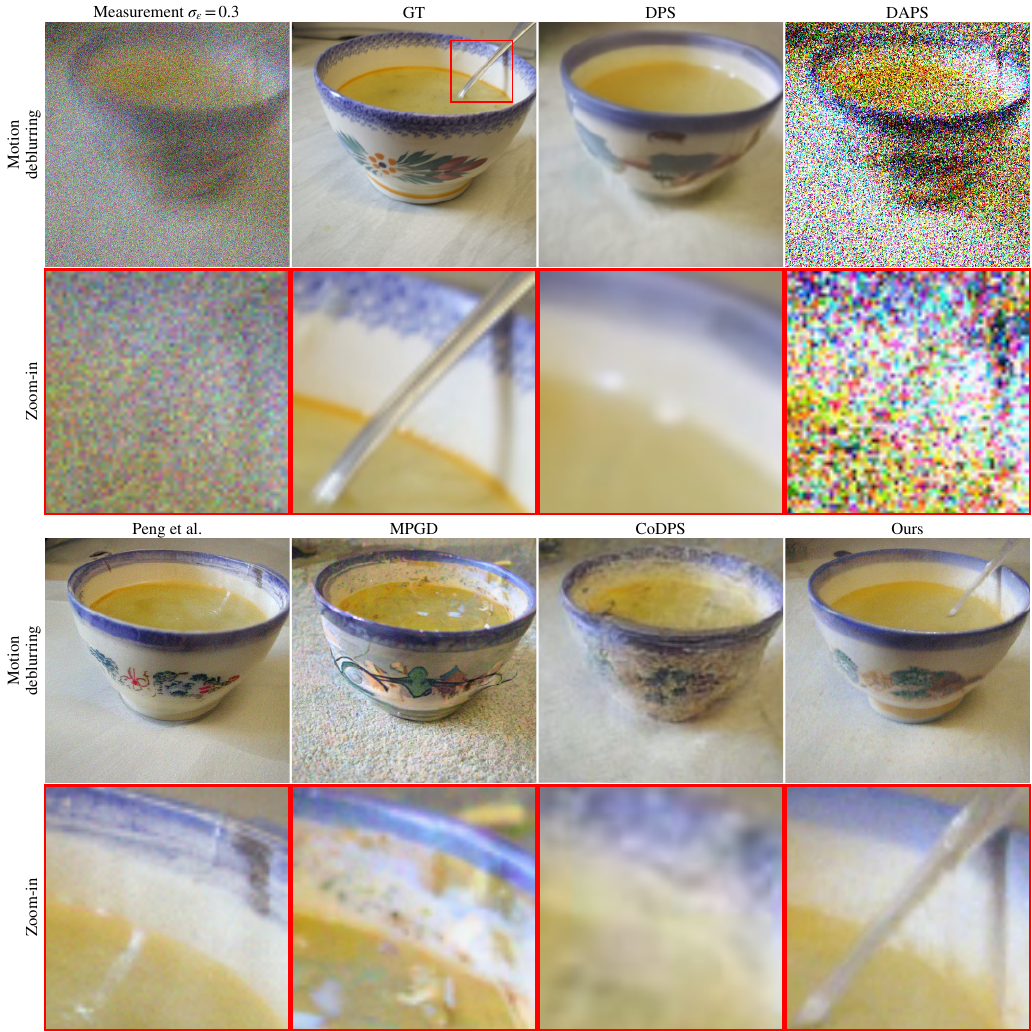}
    \caption{Qualitative comparison on ImageNet motion deblurring with $\sigma_\epsilon = 0.3$. DAPS is dominated by noise, DPS and CoDPS over-smooth the image and lose the spoon entirely, and MPGD hallucinates noisy textures on the bowl and tabletop. Peng~\etal recovers the overall bowl structure but only a fragment of the spoon.
Ours is the only method that recovers the thin spoon handle together with a clean bowl rim and background.}
    \label{fig:app_qualitative_FFHQ_GB03}
\end{figure}

\FloatBarrier

\end{document}